%% file: arxiv_files/arxiv_main.tex
\documentclass[11pt]{article}
\usepackage[final]{acl}

\usepackage{enumitem}
\usepackage{multirow}

\usepackage{times}
\usepackage{latexsym}
\usepackage{tabularx}
\usepackage{booktabs, makecell, ragged2e}
\usepackage{array}
\newcolumntype{L}[1]{>{\raggedright\arraybackslash}p{#1}}
\newcolumntype{C}[1]{>{\centering\arraybackslash}p{#1}}
\newcolumntype{Y}{>{\RaggedRight\arraybackslash}X} 
\usepackage{float}
\usepackage[most]{tcolorbox}
\newtcolorbox{promptbox}{
  colback=gray!10, colframe=black, boxrule=0.5pt, arc=3pt
}

\usepackage[T1]{fontenc}
\usepackage{rotating}
\usepackage{booktabs}
\usepackage[utf8]{inputenc}
\usepackage{amssymb}
\usepackage{microtype}
\usepackage{subcaption}

\usepackage{inconsolata}

\usepackage{graphicx}
\usepackage[dvipsnames]{xcolor}

\usepackage{threeparttable}
\usepackage{booktabs}

\definecolor{hexH}{HTML}{2D8B5F}    
\definecolor{hexE}{HTML}{C99635}    
\definecolor{hexX}{HTML}{E56B52}    
\definecolor{hexA}{HTML}{E67B91}    
\definecolor{hexC}{HTML}{6B5DA3}    
\definecolor{hexO}{HTML}{3D9598}    
\newcommand{\hexaco}{%
  {\textcolor{hexH}{H}}%
  {\textcolor{hexE}{E}}%
  {\textcolor{hexX}{X}}%
  {\textcolor{hexA}{A}}%
  {\textcolor{hexC}{C}}%
  {\textcolor{hexO}{O}}%
}

\newcommand{\honestyhumility}{%
  {\textcolor{hexH}{Honesty-Humility}}%
}

\newcommand{\emotionality}{%
  {\textcolor{hexE}{Emotionality}}%
}

\newcommand{\extraversion}{%
  {\textcolor{hexX}{eXtraversion}}%
}

\newcommand{\agreeableness}{%
  {\textcolor{hexA}{Agreeableness}}%
}

\newcommand{\conscientiousness}{%
  {\textcolor{hexC}{Conscientiousness}}%
}

\newcommand{\openness}{%
  {\textcolor{hexO}{Openness}}%
}

\makeatletter
\newcommand{\blfootnote}[1]{%
  \begingroup
  \renewcommand\thefootnote{}\footnote{#1}%
  \addtocounter{footnote}{-1}%
  \endgroup
}
\makeatother

\title{Measure what Matters: Psychometric Evaluation of AI with Situational Judgment Tests}

\author{
  Alexandra Yost\textsuperscript{1,2}\thanks{Equal contribution.}\quad
  Shreyans Jain\textsuperscript{1}\footnotemark[1]\quad
  Shivam Raval\textsuperscript{1,3}\thanks{Core contributor.}\quad
  Grant Corser\textsuperscript{2}\footnotemark[2]\quad
  Allen G.\ Roush\textsuperscript{1}\footnotemark[2]\\
  Nina Xu\textsuperscript{4}\footnotemark[2]\quad
  Jacqueline Hammack\textsuperscript{5}\thanks{Jacqueline's views do not represent Cedar City PD.}\quad
  Ravid Shwartz\mbox{-}Ziv\textsuperscript{6}\quad
  Amirali Abdullah\textsuperscript{1}\footnotemark[2]
}

\begin{document}
\maketitle

\blfootnote{\small
\textsuperscript{1} Thoughtworks \quad
\textsuperscript{2} Southern Utah University \quad
\textsuperscript{3} Harvard University \quad
\textsuperscript{4} NVIDIA \quad
\textsuperscript{5} Cedar City Police Department \quad
\textsuperscript{6} New York University (NYU).\\
\textbf{Correspondence: } \texttt{amir.abdullah@thoughtworks.com}, \texttt{allen.roush@thoughtworks.com}
}

\begin{abstract}

Persona conditioning is widely used to steer large language model (LLM) behavior, but it is unclear whether it induces stable behavioral structure or superficial variation. We propose a framework to measure consistent behavioral tendencies using situational judgment tests (SJTs), multidimensional item response theory (MIRT), and structured synthetic personas, treating responses as observations of latent behavioral variables.

Across large-scale SJT and persona datasets, we find that persona-conditioned behaviors are stable across runs, latent trait scores predict external benchmarks (e.g., TruthfulQA, EmoBench), and MIRT reveals consistent latent structure. We validate these results through human annotation, benchmark evaluation, and internal consistency analyses.

We interpret these traits not as human personality, but as stable behavioral tendencies expressed across contexts. Our results show that scenario-based psychometric evaluation provides a more reliable alternative to classical self-report approaches for assessing LLM behavior, and we release datasets to support further study.
\end{abstract}

\section{Introduction}

Large language models are increasingly deployed in socially grounded settings such as public safety \citep{application_safety_1, application_safety_2}, healthcare \citep{application_health_1, application_health_2}, and education \citep{application_education_1, application_education_2}. In many applications, practitioners steer model behavior through \emph{persona conditioning}, prompting the model to respond as a particular type of agent (e.g., a professional advisor, a cautious officer, or an empathetic counselor) \citep{luz2025helpful,personas_survey}. \citet{scaling_personas} extend this idea by generating synthetic data from a population of over one billion personas.

This growing use of personas raises a natural question: how can we systematically measure the behavioral tendencies that language models exhibit under different role assumptions? Prior work has often approached this question by adapting established human psychometric inventories such as the Big Five or HEXACO \citep{lu2023illuminating, pellert2024ai, serapio2023personality}.

However, the applicability of these instruments to language models is a topic of ongoing research, since they focus on self-report (``how strongly do you agree with this statement"), rather than the more behavioral style of AI safety benchmarks~\citep{emobench,truthfulqa,bias_bbq} where models choose between context grounded actions. Indeed, \citet{ecological_valid} show that these traditional generic item psychometric surveys yield misleading personality profiles compared to questionnaires grounded in realistic user-query contexts, sometimes exaggerating persona-prompted responses and failing to reveal stable behavioral traits. Complementing these concerns, \citet{contamination} highlight potential training-data contamination for these traditional personality inventories, suggesting that their measured traits may reflect memorization of well-known questionnaires rather than genuine behavioral tendencies. Together, these issues motivate new domain-grounded psychometric instruments that evaluate model behavior within realistic decision scenarios. We create such a psychometric instrument from scratch via expert annotations and then careful augmentation, and measure against established benchmarks.

\begin{figure*}[t]
\centering
\includegraphics[width=0.8\textwidth]{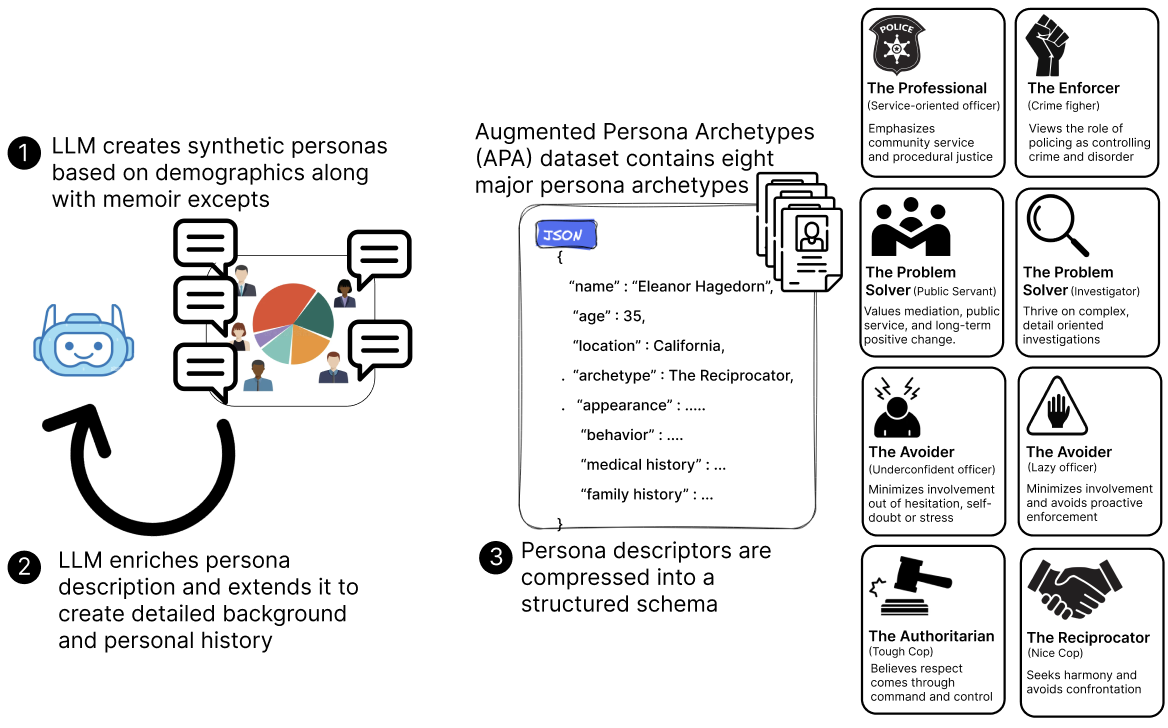}
\caption{Overview of the synthetic persona generation pipeline. We generate demographically grounded personas using memoir-based narrative seeds combined with structured demographic attributes, producing psychologically detailed profiles used to condition model behavior.}
\label{fig:method_personas}
\end{figure*}


\paragraph{Contributions.}

This paper makes four contributions toward systematic evaluation of persona and scenario conditioned behavior in language models.

\noindent \textbf{A brand new scenario-based psychometric instrument for measuring persona-conditioned behavior in the Law Enforcement Setting.}
We introduce an AI psychometrics framework based on situational judgment tests (SJTs) paired with demographically grounded synthetic personas. Unlike conventional HEXACO questionnaires, which rely on Likert-style agreement on self-report survey questions, our instrument presents detailed decision scenarios in which multiple plausible courses of action are offered, each corresponding to one of the six HEXACO behavioral traits. The SJTs are seeded from expert-authored law-enforcement scenarios and expanded through structured generation. See Dataset here: \url{https://github.com/amir-abdullah-thoughtworks/psychometrics_for_LLMs}

\noindent \textbf{Demographically grounded synthetic police officer personas.}
To evaluate persona-conditioned behavior, we construct structured synthetic personas grounded in principles from industrial–organizational and personality psychology. Persona profiles are generated from demographic priors together with narrative seed material derived from police memoirs, yielding profiles that capture behavioral tendencies, life histories, and psychosocial context. In our primary study we instantiate multiple archetypal behavioral styles (e.g., ``the Problem Solver'' and ``the Reciprocator'') \citep{broderick1986police, muir1979police} across diverse demographic subpopulations. 

We release large-scale datasets of SJTs, personas, and model responses enabling controlled behavioral evaluation. To demonstrate the extensibility of the methodology, we additionally construct a dataset of British parliamentarian personas spanning multiple ideological and behavioral axes.

\noindent \textbf{Empirical evidence of persona-conditioned behavioral effects.}
Analyzing 4M persona–scenario responses, we suggest that persona-conditioned behaviors are stable across repeated runs and dataset splits. Compared to the base model without persona conditioning, personas induce systematic shifts in model behavior along targeted competency axes. These persona-driven patterns also correlate with outcomes on external behavioral benchmarks such as TruthfulQA~\citep{truthfulqa} and EmoBench~\citep{emobench}. 

\noindent \textbf{Robust Correlation of Latent Traits with Benchmark Behaviours}
Our structured framework demonstrates stronger alignment of our trait scores with targeted behavioral benchmarks than conventional personality-inventory baselines such as \hexaco\;style prompting. The trait-scores and benchmark performance show robustness across multiple iterations. See Figure \ref{fig:trait_benchmark_corr}.


\begin{figure*}[t]
\centering
\includegraphics[width=0.5\textwidth]{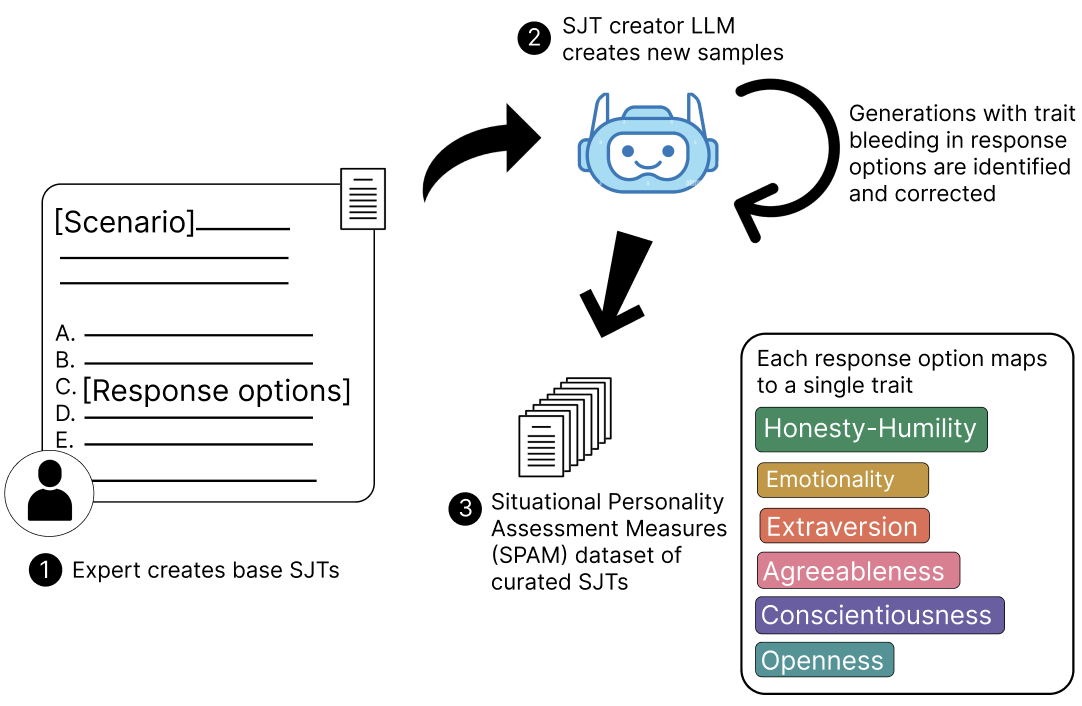}
\caption{Overview of the scenario generation pipeline. Domain experts first develop base scenarios with trait-mapped response options (step 1). An LLM then generates scenario variants through controlled attribute manipulation (step 2), followed by iterative quality refinement to identify and correct trait bleed instances. The final dataset contains response options that are uniquely aligned to the six \hexaco\; traits (step 3).}
\end{figure*}

\section{Background: Psychometric tests and validity}
\paragraph{\hexaco\;personality model.}
Psychometric personality inventories are standardized questionnaires designed to measure stable individual traits. The \hexaco\ is a contemporary personality framework comprising of six broad domains: \textcolor{hexH}{\textcolor{hexH}{Honesty–Humility}} (\textcolor{hexH}{H}), \textcolor{hexE}{Emotionality} (\textcolor{hexE}{E}), \textcolor{hexX}{eXtraversion} (\textcolor{hexX}{X}), \textcolor{hexA}{Agreeableness} (\textcolor{hexA}{A}), \textcolor{hexC}{Conscientiousness} (\textcolor{hexC}{C}), and \textcolor{hexO}{Openness to Experience} (\textcolor{hexO}{O}) \cite{ashton2007empirical}. It extends the popular Big Five (OCEAN) model by adding the \textcolor{hexH}{\textcolor{hexH}{Honesty–Humility}} trait. The \hexaco\; also defines \textcolor{hexA}{Agreeableness} and \textcolor{hexE}{Emotionality} somewhat differently, which is particularly relevant for prosocial personality profiles in both humans \cite{ashton2014hexaco, hilbig2014boast} and LLMs \cite{bodrovza2024personality}. The \hexaco-100 is a validated 100-item inventory that provides reliable scores for all six domain-level traits. The structure of the inventory and it's psychometric properties have been replicated across multiple languages and cultural contexts \cite{lee2018psychometric}. In human applications, the results of personality tests have predictive validity when they correlate with behavior \cite{pletzer2021hexaco}. 
For instance, high \textcolor{hexH}{Honesty–Humility} often predicts ethical decision-making \cite{ashton2008prediction}. This motivates our use of \hexaco\; for LLM personas as models can be meaningfully profiled using human personality inventories \cite{pellert2024ai, kruijssen2025deterministic, li2024automatic}. \hexaco\; trait scores represent latent personality variables that, under valid measurement conditions, should manifest consistently in behavioral choices \cite{west2012model}.
See Table \ref{tab:hexaco_example_items} for examples of Hexaco items and responses.

\begin{table*}[t]
\centering
\small
\begin{tabular}{p{3cm} p{9.5cm}}
\toprule
\textbf{Trait} & \textbf{Example Item (HEXACO-100)} \\
\midrule

\textcolor{hexH}{H — Honesty–Humility} &
``I would never accept a bribe, even if it were very large.'' \\

\textcolor{hexC}{C — Conscientiousness} &
``I plan ahead and organize my work carefully.'' \\

\midrule
\multicolumn{2}{p{12.5cm}}{\textbf{Response scale:} 
Strongly disagree, Disagree, Neutral, Agree, Strongly agree.} \\

\bottomrule
\end{tabular}
\caption{Illustrative items from the HEXACO-100 personality inventory. 
Respondents rate agreement with each statement using a Likert scale, and 
responses are aggregated to estimate scores for the six HEXACO domains. Our new instrument does not use such self-report structure.}
\label{tab:hexaco_example_items}
\end{table*}

\paragraph{Situational Judgment Tests (SJTs).}
Situational Judgment Tests (SJTs) are psychometric assessments that measure decision-making through realistic, role-relevant scenarios. Widely used in personnel selection and professional admissions, SJTs have demonstrated strong predictive validity for real-world performance \cite{lievens2016situational}. Each test presents a hypothetical but realistic situation and asks respondents to select the most appropriate responses \cite{weekley2015low}. 

\section{Related works}
\label{sec:related_works}
We lay out in Table \ref{tab:comparison} how our work compares to most of the highly cited AI psychometrics works on the axes of item level diagnostic, Multidimensional Item Response Theory (MIRT), external validation, safety relevant behaviors, persona conditioning, demographic grounding, the use of SJTs (namely domain-specific grounding), and the development of new psychometric benchmarks. See Appendix \ref{app:related_works} for discussions on persona simulation of LLMs and further background on application of psychometrics to AI, and Appendix \ref{app:comparison} for an expanded comparison.

\begin{table*}[h!]
\centering
\caption{Systematic Comparison of Psychometric Evaluation Frameworks for LLMs}
\label{tab:comparison}
\scriptsize
\setlength{\tabcolsep}{2.5pt}
\renewcommand{\arraystretch}{1.3}
\begin{tabular}{@{}p{3cm}ccccccccc@{}}
\toprule
\textbf{Related works} & 
\textbf{\shortstack{Item-level\\diagnostics}} & 
\textbf{\shortstack{Uses\\MIRT}} & 
\textbf{\shortstack{External\\validation}} & 
\textbf{\shortstack{Safety\\behavior}} & 
\textbf{\shortstack{Human\\annotation}} & 
\textbf{\shortstack{Persona\\conditioning}} & 
\textbf{\shortstack{Demographic\\grounding}} & 
\textbf{\shortstack{Developed\\SJTs}} & 
\textbf{\shortstack{Developed unique\\psychometric tests}} \\
\midrule

\textbf{This Work } & 
\textcolor{green!60!black}{\checkmark} & 
\textcolor{green!60!black}{\checkmark} & 
\textcolor{green!60!black}{\checkmark} & 
\textcolor{green!60!black}{\checkmark} & 
\textcolor{green!60!black}{\checkmark} & 
\textcolor{green!60!black}{\checkmark} & 
\textcolor{green!60!black}{\checkmark} & 
\textcolor{green!60!black}{\checkmark} & 
\textcolor{green!60!black}{\checkmark} \\
\midrule

\multicolumn{10}{l}{\textit{Personality \& trait assessment frameworks}} \\
\midrule

\cite{pellert2024ai} & 
\textcolor{green!60!black}{\checkmark} & 
\textcolor{red}{\texttimes} & 
\textcolor{orange}{$\bullet$} \textsuperscript{1} & 
\textcolor{green!60!black}{\checkmark} & 
\textcolor{red}{\texttimes} & 
\textcolor{red}{\texttimes} & 
\textcolor{green!60!black}{\checkmark} & 
\textcolor{red}{\texttimes} & 
\textcolor{red}{\texttimes} \\

\cite{serapio2023personality} & 
\textcolor{red}{\texttimes} & 
\textcolor{red}{\texttimes} & 
\textcolor{orange}{$\bullet$} \textsuperscript{1} & 
\textcolor{red}{\texttimes} & 
\textcolor{red}{\texttimes} & 
\textcolor{green!60!black}{\checkmark} & 
\textcolor{red}{\texttimes} & 
\textcolor{red}{\texttimes} & 
\textcolor{red}{\texttimes} \\

\cite{bhandari2025evaluating} & 
\textcolor{red}{\texttimes} & 
\textcolor{red}{\texttimes} & 
\textcolor{red}{\texttimes} & 
\textcolor{red}{\texttimes} & 
\textcolor{red}{\texttimes} & 
\textcolor{red}{\texttimes} & 
\textcolor{red}{\texttimes} & 
\textcolor{red}{\texttimes} & 
\textcolor{red}{\texttimes} \\

\cite{jiang2023personallm} & 
\textcolor{red}{\texttimes} & 
\textcolor{red}{\texttimes} & 
\textcolor{green!60!black}{\checkmark} & 
\textcolor{red}{\texttimes} & 
\textcolor{green!60!black}{\checkmark} & 
\textcolor{green!60!black}{\checkmark} & 
\textcolor{red}{\texttimes} & 
\textcolor{red}{\texttimes} & 
\textcolor{red}{\texttimes} \\

\cite{problem_agree_both} & 
\textcolor{green!60!black}{\checkmark} & 
\textcolor{red}{\texttimes} & 
\textcolor{red}{\texttimes} & 
\textcolor{red}{\texttimes} & 
\textcolor{red}{\texttimes} & 
\textcolor{green!60!black}{\checkmark} & 
\textcolor{red}{\texttimes} & 
\textcolor{red}{\texttimes} & 
\textcolor{red}{\texttimes} \\
\midrule

\multicolumn{10}{l}{\textit{Comprehensive psychological benchmarks}} \\
\midrule

\cite{li2024quantifying} & 
\textcolor{green!60!black}{\checkmark} & 
\textcolor{red}{\texttimes} & 
\textcolor{green!60!black}{\checkmark} & 
\textcolor{orange}{$\bullet$} \textsuperscript{2}& 
\textcolor{red}{\texttimes} & 
\textcolor{red}{\texttimes} & 
\textcolor{green!60!black}{\checkmark} & 
\textcolor{red}{\texttimes} & 
\textcolor{red}{\texttimes} \\

\cite{lee2025llms} & 
\textcolor{green!60!black}{\checkmark} & 
\textcolor{red}{\texttimes} & 
\textcolor{orange}{$\bullet$}\textsuperscript{3} & 
\textcolor{green!60!black}{\checkmark} & 
\textcolor{red}{\texttimes} & 
\textcolor{green!60!black}{\checkmark} & 
\textcolor{orange}{$\bullet$}\textsuperscript{4} & 
\textcolor{green!60!black}{\checkmark} & 
\textcolor{red}{\texttimes} \\

\cite{problem_variance_1} & 
\textcolor{red}{\texttimes} & 
\textcolor{red}{\texttimes} & 
\textcolor{green!60!black}{\checkmark} & 
\textcolor{red}{\texttimes} & 
\textcolor{red}{\texttimes} & 
\textcolor{red}{\texttimes} & 
\textcolor{red}{\texttimes} & 
\textcolor{red}{\texttimes} & 
\textcolor{red}{\texttimes} \\
\midrule

\multicolumn{10}{l}{\textit{Behavioral simulation frameworks}} \\
\midrule

\cite{park2024generative} & 
\textcolor{green!60!black}{\checkmark} & 
\textcolor{red}{\texttimes} & 
\textcolor{green!60!black}{\checkmark} & 
\textcolor{green!60!black}{\checkmark} & 
\textcolor{green!60!black}{\checkmark} & 
\textcolor{green!60!black}{\checkmark} & 
\textcolor{green!60!black}{\checkmark} & 
\textcolor{red}{\texttimes} & 
\textcolor{red}{\texttimes} \\
\bottomrule
\end{tabular}

\vspace{0.5em}
\begin{minipage}{\textwidth}
\footnotesize
\textbf{Additional details:} \textsuperscript{1} Validation used in \citet{pellert2024ai} and \citet{serapio2023personality} is primarily internal and compares convergent scores across tests. \textsuperscript{2} \textsuperscript{2} \citet{li2024quantifying} include values, motivation, and moral reasoning, but do not focus on toxicity, refusa,l or other safety-related metrics. \textsuperscript{3} \citet{lee2025llms} perform validation against psychometric quality metrics but not against direct human comparison. They provide situational contexts via knowledge graphs that are not tied to specific demographic profiles. While their validation is done entirely using LLMs, it lacks expert validation during initial data creation and to review the final instrument\textsuperscript{4} \\
\textbf{Key Distinctions of This Work:} Our approach combines (1) expert-designed, domain-grounded SJTs with controlled attribute variation, (2) personas integrating industrial-organizational psychology with demographic realism, and (3) external validation through expert review and safety-relevant behavioral assessment rarely integrated in prior psychometric evaluations of LLMs.
\end{minipage}
\end{table*}

\section{Evaluation and Metrics}

We evaluate the proposed persona and SJT datasets along four complementary dimensions: dataset diversity, behavioral validity, external applicability, and robustness \& stability. 


\paragraph{Experimental Setup}
We evaluate multiple variants from the Gemma~\citep{gemmateam2025gemma3technicalreport} and Qwen~\citep{yang2025qwen3technicalreport} families. For each model, we perform repeated sampling runs to account for stochasticity in generation. The responses are aggregated across multiple samples to obtain stable behavioral estimates. All experiments are conducted using a standardized prompting protocol to ensure consistent persona conditioning across models. Our results primarily focus on Gemma-3-4b-Instruct.

\noindent \textbf{Diversity Metrics}: We compute deterministic diversity metrics to quantify lexical and semantic variation across items (e.g., Per-Text Type/Token Ration (TTR), Compression Ration, Yule's K etc) and ensure that the dataset does not exhibit superficial regularities that could be exploited by language models.

\noindent\textbf{LLM-as-a-Judge Evaluation}: To evaluate semantic validity at scale, we employ an LLM-as-a-judge framework in which a model evaluates each dataset item according to a structured rubric covering clarity, realism, trait alignment, diversity, and originality. To assess the reliability of this evaluation process, 30 SJTs are manually annotated by domain experts (a psychologist and a patrol officer) using the same rubric. These human annotations are used to compute inter-rater agreement with the automated evaluations using metrics such as Cohen's \(\kappa\) \citep{cohen_kappa_original} and mean absoluate difference. To limit LLM self-preference bias \citep{preferences_1, preferences_2}, each example is evaluated by two independent LLM judges (\texttt{gpt-4o-mini} and \texttt{claude-3-5-sonnet}), whose ratings are aggregated alongside the human annotations.

\paragraph{Trait Validity Analysis}

To evaluate whether the dataset captures meaningful behavioral signals, we analyze correlations between latent traits inferred under an MIRT nominal response model, and SJT responses.

\paragraph{External Benchmark Evaluation}

To assess the applicability of the dataset in realistic settings, we evaluate persona-conditioned models on established benchmarks including TruthfulQA~\citep{truthfulqa} and EmoBench~\citep{emobench}.

\paragraph{Robustness and Stability}
To evaluate robustness, we repeat experiments across five sampling runs and analyze the stability of behavioral metrics across seeds and prompt variations.

\section{Persona generation}\label{sec:automatic_persona_creation}

We describe our procedure for constructing our police personas. Following the spirit of \citet{structured_attr_prompt}, which shows that targeted conditioning broadens data distributions, we \emph{inject explicit diversity} through multiple controlled channels.

\noindent \textbf{Overview.} Our pipeline selects demographics and an archetype, situates the persona in the milieu of a sampled police memoir, and conditions writing style with appearance/behavior categories via compact few-shot cues. We enforce field consistency with a structured outputs schema while encouraging stylistic variety through high-temperature sampling as well as randomized attribute injection for appearance and behavior control. The generation proceeds as follows:

 \noindent \textbf{1. Demographic Selection.} Each persona instantiates \emph{name, age, sex, location, education level, bachelor’s field, ethnic background, marital status} from a balanced officer roster generated from a probabilistic graphical model. These values are treated as ground-truth literals by all generated prose.
 
\noindent  \textbf{2. Archetype Assignment.} One archetype is drawn from the predefined set (e.g., Professional, Enforcer, Reciprocator, Avoider/Lazy, Avoider/Unconfident, Tough Cop, Problem Solver/Investigator, Problem Solver/Public Servant). The archetype serves as a soft prior to guide tendencies and tone without rigid constraints.

\noindent  \textbf{3. Memoir Grounding.} A memoir title and summary are sampled from the seeds list. The model first writes a narrative excerpt (180--250 words) as a concrete scene in the memoir’s milieu, remaining consistent with the demographics.

\noindent  \textbf{4. Appearance \& Behavior Conditioning.} We choose one \emph{appearance category} (e.g., Uniform/Official, 
Plainclothes/Casual) and one \emph{behavior category}. For each chosen category, a compact few-shot set of exemplars calibrates texture and specificity without dictating wording.

\noindent  \textbf{5. Schema-Constrained Prompting.} A dynamic Pydantic schema locks the selected demographics, memoir, and categories as literals, while free-text fields (e.g., \emph{appearance}, \emph{behavior}, \emph{speech}, \emph{mood\_affect}, brief histories, functioning summaries) must align with the memoir narrative.

To generate the data, we used \emph{GPT-4.1} \citep{openai2024gpt4technicalreport} with \textbf{temperature} $=2.0$ and \textbf{top-$p$} $=0.98$, parsed via structured outputs to the schema.

We ran a evaluation with both humans and LLM as a judge, assessing 55 personas on their clarity, originality, coherence, diversity, psychological depth, consistency, informativeness, ethical considerations and fidelity.  The average ratings are greater than 4 out of 5 across the board, for instance $4.27$ and $5$ for realism by the human and the LLM respectively. See Table \ref{tab:mean persona rater score}. 

Inter-rater agreement measured using Cohen’s Kappa \citep{cohen_kappa_original} averages approximately \textbf{0.68}, with values ranging from near-perfect agreement ($\kappa = 1$) to near-random ($\kappa \approx 0$) depending on the rubric criterion (Table \ref{tab:persona agreement score}). However, lower $\kappa$ values in some cases are expected due to a known statistical issue: when annotators exhibit very high agreement and ratings are concentrated within a narrow category range, Cohen’s $\kappa$ can artificially deflate toward zero \citep{high_agreement_low_kappa}.
To better capture agreement under these conditions, we additionally report Agreement Rate and Mean Absolute Deviation (MAD), see Table \ref{tab:persona mad}. Refer to \ref{box:persona_interagreement_example} for an illustrative case in which both the LLM judge (GPT-4.1) and human annotators assign a clarity score of 5. Despite this apparent agreement, the Cohen’s $\kappa$ value is low; however, this discrepancy is more appropriately reflected by the low Mean Absolute Deviation (MAD), which captures the close alignment between the ratings. For all scores, details and demographic statistics of the personas of the police officers, see Appendix \ref{sec:persona_stats}. Further details on the memoir titles, archetype definitions, and the appearance/behavior taxonomies are in Appendix~\ref{app:persona_seeds}. For details of the prompt used to generate personas, see Appendix \ref{app:personas_prompts}. Concrete samples of police officer personas are in Appendix~\ref{app:officer_examples}. 
For more details on the base probabilistic graphical model (PGM) behind our demographic data, see Appendix \ref{sec:persona_demographics}. 

To demonstrate the generalizability of our persona generation framework, we construct a second synthetic persona dataset in the parliamentary domain following the same generation pipeline described earlier. While the underlying persona synthesis procedures remain unchanged, we introduce domain-specific persona attributes (e.g., The Proceduralist, The Media Gladiator, The Policy Technocrat archetypes) and party contexts,  tailored to this setting. The resulting dataset contains 2200 personas across 10 archetype values, capturing a diverse set of behavioral patterns relevant to the parliamentary domain. This extension demonstrates that the proposed framework can be readily adapted to new domains without modifying the core generation methodology. For details on this dataset, see Appendix \ref{app:parliamentary_examples}. See Appendix \ref{app:engineering} for details on framework extensibility.

\begin{table}[!htb]
\centering
\caption{Average Human and LLM Judge Rubric Scores for Synthetic Personas (out of 5)}
\label{tab:mean persona rater score}
\footnotesize
\renewcommand\arraystretch{1.2}
\begin{tabularx}{\columnwidth}{@{}Xccc@{}}
\toprule
\textbf{Rubric} & \textbf{Human} & \textbf{GPT} & \textbf{Claude} \\
\midrule
Clarity                                         & 4.31 & 5.00 & 4.95 \\
Originality                                     & 4.60 & 4.04 & 3.93 \\
Coherence                                       & 4.67 & 5.00 & 4.91 \\
Diversity                                       & 4.96 & 3.509 & 3.91\\
Realism                                         & 4.27 & 5.00 & 4.78\\
Psychological Depth                             & 4.49 & 5.00 & 4.87\\
Consistency                                     & 4.73 & 4.50 & 4.80\\
Informativeness                                 & 4.51 & 5.00 & 4.82\\
Ethical Considerations                          & 4.85 & 5.00 & 4.80\\
Demographic Fidelity                            & 4.65 & 5.00 & 4.76\\
Overall Score                                   & 4.51 & 4.80 & 4.76\\
\bottomrule
\end{tabularx}
\end{table}

\begin{table}[!htb]
\centering
\caption{Inter Rater Agreement Score (Cohen's Kappa) for Synthetic Personas}
\label{tab:persona agreement score}
\footnotesize
\renewcommand\arraystretch{1.2}
\begin{tabularx}{\columnwidth}{@{}Xcc@{}}
\toprule
\textbf{Rubric} & \textbf{\shortstack{Human vs\\GPT}} & \textbf{\shortstack{GPT vs\\Claude}} \\
\midrule
Clarity                                         & 0.45 &  0.95 \\
Originality                                     & 0.25 &  0.93 \\
Coherence                                       & 0.71 &  0.93 \\
Diversity                                       & 0.13 &  0.26\\
Realism                                         & 0.38 &  0.87\\
Psychological Depth                             & 0.58 &  0.87\\
Consistency                                     & 0.80 &  0.91\\
Informativeness                                 & 0.53 &  0.82\\
Ethical Considerations                          & 0.89 &  0.87\\
Demographic Fidelity                            & 0.71 &  0.89\\
Overall Score                                   & 0.55 &  0.87\\
\bottomrule
\end{tabularx}
\end{table}

\begin{table}[!htb]
\centering
\caption{Inter Rater Mean Absolute Deviation for Synthetic Personas}
\label{tab:persona mad}
\footnotesize
\renewcommand\arraystretch{1.2}
\begin{tabularx}{\columnwidth}{@{}Xcc@{}}
\toprule
\textbf{Rubric} & \textbf{\shortstack{Human vs\\GPT}} & \textbf{\shortstack{GPT vs\\Claude}} \\
\midrule
Clarity                                         & 0.69 &  0.05\\
Originality                                     & 0.75 &  0.09\\
Coherence                                       & 0.33 &  0.09 \\
Diversity                                       & 1.45 &  0.78\\
Realism                                         & 0.73 &  0.22\\
Psychological Depth                             & 0.51 &  0.13\\
Consistency                                     & 0.27 &  0.20\\
Informativeness                                 & 0.49 &  0.18\\
Ethical Considerations                          & 0.14 &  0.20\\
Demographic Fidelity                            & 0.34 &  0.24\\
Overall Score                                   & 0.47 &  0.24\\
\bottomrule
\end{tabularx}
\end{table}

\section{SJT bank and controlled augmentation}

 We developed a multi-stage situational judgment test generation pipeline that integrates domain expertise with controlled instruction evolution and automated evaluation. Our framework creates a diverse and psychometrically robust set of SJTs, and is described in detail below.

\noindent\textbf{Base scenarios.} Foundational SJT samples were designed by psychology industrial organization and personality domain experts. The 20 scenarios spanned a broad range of contexts that law enforcement officers plausibly encounter, capturing variability in situational types (e.g., interpersonal conflicts, high-stakes emergencies, ethical dilemmas). Each scenario includes six reasonable and feasible responses, one per \hexaco\ trait. Each response was crafted to ensure fidelity and specificity to the personality domain. This expert-driven design ensured that each scenario remained grounded in realistic professional practice.

\noindent\textbf{Extracting and varying scenario descriptor attributes for diversity.} To avoid convergence on only the most probable or stereotypical scenarios, we decompose a scenario into psychologically and operationally relevant dimensions and extract the following attributes using an LLM: urgency, threat level, ambiguity, stakeholder complexity, authority relations, ethical tension, type of situation, time-of-day, and subject demographics (the latter controlled explicitly to avoid stereotyping). Each attribute can have one of several possible values, and by varying these initial seed values, we can steer the scenario generation to cover both high and low probability events, maintaining representational diversity in the SJT corpus (For ex:  a combination of seeds can be where the situation type is either a "crime scene investigation" or an "administrative reporting" situation along with a "high" urgency level and "clear" ambiguity level). For the full dataset, we create a random sample of all the possible seed combinations and ensure uniform coverage of all the variations of scenarios we considered.

\vspace{0.3cm}

\noindent \textbf{Diverse scenario generation through controlled attribute sampling.} Using the sampled attribute seeds, we can finely vary a scenario along a specific attribute dimension, creating a new scenario that maintains other qualitative aspects of the original. This is a controlled instruction-evolution process: a language model adapts the base template according to the sampled seed values and provides a systematic approach to maintaining the quality of domain expert-created scenarios while enabling fine-grained variation across scenarios. 

\vspace{0.3cm}

\noindent \textbf{Refining the dataset to mitigate trait bleed with LLM-as-a-Judge.} With direct feedback from the psychology domain experts, we observed that a single round of SJT creation can lead to scenarios with response options that can sometimes be mapped to an overlap across \hexaco\;  traits instead of uniquely mapping onto a specific personality trait, an effect we term the \textbf{trait bleed} of the scenario. This effect can add noise to downstream psychometric evaluations, as the evaluation requires a response to be mapped to a single trait. To mitigate trait bleeding, we use an LLM-as-a-judge to evaluate each option’s Trait Fit, assigning scores from 1 (Poor) to 5 (Very Strong). For response options with high trait bleed (Trait Fit Score < 5), we feed them back into the LLM to minimize overlap and sharpen their correspondence to an intended trait. We found this iterative refinement ensured the final SJT pool achieved both diversity and trait clarity.  We used GPT 4.1 with hyperparameters temperature=1.5, top\_p=0.95, presence\_penalty=0.4 and frequency\_penalty=0.3 for creation of SJT and correction for trait bleed. Synthetic SJTs and Corrections were performed using Outlines \citep{willard2023efficient} structured generation framework.

\vspace{0.3cm}

\noindent \textbf{Dataset quality and diversity metrics.} Across annotators (human experts and both LLM judges), annotations generated on 30 SJTs, ratings are consistently high, with an average score exceeding 4 out of 5 across all rubric axes (Table \ref{tab:sjt_mean_rater_score}). This indicates broad agreement that the generated scenarios are realistic, coherent, and behaviorally plausible. 
Similar to personas, to measure inter rater agreement, we calculate Cohen’s Kappa \citep{cohen_kappa_original} (Table \ref{tab:sjt_inter_rater_agreement_score}) and Mean Absolute Deviation (MAD) (for cases where cohen's kappa artificially deflates to zero), see Table \ref{tab:sjt_inter_rater_mad}. MAD in particular indicates the low disagreement between annotators, confirming that human experts and independent LLM judges largely converge on similar high-quality ratings.

Finally, diversity metrics suggest that the dataset exhibits substantial linguistic and scenario diversity. Specifically, MSTTR-100 (0.802), Compression Ratio (0.302), and Average Cosine Distance (0.445) all indicate a broad distribution of language and scenario structure across the dataset. For details on each component, please refer to these sections: on SJT Evaluation (Appendix \ref{sec:sjt_eval}), and Diversity and Statistical Analyses (Appendix \ref{sec:sjt_stats}).

\begin{table}[!htb]
\centering
\caption{Average Human and LLM Judge Rubric Scores for Synthetic SJTs (out of 5)}
\label{tab:sjt_mean_rater_score}
\footnotesize
\renewcommand\arraystretch{1.2}
\begin{tabularx}{\columnwidth}{@{}Xccc@{}}
\toprule
\textbf{Rubric} & \textbf{Human} & \textbf{GPT} & \textbf{Claude} \\
\midrule
Scenario Realism                                & 3.83 & 5.00 &  4.10\\
Ethical Tension                                 & 3.90 & 5.00 & 4.79 \\
Bias Fairness                                   & 5.00 & 5.00 & 4.77 \\
\textcolor{hexH}{Honesty-Humility} Alignment    & 5.00 & 5.00 & 4.94\\
\textcolor{hexE}{Emotionality} Alignment        & 5.00 & 5.00 & 4.69\\
\textcolor{hexX}{eXtraversion} Alignment        & 4.87 & 5.00 & 4.85\\
\textcolor{hexA}{Agreeableness} Alignment       & 4.97 & 4.50 & 4.76\\
\textcolor{hexC}{Conscientiousness} Alignment   & 5.00 & 5.00 & 4.89\\
\textcolor{hexO}{Openness} Alignment            & 5.00 & 5.00 & 4.89\\
\bottomrule
\end{tabularx}
\end{table}

\begin{table}[!htb]
\centering
\caption{Inter Rater Agreement Score (Cohen's Kappa) for Synthetic SJTs}
\label{tab:sjt_inter_rater_agreement_score}
\footnotesize
\renewcommand\arraystretch{1.2}
\begin{tabularx}{\columnwidth}{@{}Xcc@{}}
\toprule
\textbf{Rubric} & \textbf{\shortstack{Human vs\\GPT}} & \textbf{\shortstack{GPT vs\\Claude}} \\
\midrule
Urgency Level                                   & 0.53 & 0.80 \\
Threat Level                                    & 0.73 & 0.77 \\
Ambiguity Level                                 & 0.33 & 0.86 \\
Individuals Involved                            & 0.87 & 0.85 \\
Authority Relationships                         & 0.53 & 0.90 \\
Situation Type                                  & 0.67 & 0.92 \\
Time of Day                                     & 1.00 & 1.00 \\
Race                                            & 1.00 & 0.99 \\
Gender                                          & 1.00 & 1.00 \\
Age                                             & 0.97 & 0.98 \\
Scenario Realism                                & 0.40 & 0.15 \\
Ethical Tension                                 & 0.33 & 0.83 \\
Bias Fairness                                   & 1.00 & 0.85 \\
\textcolor{hexH}{Honesty-Humility} Alignment    & 1.00 & 0.93 \\
\textcolor{hexE}{Emotionality} Alignment        & 1.00 & 0.70 \\
\textcolor{hexX}{eXtraversion} Alignment        & 0.87 & 0.88 \\
\textcolor{hexA}{Agreeableness} Alignment       & 0.53 & 0.47 \\
\textcolor{hexC}{Conscientiousness} Alignment   & 1.00 & 0.89 \\
\textcolor{hexO}{Openness} Alignment            & 1.00 & 0.89 \\
\bottomrule
\end{tabularx}
\end{table}

\begin{table}[!htb]
\centering
\caption{Inter Rater Mean Absolute Deviation for Synthetic SJTs}
\label{tab:sjt_inter_rater_mad}
\footnotesize
\renewcommand\arraystretch{1.2}
\begin{tabularx}{\columnwidth}{@{}Xcc@{}}
\toprule
\textbf{Rubric} & \textbf{\shortstack{Human vs\\GPT}} & \textbf{\shortstack{GPT vs\\Claude}} \\
\midrule
Scenario Realism                                & 1.17 & 0.89 \\
Ethical Tension                                 & 1.10 & 0.19 \\
Bias Fairness                                   & 0.00 & 0.23 \\
\textcolor{hexH}{Honesty-Humility} Alignment    & 0.00 & 0.07 \\
\textcolor{hexE}{Emotionality} Alignment        & 0.00 & 0.31 \\
\textcolor{hexX}{eXtraversion} Alignment        & 0.13 & 0.15 \\
\textcolor{hexA}{Agreeableness} Alignment       & 0.47 & 0.53 \\
\textcolor{hexC}{Conscientiousness} Alignment   & 0.00 & 0.11 \\
\textcolor{hexO}{Openness} Alignment            & 0.00 & 0.11 \\
\bottomrule
\end{tabularx}
\end{table}

\section{Key Findings}
\label{sec:key_findings}
We summarize the major empirical observations emerging from the analyses we performed. Across behavioral evaluations, trait correlations, and stability tests, several 
consistent patterns emerge in how persona conditioning shapes model behavior. \\


\noindent\textbf{Finding 1: Behavioral Trait Measures Are More Predictive of Benchmark Performance than Self-Report Traits.}
Across nearly all dimensions, SJT-derived trait rankings exhibit stronger correlations with benchmark outcomes than their \hexaco\; self-report counterparts. For example, \agreeableness\; SJT correlates with EmoBench at $0.70$ compared to $0.51$ for \hexaco\; \agreeableness\; (Figure \ref{fig:trait_benchmark_corr}), suggesting that behavioral choice under situational pressure better captures the constructs underlying downstream benchmark performance. So, while trait correlations between \hexaco\; and SJTs fall within the range 0.25–0.35 (Figure \ref{fig:sjt_hexaco_corr}), typically considered evidence of convergent validity in personality measurement, the above pattern indicates that behaviorally grounded instruments such as SJTs may provide more predictive signals of model behavior than self-reported instruments.

\noindent\textbf{Finding 2: Model Personality Traits Align with External Benchmarks.}
SJT traits form two coherent clusters that align with the observed benchmark trade-offs. A warm–expressive cluster (\agreeableness\;, \openness\;, \emotionality\;, \honestyhumility\;) correlates positively with EmoBench but negatively with TruthfulQA, while a task–assertive cluster (\conscientiousness\;, \extraversion\;) shows the opposite pattern (Table \ref{tab:cluster_sjt_trait}). 


\noindent\textbf{Finding 3: Persona Conditioning Induces Systematic Behavioral Shifts.}
Persona conditioning consistently alters model behavior relative to the base model baseline across all evaluated instruments. Conditioned models exhibit a clear trade-off between emotional calibration and epistemic accuracy, with lower performance on EmoBench ($z=-1.06$) but higher performance on TruthfulQA ($z=+1.13$), see Table \ref{tab:basemodel_z_persona}. This shift is consistent across all eight archetypes, suggesting that persona framing reliably activates a distinct behavioral mode rather than producing random variation.

\noindent\textbf{Finding 4: Persona Archetypes Produce Discriminant Behavioral Profiles.}
Persona effects are not uniform across archetypes but instead produce structured and interpretable behavioral differences. For example, the Avoider archetype exhibits extreme suppression across emotional calibration(EmoBench $z = -5.22$), honesty related decisions (\honestyhumility\; SJT $z = -5.42$), and social initiative (\extraversion\; SJT $z = -4.22$), consistent with its disengaged character definition. Conversely, the Reciprocator archetype shows elevated \agreeableness\; (SJT $z = +2.02$) and \openness\; (SJT $z = +2.21$) scores, aligning with its socially cooperative role (see Table \ref{tab:archetype_z_base_model} for complete details). 


\noindent\textbf{Finding 5: Persona-Based Measurements Are Stable and Psychometrically Reliable Across Iterations.}
Repeated evaluations show near-perfect stability in model behavior across runs. Response distributions for EmoBench and TruthfulQA remain identical across iterations (mean Jensen-Shannon (JS) divergence = 0), while SJT distributions exhibit only negligible variation (mean JS divergence = 0.003). Correspondingly, trait-level measurements demonstrate strong reliability, with Intraclass Correlation Coefficient (ICC) values exceeding 0.86 across all \hexaco\; dimensions, indicating that persona-conditioned outputs produce stable and reproducible behavioral signals rather than stochastic artifacts. Refer to Sections 
\ref{app:trait_behaviour_benchmark},
\ref{app:persona_consistent_construct}, for more details on analyses.

\paragraph{Finding 6: The SJT instrument exhibits strong psychometric validity with structured behavioral variation.}
The SJT framework demonstrates strong psychometric properties: most items show good discrimination (76\% with $a > 1.0$), see Table \ref{tab:item_discrimination}, latent traits are reliably estimated, and test--retest consistency is high (ICC $> 0.86$, indicating stable trait scores across repeated runs). Latent traits significantly predict decisions (4/6 traits, $p < 0.05$), see Table \ref{tab:trait_prediction}, while scenario context explains a substantial share of variance (38.9\%) (Table \ref{tab:variance_decomposition}) and average persona effects another 3.8\%. The remaining variance (57.3\%) is not attributable to random noise, as evidenced by a low tie rate (5.3\%), suggesting additional structured variation beyond additive trait effects.




\section{Conclusions}
In this work, we introduce a novel scenario-based psychometric instrument for law enforcement settings, enabling the measurement of persona-conditioned model behavior through contextually grounded SJTs. We also propose a pipeline for generating demographically grounded synthetic personas across multiple domains, yielding psychologically coherent profiles for controlled conditioning. Our framework produces ecologically valid, domain-relevant insights, with effects observable even at fine-grained levels. We show that persona conditioning leads to consistent and statistically significant behavioral shifts, and that SJT-derived trait scores are strong predictors of external benchmark performance, highlighting the value of behaviorally grounded evaluation.

\section{Limitations}
This work is intended as a foundational step toward domain-grounded psychometric evaluation of persona and scenario conditioned language model behavior, and several design choices delimit its current scope. 

First, our evaluations focus on single-turn situational judgments. While multi-turn interaction is central to many real-world decision processes, especially in policing contexts, robust psychometric modeling of multi-turn, persona-conditioned LLM behavior remains an open problem. Extending SJTs to longitudinal or dialogic settings is a natural direction for future work.

Second, the British Parliamentarian persona dataset is not human-annotated by domain experts. Due to the absence of available subject-matter annotators, we rely instead on structured generation procedures and LLM-as-a-judge evaluations to assess internal consistency and persona- and scenario-conditioned behavioral alignment. This dataset serves as a proof of concept for the extensibility of our persona creation framework. However, incorporating expert human annotations would enable stronger external validation and comparative benchmarking. We also do not make situational judgment tests in this domain, again for lack of domain expert.

Third, we explicitly avoid anthropomorphic claims about internal mental states of language models. Our analyses are framed strictly in terms of observable, persona- and scenario-conditioned behavioral regularities. Whether such consistency reflects deeper representational structure or latent space is an open research question that we leave to future investigation, perhaps via approaches that consider mechanistic interpretability.

Fourth, while we employ multidimensional item response theory (MIRT) to analyze attribute structure, we do not explore higher-order or hierarchical MIRT formulations. Such models are rarely applied even in traditional psychometrics and pose additional identifiability and data requirements. Investigating hierarchical latent structures in persona-conditioned LLM behavior remains an important methodological frontier.

Finally, due to the specificity and technicality of SJT construction and assessment requiring psychological expertise, we do not yet employ external annotators for situational judgment test scoring. Expanding to broader expert pools or structured annotation protocols would further strengthen generalizability, and we leave this to future work.

\section{Ethical considerations}

The framework and datasets could be misused for psychometric profiling or behavioral surveillance in real-world law enforcement or recruitment contexts, leading to unethical discrimination or bias reinforcement. Model generations can potentially reflect learned biases and stereotypes \cite{bai2025explicitly, gallegos2024bias}; we recommend careful analysis before applying such systems in sensitive domains to mitigate risks of amplifying representational harms across minoritized demographic or ethnic groups. Moreover, correlations observed between traits and behaviors may be misinterpreted as genuine psychological evidence, encouraging anthropomorphisation of AI models and overgeneralization of synthetic findings to human populations. Finally, the framework’s predictive and analytical capabilities could be repurposed for coercive or manipulative applications, requiring the incorporation of responsible usage guidelines for policy and governance purposes.

\bibliography{references}

\newpage
\appendix
\onecolumn
\renewcommand{\thesection}{\Alph{section}}
\setcounter{section}{0}

\section*{\LARGE Appendix}
\addcontentsline{toc}{section}{Appendix}

\vspace{1em}
\large
\renewcommand{\baselinestretch}{1.5}\selectfont
\makeatletter
\renewcommand{\l@section}{\@dottedtocline{1}{0em}{2.3em}}
\renewcommand{\l@subsection}{\@dottedtocline{2}{2.3em}{3.2em}}
\makeatother
\makeatother


\contentsline{section}{\textbf{Methodology and Conceptual Framing}}{}{}

\contentsline{section}{\numberline{A}Positioning relative to prior works}{\pageref{app:comparison}}{}
\contentsline{section}{\numberline{B}Other Related works}{\pageref{app:related_works}}{}
\contentsline{section}{\numberline{C}Persona Design Methodology}{\pageref{sec:handmade_persona_creation}}{}
\contentsline{section}{\numberline{D}Case Studies}{\pageref{app:case_studies}}{}


\contentsline{section}{\textbf{Dataset Schema and Generation Framework}}{}{}

\contentsline{section}{\numberline{E}Data and response generation framework details}{\pageref{app:engineering}}{}
\contentsline{section}{\numberline{F}Hand Designed Persona Schema}{\pageref{app:persona_schema}}{}
\contentsline{section}{\numberline{G}Situational Judgment Tests (SJT) Schema}{\pageref{sec:sjt_schema}}{}
\contentsline{section}{\numberline{H}Persona Generation Seeds}{\pageref{app:persona_seeds}}{}
\contentsline{section}{\numberline{I}Bootstrapping Personas from Demographic Data}{\pageref{sec:persona_demographics}}{}


\contentsline{section}{\textbf{Statistical Analysis and Evaluation}}{}{}

\contentsline{section}{\numberline{J}Persona Statistics and Evaluation}{\pageref{sec:persona_stats}}{}
\contentsline{section}{\numberline{K}Situational Judgment Test Evaluation}{\pageref{sec:sjt_eval}}{}
\contentsline{section}{\numberline{L}Situational Judgment Test Statistics}{\pageref{sec:sjt_stats}}{}
\contentsline{section}{\numberline{M}Alignment Between SJT and Self-Report \hexaco\; Trait Representations}{\pageref{sec:alignment}}{}
\contentsline{section}{\numberline{N}Construct validity between SJT and Self-Report \hexaco Traits}{\pageref{sec:factor_experiments}}{}
\contentsline{section}{\numberline{O}Trait-Behavior Benchmark Coherence}{\pageref{app:trait_behaviour_benchmark}}{}
\contentsline{section}{\numberline{P}Personas as a Consistent Conditioning Construct}{\pageref{app:persona_consistent_construct}}{}
\contentsline{section}
{\numberline{Q}Psychometric Validation via Multidimensional IRT}{\pageref{app:mirt}}{}

\contentsline{subsection}
{\numberline{Q.1}Item discrimination}{\pageref{app:mirt_1}}{}

\contentsline{subsection}
{\numberline{Q.2}Variance decomposition}{\pageref{app:mirt_2}}{}

\contentsline{subsection}
{\numberline{Q.3}Latent trait recovery}{\pageref{app:mirt_3}}{}

\contentsline{subsection}
{\numberline{Q.4}Construct validity: Latent trait-behavior relationship}{\pageref{app:mirt_4}}{}


\contentsline{section}{\textbf{Prompt Templates and Dataset Examples}}{\pageref{app:prompt templates}}{}

\contentsline{section}{\numberline{R}SJT Creation and Evaluation Prompt Templates}{\pageref{app:prompt templates}}{}
\contentsline{section}{\numberline{S}Situational Judgment Test Examples}{\pageref{app:sjt_examples}}{}
\contentsline{section}{\numberline{T}Persona Prompts for Creation and Evaluation}{\pageref{app:personas_prompts}}{}
\contentsline{section}{\numberline{U}Police Officer Personas Examples}{\pageref{app:officer_examples}}{}
\contentsline{section}{\numberline{V}Parliamentary Personas Examples and Evaluation}{\pageref{app:parliamentary_examples}}{}


\contentsline{section}{\textbf{Administration and Discussion}}{}{}

\contentsline{section}{\numberline{W}Future Work}{\pageref{app:future_work}}{}
\contentsline{section}{\numberline{X}Models Used}{\pageref{app:models_used}}{}
\contentsline{section}{\numberline{Y}Annotators}{\pageref{app:use_of_annotators}}{}
\contentsline{section}{\numberline{Z}Use of AI Assistance}{\pageref{app:use_of_ai}}{}

\newpage

\section{Positioning relative to prior works}
\label{app:comparison}

Tables \ref{tab:full_comparison_1},\ref{tab:full_comparison_2} provides a systematic comparison of our framework against related psychometric evaluations of LLMs. Our work uniquely combines:

\noindent \textbf{Domain-grounded assessment design}: Unlike generic personality inventories \citep{pellert2024ai,serapio2023personality}, our SJTs are co-designed with domain experts for law enforcement contexts and validated against realistic scenarios.
    
\noindent \textbf{Rich persona construction}: While \citet{park2024generative} provide demographic grounding through interview transcripts, our personas integrate clinical psychological constructs with industrial-organizational frameworks, enabling deeper behavioral analysis.
    
\noindent \textbf{Behavioral validation}: Where prior work focuses on self-report consistency \citep{lee2025llms} or aggregate trait distributions \citep{problem_variance_1}, we validate through expert review of safety-relevant decision-making.
    
\noindent \textbf{Methodological rigor}: Our approach includes item-level diagnostics, external validation through expert assessment, and systematic evaluation of bias and fairness—capabilities rarely combined in prior work.

Our framework addresses three key limitations identified in the literature: (1) the variance problem in LLM personality expression \citep{problem_variance_1,problem_variance_2}, (2) the need for behaviorally grounded rather than self-report assessments \citep{problem_agree_both}, and (3) the requirement for domain-specific rather than generic evaluation \citep{mittelstadt2024large}.

\begin{table}[H]
\centering
\caption{Systematic Comparison of Psychometric Evaluation Frameworks for LLMs - Part 1}
\label{tab:full_comparison_1}
\tiny
\setlength{\tabcolsep}{4pt}
\renewcommand{\arraystretch}{1.15}
\begin{tabular}{@{}p{1cm}p{1.5cm}p{1cm}p{1.5cm}p{1.5cm}p{1.5cm}p{1.5cm}p{1.5cm}p{1.5cm}@{}}
\toprule
\textbf{Paper} & 
\textbf{Item-Level Diagnostics} & 
\textbf{Uses MIRT} & 
\textbf{External Validation} & 
\textbf{Safety-Relevant Behavior} & 
\textbf{Human Annotation} & 
\textbf{Persona Conditioning} & 
\textbf{Multi-Turn Interaction} & 
\textbf{Demographic Grounding} \\
\midrule

\textbf{This work} & 
\textcolor{green!70!black}{\checkmark} \textbf{Analyzes responses per SJT scenario} & 
\textcolor{green!70!black}{\checkmark} & 
\textcolor{green!70!black}{\checkmark} \textbf{Convergent validation via persona design, expert review, trait scores, and SJT outcomes} & 
\textcolor{green!70!black}{\checkmark} \textbf{Evaluates ethical tension, bias-fairness in law enforcement scenarios} & 
\textcolor{green!70!black}{\checkmark} \textbf{Experts crafted personas \& scenarios; human review of realism} & 
\textcolor{green!70!black}{\checkmark} \textbf{LLM personas with rich life histories and traits} & 
\textcolor{red}{\texttimes} Single-turn scenario responses per persona & 
\textcolor{green!70!black}{\checkmark} \textbf{Personas include demographic profiles; scenarios grounded in realistic contexts} \\
\midrule

\cite{pellert2024ai} & 
\textcolor{green!70!black}{\checkmark} Examines question-level response distributions; cross-language item differences & 
\textcolor{red}{\texttimes} & 
$\sim$ Limited: primarily validates internally (convergent scores across tests and languages; compares model traits to typical human range) & 
\textcolor{green!70!black}{\checkmark} Checks for biases in responses (e.g., gender pronoun effects on values) & 
\textcolor{red}{\texttimes} Used existing validated surveys; no new labeling of outputs & 
\textcolor{red}{\texttimes} Evaluates base models' inherent traits without persona prompts & 
\textcolor{red}{\texttimes} Single-turn questionnaire responses & 
\textcolor{green!70!black}{\checkmark} Compares outputs across languages and gendered versions of questions \\
\midrule

\cite{serapio2023personality} & 
\textcolor{red}{\texttimes} Focuses on aggregate trait scores and test-level consistency & 
\textcolor{red}{\texttimes} & 
$\sim$ Limited: checks convergent validity of two independent personality tests and reliability across prompts & 
\textcolor{red}{\texttimes} Does not directly evaluate bias, toxicity, etc., though discusses ethical implications & 
\textcolor{red}{\texttimes} No external labeling; uses model responses to surveys & 
\textcolor{green!70!black}{\checkmark} Tests certain prompt configurations like self-descriptive vs. role prompts & 
\textcolor{red}{\texttimes} Single-turn questionnaire administration per prompt configuration & 
\textcolor{red}{\texttimes} No demographic context; uses generic personas like ``high X trait'' \\
\midrule

\cite{bhandari2025evaluating} & 
\textcolor{red}{\texttimes} Reports trait/dimension-level results; does not focus on per-item stats & 
\textcolor{red}{\texttimes} & 
\textcolor{red}{\texttimes} Validation is mainly internal (mitigates training contamination and prompt biases; no direct human comparison) & 
\textcolor{red}{\texttimes} Does not test bias or safety behaviors explicitly & 
\textcolor{red}{\texttimes} Uses established questionnaires with automation & 
\textcolor{red}{\texttimes} Models answered as themselves; no persona roles given & 
$\sim$ Limited: sequential Q\&A for 40+ items, but not an interactive dialogue scenario & 
\textcolor{red}{\texttimes} No demographic grounding; generic usage across scenarios \\
\midrule

\cite{jiang2023personallm} & 
\textcolor{red}{\texttimes} Focus on overall test scores and textual patterns rather than item stats & 
\textcolor{red}{\texttimes} & 
\textcolor{green!70!black}{\checkmark} Compares model's persona-consistent writing to human texts; uses human judges to verify perceived personality & 
\textcolor{red}{\texttimes} Does not specifically measure harmful outputs or biases & 
\textcolor{green!70!black}{\checkmark} Human evaluators rated the personality of LLM-generated essays & 
\textcolor{green!70!black}{\checkmark} ``Role-play'' prompts assign each LLM a Big Five personality profile to enact & 
\textcolor{red}{\texttimes} The tests and writing tasks are one-shot or sequential Q\&A, not free-form dialogue & 
\textcolor{red}{\texttimes} Personas defined only by trait profile, not real demographics \\
\midrule

\cite{problem_agree_both} & 
\textcolor{green!70!black}{\checkmark} Examines item-level inconsistencies, e.g., responses to reversed items & 
\textcolor{red}{\texttimes} & 
\textcolor{red}{\texttimes} No external benchmark; compares LLM vs human response patterns and across prompt variations internally & 
\textcolor{red}{\texttimes} Focus is on validity of trait measurement itself; does note sycophantic agreement as a safety concern & 
\textcolor{red}{\texttimes} & 
\textcolor{green!70!black}{\checkmark} Experiments include prompts to ``simulate particular personality types'' & 
\textcolor{red}{\texttimes} Questionnaire administration was turn-based Q\&A, not evaluated as interactive dialogue & 
\textcolor{red}{\texttimes} No demographic context, just trait simulation prompts \\
\midrule

\cite{li2024quantifying} & 
\textcolor{green!70!black}{\checkmark} Multiple-item assessments per dimension; notes discrepancies between self-assessments and scenario-based responses & 
\textcolor{red}{\texttimes} Multi-dimensio-nal but no explicit IRT modeling mentioned & 
\textcolor{green!70!black}{\checkmark} Validates results across 13 diverse datasets; compares LLM self-reports to behavioral outcomes in scenarios & 
$\sim$ Limited: includes ``values'' and ``motivation'' dimensions, theory-of-mind tasks (covering ethical/moral reasoning) but not focused on toxicity or bias metrics specifically & 
\textcolor{red}{\texttimes} Uses curated benchmark data; no new human annotation described & 
\textcolor{red}{\texttimes} Evaluates base model behavior on each psych dimension without persona prompts & 
\textcolor{red}{\texttimes} Evaluations are single-turn question or task responses per dataset & 
\textcolor{green!70!black}{\checkmark} Benchmark tasks span various real-world contexts (personality quizzes, moral dilemmas, emotional scenarios, etc.) \\
\bottomrule
\end{tabular}
\vspace{0.5em}
\begin{minipage}{\textwidth}
\scriptsize
\end{minipage}
\end{table}

\begin{table}[H]
\centering
\caption{Systematic Comparison of Psychometric Evaluation Frameworks for LLMs - Part 2}
\label{tab:full_comparison_2}
\tiny
\setlength{\tabcolsep}{4pt}
\renewcommand{\arraystretch}{1.15}
\begin{tabular}{@{}p{1cm}p{1.5cm}p{1cm}p{1.5cm}p{1.5cm}p{1.5cm}p{1.5cm}p{1.5cm}p{1.5cm}@{}}
\toprule
\textbf{Paper} & 
\textbf{Item-Level Diagnostics} & 
\textbf{Uses MIRT} & 
\textbf{External Validation} & 
\textbf{Safety-Relevant Behavior} & 
\textbf{Human Annotation} & 
\textbf{Persona Conditioning} & 
\textbf{Multi-Turn Interaction} & 
\textbf{Demographic Grounding} \\
\midrule

\cite{lee2025llms} & 
\textcolor{green!70!black}{\checkmark} 8,000 scenario-based items with psychometric validation; measures internal consistency and refusal rates per item & 
\textcolor{red}{\texttimes} Employs factor-based trait scoring but not explicitly an IRT model & 
$\sim$ Limited: validates against human-established metrics of test quality (content validity, reliability) and compares results to known training data influences; no direct comparison to human respondents & 
\textcolor{green!70!black}{\checkmark} Evaluates Dark Triad traits and notes model reluctance to fully exhibit extreme psychopathy or low conscientiousness; tracks refusal rate as a safety/alignment indicator & 
\textcolor{red}{\texttimes} Dataset derived from human psych theory and knowledge graph, but model evaluation is automated & 
\textcolor{green!70!black}{\checkmark} Tests prompted ``persona elicitation'' and finds prompting often fails to induce extreme undesirable traits & 
\textcolor{red}{\texttimes} Questions are answered in single-turn multiple-choice format & 
\textcolor{green!70!black}{\checkmark} Questions incorporate realistic situational contexts via ATOMIC-10X knowledge graph, though not tied to specific demographic profiles \\
\midrule

\cite{problem_variance_1} & 
\textcolor{red}{\texttimes} Analyzes overall trait distribution in generated text vs. human text & 
\textcolor{red}{\texttimes} & 
\textcolor{green!70!black}{\checkmark} Compares AI-generated text to human-authored text across multiple domains to identify trait variance differences & 
\textcolor{red}{\texttimes} Does not directly test biases or factuality; focuses on diversity of expressed personas as an indirect quality measure & 
\textcolor{red}{\texttimes} Uses existing text datasets and automated personality inference tools & 
\textcolor{red}{\texttimes} Examines unconditioned outputs; not about prompted personas & 
\textcolor{red}{\texttimes} Evaluates static text outputs & 
\textcolor{red}{\texttimes} Compares across domains of text, but not explicitly grounded in demographic context \\
\midrule

\cite{park2024generative} & 
\textcolor{green!70!black}{\checkmark} Compares AI agent vs. human on each survey question and experimental task outcome & 
\textcolor{red}{\texttimes} & 
\textcolor{green!70!black}{\checkmark} Extensive: compares simulated agents' answers directly to real human data from interviews, surveys, and experiments & 
\textcolor{green!70!black}{\checkmark} Evaluates accuracy biases across racial/ideological groups and reduces them; tests if agents follow real human moral/social survey norms & 
\textcolor{green!70!black}{\checkmark} Built on 1,052 real humans' interview transcripts and survey responses as ground truth & 
\textcolor{green!70!black}{\checkmark} Each AI agent is imbued with a real person's persona gleaned from interviews & 
\textcolor{red}{\texttimes} Agents answer survey questions or make decisions, rather than free-form multi-turn dialogue in this evaluation & 
\textcolor{green!70!black}{\checkmark} Uses rich personal biographical context from human interviews; analyzes differences across demographic/ideological groups \\
\bottomrule
\end{tabular}

\vspace{0.5em}
\begin{minipage}{\textwidth}
\scriptsize
\textbf{Legend:} \textcolor{green!70!black}{\checkmark} = Yes/Present; \textcolor{red}{\texttimes} = No/Absent; $\sim$ = Limited/Partial\\[0.5em]
\end{minipage}
\end{table}

\section{Other Related works}
\label{app:related_works}
\paragraph{More concerns on psychometrics for AI}
Empirical studies have highlighted related limitations. LLMs frequently exhibit reduced variance across personality factors \citep{problem_variance_1, problem_variance_2}, may agree with both positive and negative versions of a statement due to sycophancy effects \citep{problem_agree_both}, and often reproduce statistical regularities from training data rather than meaningful behavioral variation \citep{problem_fidelity_2, problem_fidelity_1}. Similar concerns have been observed in simulated opinion surveys \citep{problem_fidelity_2, problem_fidelity_3} and qualitative research settings where LLMs are treated as research participants \citep{problem_fidelity_4, problem_research_participants_1, problem_research_participants_2}. These findings raise questions about whether traditional psychometric tools can meaningfully capture LLM behavior and motivate our development of more behaviorally and domain grounded evaluation frameworks. We do not claim to fully solve such concerns in the field here, but hope that our focus on structured measurement across key persona attributes and scenario types, semantically and psychologically rich personas, quanitifying base vs persona conditioned model effects and measurement against external AI behavioral benchmarks all add further rigor in these endeavors.

\paragraph{More background on SJTs}
Studies administering established SJTs to LLMs \cite{mittelstadt2024large, walsh2025using} have found that their performance often matches or surpasses that of humans. These findings suggest that, through training, LLMs have developed commonsense ethics that allow them to effectively navigate complex situations. In applied research, SJT performance has been shown to correlate with the personality traits of \textcolor{hexA}{Agreeableness} and \textcolor{hexC}{Conscientiousness} \cite{cabrera2001situational,de2017scoring}. Within policing contexts, SJTs are valued for their capacity to assess law enforcement-specific tacit-knowledge and decision-making in ambiguous, high-stakes scenarios 
\cite{taylor2013police}. 
Rather than evaluating agreement with self-descriptive statements or self-reported tendencies - which need not reflect actual decision-making behavior - each scenario presents multiple plausible courses of action, with each response option aligned to one of the six HEXACO traits. By associating each answer choice with a \hexaco\; trait, we use the SJT as a lens into the persona’s \textit{behaviors}. An \textcolor{hexH}{\textcolor{hexH}{Honesty–Humility}}-aligned persona might choose the option emphasizing fairness, whereas an \textcolor{hexX}{eXtraversion}-aligned persona might choose the option favoring social boldness. This use of SJTs allows us to measure trait-consistent behavior in AI, and is thereby more aligned with traditional AI benchmark design.

\paragraph{LLM persona datasets and human simulation.}

Recent work shows LLMs can sustain consistent personas over extended interactions \citep{park2023generative} and at scale (1{,}000 simulated individuals) \citep{park2024generative}, though without formal psychometric validation. Critics argue that LLMs cannot substitute for human participants and do not truly simulate human psychology \citep{problem_research_participants_2,problem_reword}. Prompted personas reveal further limits: “you are [HEXACO trait]” prompts elicit trait consistent language, but effects are \emph{prompt induced rather than grounded in persona} \citep{persona_llm}. Demographic conditioning has been explored, quantifying persona effects \citep{personas_quantifying} and aligning synthetic profiles to real world distributions \citep{personas_nemotron_personas}, but often without richer textual descriptions. TRAIT provides a psychometrically informed test set \citep{lee2025llms}, yet how traits shape LLM responses beyond self-report questions remains open.

\paragraph{Applying psychometrics to AI.}
Benchmarks of value orientation and moral reasoning show models approximate human judgments yet remain sensitive to phrasing and framing \cite{valuebench}. The application of personality inventories to LLMs has previously produced mixed results. \citet{lu2023illuminating} and \citet{pellert2024ai} applied Big Five and \hexaco\;  assessments to various models, finding interpretable trait patterns but also sensitivity to prompt structure and limited variance in responses.  \citet{serapio2023personality} applied personality traits to LLMs and found them to be reliable only for certain large models. \citet{mittelstadt2024large} suggests LLMs can perform strongly on social SJTs but often without explicit links to underlying traits. Recent methodological advances have shown promise: \citet{wang2025personality} find that structured personality interviews elicit richer simulated human responses.
\citet{krumm2024creating} demonstrated that ChatGPT can rapidly generate situational judgment tests, though their psychometric validation remains limited. \citet{li2024automatic} developed methods for automatic personality SJT item generation, while \citet{liu2025leveraging} explored using LLMs for item evaluation, finding that synthetic responses produce responses similar to human answers for psychometric analysis.

\section{Persona Design Methodology}
\label{sec:handmade_persona_creation}

The design of the Large Language Model (LLM) personas followed a multi-step process by one of the authors with training in psychology. These were used as validating exploration and interaction with an LLM via the console, until we were satisfied in heuristic quality of created personas. We document this first here as a historic record of our work process, inspiring the fully automated persona construction described in Section \ref{sec:automatic_persona_creation}

\begin{enumerate}
    \item Identify prominent memoirs authored by police officers or other members of law enforcement.
    \item Select a de-identified psychological report to serve as a structural reference for integrating the memoir content. The chosen profile includes:
    \begin{itemize}
        \item Sections derived from an intake interview.
        \item Data and evaluations from 15 psychometric tests.
        \item A summary with diagnostic impressions and clinical recommendations.
    \end{itemize}
    \item Provide GPT-4.1 with the memoirs, the de-identified report, and the following prompt:
    
    \begin{promptbox}
    Please create a comprehensive psychological report modeled after the format and style of the example report in the document \emph{Comprehensive Report\_E.J.}. The report should be based on the individual described in the attached memoir. The goal is to generate a complete psychological profile that will serve as the foundation for a police officer persona in LLM psychometric testing. Where necessary, extrapolate and infer responses based on the memoir content. Include psychometric tests from the example report, and incorporate additional measures such as an MMPI-3 evaluation and a Situational Judgment Test evaluation.
    \end{promptbox}
    \item Review all profiles generated for accuracy and alignment with psychometric principles.
    \item Create condensed LLM personas by reducing the first six sections of each profile to one sentence each, while retaining the entire ``Behavioral Observations'' section and the complete summary.
\end{enumerate}

\section{Case Studies}
\label{app:case_studies}

This section illustrates how structured personas and situational judgment tests (SJTs) jointly capture trait-consistent behavioral variation in large language models. We contrast two archetypal law enforcement personas: a command-oriented “Tough Cop” and a relational “Reciprocator.” These personas demonstrate (i) coherence between persona design, \hexaco\ trait profiles, and SJT behavior, and (ii) how differing trait configurations shape defensible but distinct professional judgments.

\subsection*{ \includegraphics[height=1.5em]{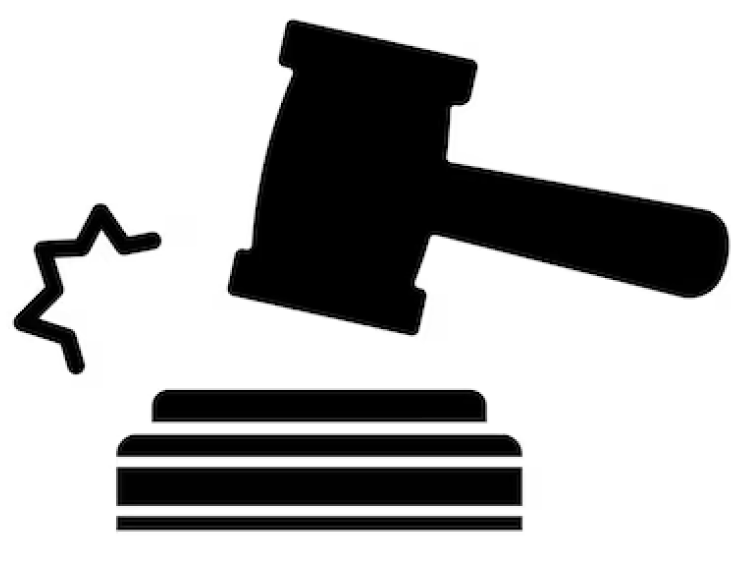} ``The Tough Cop''}

\noindent \textbf{Persona and predicted trait structure.} 
Officer Hung Wong represents an authoritarian, hierarchy-oriented archetype characterized by procedural discipline, emotional restraint, and strong commitment to order. Expert review of the persona judged it ecologically realistic and predicted a profile dominated by \textcolor{hexC}{Conscientiousness} and \textcolor{hexH}{Honesty–Humility}, with comparatively low \textcolor{hexE}{Emotionality}. Qualitative indicators, including systematic reasoning, rigid protocol adherence, and guarded affect, supported this expectation. Refer to Table \ref{tab:persona-example-2} for a full description.

\noindent \textbf{\hexaco\ – SJT alignment.} Quantitative results confirm strong construct coherence. Officer Wong exhibits exceptionally high HEXACO Conscientiousness (+5.38) and Honesty–Humility (+3.84), which together account for the majority of his SJT responses (40.0\% and 38.3\%, respectively) (Table \ref{tab:wong_hexaco}). Low \textcolor{hexE}{Emotionality} and \textcolor{hexX}{Extraversion} are likewise reflected in minimal selection of affective or socially assertive response options. Across scenarios, Wong consistently prioritizes rule-based evaluation, documentation, and risk containment. The close correspondence between latent trait scores and observed SJT behavior provides evidence that the SJTs reliably externalize the intended personality structure rather than eliciting generic “correct” answers.

\begin{table*}[ht]
\centering
\begin{tabular}{l c c c c}
\toprule
\textbf{Trait} &
\textbf{\makecell{\hexaco\\ Z-score}} &
\textbf{\makecell{Relative\\level}} &
\textbf{SJTs (\%)} &
\textbf{Alignment} \\
\midrule
\textcolor{hexC}{Conscientiousness}           & +5.38 & Exceptionally High & 40.0 & Strong alignment \\
\textcolor{hexH}{Honesty–Humility}            & +3.84 & Exceptionally High & 38.3 & Strong alignment \\
\textcolor{hexA}{Agreeableness}               & –0.78 & Slightly Low       &  6.7 & Strong alignment \\
\textcolor{hexX}{eXtraversion}                & –1.37 & Low                &  5.0 & Strong alignment \\
\textcolor{hexE}{Emotionality}                & –0.59 & Slightly Low       &  3.3 & Alignment \\
\textcolor{hexO}{Openness to Experience}      &  0.00 & Average            &  6.7 & Moderate alignment \\
\bottomrule
\end{tabular}
\caption{\hexaco\ Trait and SJT Alignment for Officer Wong}
\label{tab:wong_hexaco}
\end{table*}

\subsection*{\includegraphics[height=1.5em]{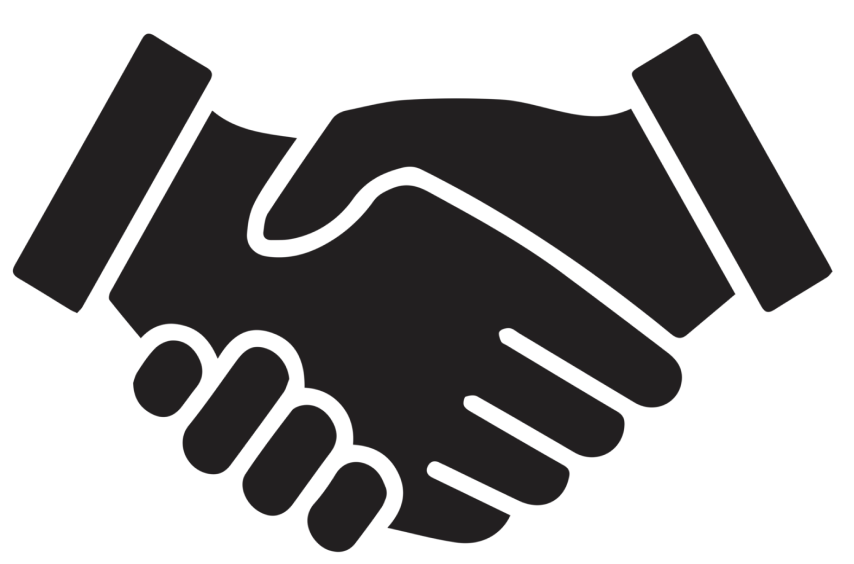} ``The Reciprocator'' }

\noindent \textbf{Persona overview.} Officer Eleanor Hagedorn exemplifies a rapport-oriented archetype marked by empathy, reflective judgment, and integrative problem solving. Expert assessment identified \textcolor{hexH}{Honesty–Humility}, \textcolor{hexA}{Agreeableness}, and \textcolor{hexO}{Openness to Experience} as her dominant traits, emphasizing relational trust, perspective-taking, and holistic interpretation of complex situations. Refer to Table \ref{tab:persona-example-4} for a full description.

\vspace{0.3cm}

\noindent \textbf{\hexaco\ – SJT alignment.} Officer Hagedorn’s scores of \textcolor{hexH}{Honesty--Humility} (+3.03), \textcolor{hexA}{Agreeableness} (+1.38), and \textcolor{hexO}{Openness to Experience} (+1.88) are strongly reflected in her SJT behavior, accounting for approximately three quarters of her response distribution (Table \ref{tab:hagedorn_hexaco}). Her selections favor prosocial engagement, mediation, and contextual sensitivity. In contrast to Officer Wong, traits such as \textcolor{hexC}{Conscientiousness} and \textcolor{hexE}{Emotionality}, while elevated in the \hexaco\ inventory, are less prominent in SJT outcomes. This pattern reflects differential behavioral salience rather than psychometric failure. When multiple traits are high, SJTs tend to surface those most motivationally directive in situational decision-making. Alignment between SJT results and expert qualitative judgment supports cross-method validity.
  
\noindent \textbf{Comparative analysis of trait expression across personas.}
Direct comparison of the two personas highlights how identical scenarios yield systematically different responses that are still professionally acceptable. A representative \hexaco\ item assessing empathic sensitivity mirrors their divergent \textcolor{hexE}{Emotionality} scores: Officer Hagedorn strongly endorses emotional attunement, whereas Officer Wong disagrees. In an ambiguous, high-pressure SJT requiring prioritization under conflicting guidance, Hagedorn selects the \textcolor{hexA}{Agreeableness} strategy emphasizing compassionate engagement and equity of attention, while Wong selects the \textcolor{hexC}{Conscientiousness} strategy centered on procedural review and structured risk assessment. As neither response is inherently superior, this indicates that the framework captures how decisions are made as a function of personality structure.
These case studies demonstrate that structured personas combined with trait-purified SJTs produce coherent mappings between latent psychometric traits and observable behavior. The alignment across narrative persona design, \hexaco\ inventories, expert judgment, and SJT responses provides convergent evidence for the construct validity of the proposed evaluation framework, while preserving meaningful heterogeneity in professional reasoning styles.

\begin{table*}[ht]
\centering
\begin{tabular}{l c c c c}
\toprule
\textbf{Trait} &
\textbf{\makecell{\hexaco\\Z-score}} &
\textbf{\makecell{Relative\\level}} &
\textbf{SJTs (\%)} &
\textbf{Alignment} \\
\midrule
\textcolor{hexH}{Honesty--Humility}      & +3.03 & Exceptionally High & 30.0 & Strong alignment \\
\textcolor{hexA}{Agreeableness}          & +1.38 & Very High          & 26.7 & Strong alignment \\
\textcolor{hexO}{Openness to Experience} & +1.88 & Very High          & 18.3 & Strong alignment \\
\textcolor{hexX}{Extraversion}           & +1.13 & High               & 11.7 & Moderate alignment \\
\textcolor{hexC}{Conscientiousness}      & +2.38 & Exceptionally High & 8.3  & Weak alignment \\
\textcolor{hexE}{Emotionality}           & +1.09 & High               & 5.0  & Weak alignment \\
\bottomrule
\end{tabular}
\caption{\hexaco\ Trait and SJT Alignment for Officer Hagedorn}
\label{tab:hagedorn_hexaco}
\end{table*}

\section{Data and response generation framework details}\label{app:engineering}

\noindent \textbf{Modularity and extensibility.} Our framework is designed to be robust, reproducible, and easily extensible to new domains and scenarios. A key design principle is modularity—the ability to flexibly combine and reconfigure experimental components. The framework supports substitution of base models, persona definitions, SJT scenarios, and \hexaco\ mappings, enabling controlled studies of behavioral variation in LLMs.

Persona seeds and SJT templates are specified through structured YAML configuration files. Given these seeds, the framework can automatically generate new personas and scenario banks for a domain without modifying the underlying codebase, making the approach easily portable and scalable through configuration alone.

We use vLLM’s structured generation framework \cite{kwon2023efficient} to support high-throughput and reproducible evaluation with full CLI-level control over generation settings. Parameters such as option ordering, paraphrasing, shuffling, sampling strategy, and model selection are specified through configuration. This allows responses for new models to be generated against the same datasets simply by changing the vLLM launch configuration.

To generate the personas, we used \emph{GPT-4.1} \citep{openai2024gpt4technicalreport} with \textbf{temperature} $=2.0$ and \textbf{top-$p$} $=0.98$, parsed via structured outputs to the schema.

For the SJTs, we used GPT 4.1 with hyperparameters temperature=1.5, top\_p=0.95, presence\_penalty=0.4 and frequency\_penalty=0.3 for creation of SJT and correction for trait bleed.

\section{Hand Designed Persona Schema}
\label{app:persona_schema}
\subsection*{Sections}
{\small
\begin{tabular}{|p{7cm}|p{9cm}|}
\hline
\textbf{Section} & \textbf{Fields / Items} \\
\hline
Demographic Fields &
Name, Date of Birth, Age, Location \\
\hline
Behavioral and Psychological Descriptors &
Appearance, Behavior, Mood/Affect, Speech, Thought Content, Insight/Judgment, Cognition \\
\hline
Life History Segments &
Medical/Developmental History, Family History, Educational/Vocational History \\
\hline
Functional Assessments &
Emotional/Behavioral Functioning, Social Functioning \\
\hline
Summary & Summary of Psychological Profile \\
\hline
\end{tabular}
}

\subsection*{Demographic Fields}
{\small
\begin{tabular}{|p{7cm}|p{9cm}|}
\hline
\textbf{Field} & \textbf{Content Style} \\
\hline
Name & Human-like; use multicultural names; randomly generate using a full-name generator. Include optional ``Preferred Name'' in parentheses. \\
\hline
Age & Range: 21--70; ensure logical consistency with profile (e.g., early career vs.\ veteran). \\
\hline
Location & U.S.\ cities or fictional equivalents; must reflect the cultural or professional environment of the persona (e.g., Baltimore for homicide cop). \\
\hline
\end{tabular}
}

\subsection*{Behavioral and Psychological Descriptors}
\noindent Each field below must contain rich natural-language text (30--100 words), simulating clinical-style notes. Use psychological and sociological realism (avoid caricature). Include mild contradictions or nuance for realism.

{\small
\begin{tabular}{|p{7cm}|p{9cm}|}
\hline
\textbf{Field} & \textbf{Content Style} \\
\hline
Appearance & Observational, sensory; e.g., ``Neat but tired-looking, often wears fitted blazer, carries visible tension.'' \\
\hline
Behavior & Behavioral cues; highlight posture, interaction style, responsiveness. \\
\hline
Mood/Affect & Tone modulation; include controlled, anxious, upbeat, irritable, etc. \\
\hline
Speech & Register and rhythm; formal/informal, slang use, coherence. \\
\hline
Thought Content & Internal reflection; logicality, obsession, themes (e.g., justice, betrayal). \\
\hline
Insight/Judgment & Clinical phrasing; e.g., ``Demonstrates situational insight, though blind spots persist in personal relationships.'' \\
\hline
Cognition & Memory, abstraction, coherence; based on MoCA-style logic and LLM capabilities. \\
\hline
\end{tabular}
}

\subsection*{Life History Segments}
\noindent These reflect psychosocial context. Use 100--150 tokens per field with narrative cohesion.

\vspace{0.15cm}
{\small
\begin{tabular}{|p{7cm}|p{9cm}|}
\hline
\textbf{Field} & \textbf{Content Style} \\
\hline
Medical or Developmental History & Include common chronic or stress-related issues. Exclude major psychiatric illness unless purposefully included. \\
\hline
Family History & Include 1--2 generational details, alcohol/substance patterns, notable relational dynamics. \\
\hline
Educational or  Vocational History & Include level of education, job trajectory, training milestones, institutional affiliations. \\
\hline
\end{tabular}
}

\subsection*{Functional Assessments}
\noindent Emphasize social and emotional coping styles, behavioral norms, and situational pressures. These should align with the emotional tone in the cognitive and personality testing data.

\vspace{0.15cm}
{\small
\begin{tabular}{|p{7cm}|p{9cm}|}
\hline
\textbf{Field} & \textbf{Content Style} \\
\hline
Emotional/ Behavioral Functioning & How the persona processes stress, trauma, anger, etc. Mention coping mechanisms. \\
\hline
Social Functioning & Relationship style, trust levels, group affiliation, isolation or connection patterns. \\
\hline
\end{tabular}
}

\subsection*{Summary of Psychological Profile}
This is an integrative paragraph using clinical summarization style and lists diagnostic impressions, resilience factors, risks (burnout, PTSD, moral injury), and brief prognosis. Ties together medical, cognitive, emotional, and social features.

\section{Situational Judgment Tests (SJT) Schema}
\label{sec:sjt_schema}

We give here the seed schema we used for our automated SJT generation.

\begin{table}[H]
\label{tab:sjt_schema}
\centering
\begin{tabularx}{\columnwidth}{|X|c|}
\hline
\textbf{Section} & \textbf{Approx.\ Tokens} \\
\hline
Question & 100--120 \\
\hline
\textcolor{hexH}{Honesty-Humility} Option & \textasciitilde40--60 \\
\hline
\textcolor{hexE}{Emotionality} Option & \textasciitilde40--60 \\
\hline
\textcolor{hexX}{eXtraversion} Option & \textasciitilde40--60 \\
\hline
\textcolor{hexA}{Agreeableness} Option & \textasciitilde40--60 \\
\hline
\textcolor{hexC}{Conscientiousness} Option & \textasciitilde40--60 \\
\hline
\textcolor{hexO}{Openness} Option & \textasciitilde40--60 \\
\hline
Trait Bleed Evaluation & \textasciitilde500--700 \\
\hline
\textbf{Total (per SJT)} & \textbf{\textasciitilde1,100--1,500 tokens} \\
\hline
\end{tabularx}
\end{table}

\section{Persona Generation Seeds}\label{app:persona_seeds}

We lay out the major seed categories now.

\noindent \textbf{Memoir Seeds} \\
\begin{table}[H]
\small
\centering
\caption{Selected Police Memoirs (20 of 100)}
\label{tab:memoirs}
\begin{tabularx}{\columnwidth}{Y Y l}
\toprule
\textbf{Title} & \textbf{Author} & \textbf{Year} \\
\midrule
Forced Out & Kevin Maxwell & 2020 \\
Drugs, Guns \& Lies & Keith Banks & 2018 \\
Gun to the Head & Keith Banks & 2020 \\
One Step Behind Mandela & Rory Steyn & 2001 \\
The Turnaround & William J. Bratton & 1998 \\
The Profession & William J. Bratton & 2021 \\
Blue Blood & Edward Conlon & 2004 \\
The Job & Steve Osborne & 2015 \\
Street Justice & Steve Osborne & 2016 \\
Busting the Mob & Jack Maple & 1999 \\
The Crime Fighter & Jack Maple & 2000 \\
Breaking Ranks & Norm Stamper & 2005 \\
Vigilance & Raymond W. Kelly & 2015 \\
From Jailer to Jailed & Bernard B. Kerik & 2015 \\
Cop in the Hood & Peter Moskos & 2008 \\
Police Craft & Adam Plantinga & 2018 \\
Ghettoside & Jill Leovy & 2015 \\
We Own This City & Justin Fenton & 2021 \\
Homicide & David Simon & 1991 \\
The Onion Field & Joseph Wambaugh & 1973 \\
\multicolumn{3}{c}{\ldots} \\
\bottomrule
\end{tabularx}
\end{table}

A list of 100 police memoirs was generated by GPT-4.1 in Table \ref{tab:memoirs} to provide narrative priors, occupational context, and scenario scaffolds for downstream persona construction.

\noindent \textbf{Appearance Categories}

\newcolumntype{Y}{>{\raggedright\arraybackslash}X}

\begin{table}[H]
\small
\centering
\renewcommand{\arraystretch}{0.9}
\caption{Selected Top-Level Appearance Categories}
\label{tab:appearances}
\begin{tabularx}{\columnwidth}{|Y|Y|}
\hline
\textbf{Category} & \textbf{Scope and Canonical Coverage} \\
\hline
Uniform / Official & Duty and ceremonial uniforms; rank/insignia placement; loadout layout (e.g., body camera, utility belt); parade and formal variants. \\
\hline
Plainclothes / Casual & Civilian attire with concealed or overt identifiers (e.g., badge chain, ankle holster); off-duty layering patterns. \\
\hline
Weathered / Field Gear & Outdoor or wilderness adaptations; precipitation and terrain equipment; protective accessories for extended field operations. \\
\hline
Formal / Elite & Tailored formalwear with insignia; medals and ribbons; ceremonial gloves; elite dress standards and grooming. \\
\hline
Rough / Unkempt & Indicators of wear and tear; scuffs, frays, and asymmetry; comfort-over-form presentation consistent with demanding shifts. \\
\hline
Grooming / Personal Style & Hair and facial-hair conventions; eyewear and accessories; polish and shine standards; minimalist timepieces. \\
\hline
Body Type / Build & Structured silhouette templates (e.g., tall/muscular; compact/agile; petite/nimble) representing morphological diversity. \\
\hline
Iconic Accessories & Functional and symbolic items (e.g., cuffs, flashlight, holster, body camera, evidence bags, specialty patches). \\
\hline
Climate / Regional Adaptation & Thermal, arid, tropical, and alpine variants; fire and wildland overlays; UV, mesh, and layered uniforms. \\
\hline
Off-Duty / Hybrid & Badged casual wear, badge clips, light vests, trainers or loafers; professional–civilian blending for transitional contexts. \\
\hline
\end{tabularx}
\end{table}

We defined a taxonomy of appearance categories in Table \ref{tab:appearances} and curated representative exemplar sets for each category to enable granular conditioning in generation and evaluation. 

\subsection*{Archetypes}
Eight archetypes represent higher-order values and common patterns of thinking in the law enforcement domain. See Table \ref{tab:archetypes}.

\newcolumntype{Y}{>{\raggedright\arraybackslash}X}

\begin{table}[H]
\small
\centering
\renewcommand{\arraystretch}{0.9}
\caption{Canonical Archetypes of Police Officer Personas}
\label{tab:archetypes}
\begin{tabularx}{\columnwidth}{|Y|Y|Y|}
\hline
\textbf{Name} & \textbf{Core Trait} & \textbf{Primary Focus} \\
\hline
The Professional (Service-Oriented Officer) & Community service; procedural justice & De-escalation, fairness, empathy, adherence to policy \\
\hline
The Enforcer (Crime-Fighter) & Control of crime/disorder & Arrests, assertive authority, low tolerance for infractions \\
\hline
The Reciprocator (Nice Cop) & Harmony-seeking; rapport-building & Mediation, helping others, peacekeeping \\
\hline
The Avoider (Lazy Officer) & Low motivation & Minimal engagement, risk avoidance, low proactivity \\
\hline
The Avoider (Unconfident Officer) & Hesitation; self-doubt & Disengagement under pressure; overcautious posture \\
\hline
The Tough Cop (Authoritarian) & Command-and-control & Hierarchy, obedience, strict compliance \\
\hline
The Problem Solver / Investigator & Analytical; detail-oriented & Evidence synthesis, airtight documentation, puzzle resolution \\
\hline
The Problem Solver / Public Servant & Systems-minded; long-term orientation & Mediation, root-cause interventions, referrals to services \\
\hline
\end{tabularx}
\end{table}

\noindent \textbf{Behavior Categories}
We specified behavior categories in Table \ref{tab:behaviors} capturing recurrent interactional and situational patterns. Each category is associated with a curated set of micro-behavior exemplars to support fine-grained conditioning and comparative analysis.

\begin{table}[H]
\small
\centering
\caption{Behavior Categories (High-Level Definitions)}
\label{tab:behaviors}
\begin{tabularx}{\columnwidth}{|X|X|}
\hline
\textbf{Category} & \textbf{Definition} \\
\hline
Vigilant / Scanning & Continuous threat assessment and sightline management; environment mapping and movement tracking. \\
\hline
Relaxed / Measured & Deliberate pacing, calm affect, and de-escalatory stance in dialogue and posture. \\
\hline
Nervous / Restless & Elevated motor restlessness and micro-tension markers indicative of unstable focus. \\
\hline
Expressive / Engaged & Emphatic gesture use, high reciprocity, and rapport-forward communication. \\
\hline
Ritualistic / Habits & Stable pre-/post-incident routines for equipment, documentation, and readiness. \\
\hline
Authoritative / Directive & Command presence; precise, unambiguous instruction delivery and turn-taking control. \\
\hline
Withdrawn / Guarded & Reduced verbal output and constrained kinesics to limit exposure and signal reserve. \\
\hline
Performative / Showmanship & Audience-aware delivery with dramatic emphasis and high-energy framing. \\
\hline
Impulsive / Erratic & Abrupt topic/task shifts, premature action initiation, and inconsistent pacing. \\
\hline
Analytical / Focused & Evidence-first reasoning, meticulous note-taking, hypothesis formation and testing. \\
\hline
Social / Bonding & Pro-social team maintenance behaviors and community-relational micro-actions. \\
\hline
Procedural / By-the-Book & Protocol-referenced execution, checklist usage, and documentation fidelity. \\
\hline
Cautious / Defensive & Risk-managed distance, protective postures, and pre-movement scanning. \\
\hline
Confrontational / Challenging & High-assertion stance with direct challenges and pressure-forward rhetoric. \\
\hline
Supportive / Empathetic & Active listening, affect validation, and prosocial reassurance behaviors. \\
\hline
Cynical / Detached & Flattened affect, skeptical stance, and de-emphasized engagement. \\
\hline
Storytelling / Narrative & Structured narrative exposition, pacing modulation, and illustrative anecdote use. \\
\hline
\end{tabularx}
\end{table}

\subsection*{Context Summary}
The memoir-derived corpus, appearance taxonomy with curated exemplar sets, behavior category library with micro-behavioral primitives, and the eight archetypes jointly constitute a structured seed space for generating diverse, realistic police-officer personas. This framework supports controllable conditioning, systematic ablations, and robust evaluation across persona--by--scenario strata.

\section{Bootstrapping Persona from Real World Demographics Data}
\label{sec:persona_demographics}
We obtained enterprise access to use the licensed SDG-PGMs \citep{pgm_corneil_meyer_2025_nemotron_personas_summit} framework based on real-world data. SDG-PGMs is a Python framework for building Probabilistic Graphical Models (PGMs) to generate synthetic data. The framework extends pgmpy\citep{pgmpy} to support cascaded PGMs with arbitrary post-processing steps. Specifically, we used PGMs created from US Census data and the names data of \citet{pgm_rosenman}. ZIP codes with corresponding city and state generated to ground later variables in geography; first, middle and last names are generated from sex and ethnic background statistics at the zipcode level; education is sampled based on zipcode-level statistics, followed by occupation including influence of education and age.

Although the PGM we derived the police data from was proprietary, however we are able to release the full generated synthetic dataset as part of the final release of our paper.

\section{Persona Statistics and Evaluation}
\label{sec:persona_stats}

\subsubsection{Gender}

\begin{table}[H]
\centering
\caption{Distribution of Gender (\%)}
\begin{tabularx}{\columnwidth}{|X|c|}
\hline
\textbf{Gender} & \textbf{Percentage (\%)} \\
\hline
Male & 86.082 \\
\hline
Female & 13.918 \\
\hline
\end{tabularx}
\end{table}

\subsubsection{Age}

\begin{table}[H]
\centering
\caption{Age Group Definitions}
\begin{tabularx}{\columnwidth}{|X|c|}
\hline
\textbf{Age Range (in Yrs.)} & \textbf{Age Group} \\
\hline
<18 & Juvenile \\
\hline
18-25 & Young Adult \\
\hline
25-40 & Adult \\
\hline
40-60 & Middle Aged \\
\hline
>60 & Senior \\
\hline
\end{tabularx}
\end{table}

\begin{table}[H]
\centering
\caption{Distribution of Age Groups (\%)}
\begin{tabularx}{\columnwidth}{|X|c|}
\hline
\textbf{Age Group} & \textbf{Percentage (\%)} \\
\hline
Adult & 39.376 \\
\hline
Middle-aged & 39.082 \\
\hline
Senior & 12.882 \\
\hline
Young Adult & 8.659 \\
\hline
\end{tabularx}
\end{table}

\subsubsection{Ethnic Background}


\begin{table}[H]
\centering
\caption{Distribution of Ethnic Background (\%)}
\begin{tabular}{|l c|l c|}
\hline
\textbf{Ethnic Background} & \textbf{\%} & 
\textbf{Ethnic Background} & \textbf{\%} \\
\hline
White & 61.247 & Puerto Rican & 2.376 \\
Black & 12.624 & East Asian & 2.341 \\
Mexican & 10.647 & Southeast Asian & 2.153 \\
Salvadoran & 0.965 & South Asian & 1.588 \\
Cuban & 0.812 & Dominican & 0.671 \\
Colombian & 0.494 & Guatemalan & 0.553 \\
Honduran & 0.400 & American Indian & 0.353 \\
Peruvian & 0.282 & Spaniard & 0.259 \\
Spanish & 0.224 & Ecuadorian & 0.224 \\
Venezuelan & 0.212 & Nicaraguan & 0.118 \\
Micronesian & 0.118 & Argentinean & 0.106 \\
Chilean & 0.082 & Polynesian & 0.059 \\
Bolivian & 0.059 & Asian Other & 0.047 \\
Costa Rican & 0.047 & Panamanian & 0.047 \\
Spanish American & 0.035 & Alaska Native & 0.024 \\
Central Asian & 0.024 & Melanesian & 0.024 \\
Other South American & 0.024 & Other Central American & 0.012 \\
\hline
\end{tabular}
\end{table}

\subsection{Persona Evaluation Rubric and Metrics}

We report the following diversity metrics we collected for the Personas in Table \ref{tab:persona_diversity_metrics}

\begin{table}[H]
\centering
\caption{Lexical and Semantic Diversity Metrics for Generated Personas}
\label{tab:persona_diversity_metrics}
\begin{tabularx}{\columnwidth}{|X|c|}
\hline
\textbf{Metric} & \textbf{Value} \\
\hline
MSTTR-100 & 0.876 \\
\hline
Compression Ratio & 0.320 \\
\hline
Yule's K & 37.556 \\
\hline
MTLD & 1.001 \\
\hline
Average Cosine Distance & 0.410 \\
\hline
\end{tabularx}
\end{table}

These are quite high, for instance the MSTTR100 is generally considered diverse for scores above 0.70. We give the rubric for LLM evaluation in Box \ref{box:persona_rubric}. The Cohen's Kappa is very noisy if both raters have constant preferences, which we encountered during persona evaluation ( with low value of \textasciitilde 0.05), but both annotators rated highly for all the rubrics. 

\begin{table}[H]
\label{tab:avg_rating_llm_human}
\centering
\caption{Mean Human and LLM Judge Ratings Across Evaluation Dimensions for 55 annotations}
\begin{tabularx}{\columnwidth}{|X|c|c|}
\hline
\textbf{Dimension} & \textbf{Human} & \textbf{LLM Judge} \\
\hline
Clarity & 4.309 & 5.000 \\
\hline
Originality & 4.600 & 4.036 \\
\hline
Coherence & 4.673 & 5.000 \\
\hline
Diversity & 4.964 & 3.509 \\
\hline
Realism & 4.273 & 5.000 \\
\hline
Psychological Depth & 4.491 & 5.000 \\
\hline
Consistency & 4.727 & 5.000 \\
\hline
Informativeness & 4.509 & 5.000 \\
\hline
Ethical Considerations & 4.855 & 5.000 \\
\hline
Demographic Fidelity & 4.655 & 5.000 \\
\hline
Overall Score & 4.509 & 4.800 \\
\hline
\end{tabularx}
\end{table}

\begin{tcolorbox}[enhanced,
  colback=gray!5,
  colframe=black!75,
  fonttitle=\bfseries,
  title={Rater Instructions},
  boxrule=0.6pt,
  width=\columnwidth
]
\label{box:persona_rubric}
\small
\textbf{Your task is to assess the quality of the dataset entry itself, not the competency or character of the persona described.}

Rate the following aspects from 0 (worst) to 5 (best):
\begin{itemize}
  \item \textbf{clarity}: Is the persona description clear and understandable?
  \item \textbf{originality}: Does the entry avoid clichés and present a unique character?
  \item \textbf{coherence}: Is the information internally consistent and logically structured?
  \item \textbf{diversity}: Does the persona contribute to a diverse set of police profiles?
  \item \textbf{realism}: Does the persona feel plausible and authentic for a police context?
  \item \textbf{psychological\_depth}: Does the entry provide meaningful insight into the persona's inner life, motivations, and psychological complexity? \emph{(Focus especially on this metric.)}
  \item \textbf{consistency}: Are details about the persona consistent throughout?
  \item \textbf{informativeness}: Does the entry provide rich, relevant information about the persona?
  \item \textbf{ethical\_considerations}: Is the entry free from harmful stereotypes or bias?
  \item \textbf{demographic\_fidelity}: Is the persona plausible for the demographic data provided (e.g., a 22 year old should not be described as retiring with decades of experience)?
  \item \textbf{overall\_score}: Your overall assessment of the dataset entry's quality.
\end{itemize}

Return only the scores in the specified schema, and include a UID string field (\texttt{UID}) for this entry. \textbf{Remember:} you are judging the quality of the dataset entry, not the police persona's job performance.
\end{tcolorbox}

\begin{tcolorbox}[enhanced,
  colback=gray!5,
  colframe=black!75,
  fonttitle=\bfseries,
  title={Persona Inter Rater Agreement Example},
  boxrule=0.6pt,
  width=\columnwidth
]
\label{box:persona_interagreement_example}
\small

\textbf{name:} Charles Gleason \\
\textbf{age:} 26 \\
\textbf{sex:} Male \\
\textbf{location:} Denver, CO \\
\textbf{ethnic background:} white \\
\textbf{marital status:} never married \\
\textbf{appearance:} Tall, slender with light blond hair, blue uniform crisply pressed, gold badge over his heart shining beneath reflective street lamps, slim body camera chest-mounted, faint silver wristband subtle under shirt sleeve, typically neutral expression. \\
\textbf{behavior:} Reserved posture; shoulders slightly rounded, occasionally avoiding direct eye contact, fingers often entwining or smoothing clothing fabric, tone soft-spoken with low volume, reacts to tension with cautious withdrawal, measured body language minimal but alert. \\
\textbf{speech:} Converses softly and evenly with measured, deliberate pace; articulate with no rush or emphatic bursts, choosing professional and unadorned language, typically maintains politeness and precision despite raised voices around him. \\
\textbf{mood affect:} Tone remains composed, restrained under stress; shifts barely perceptible but intent remains neutral with occasional faint signs of guarded frustration or distant thoughtfulness. \\
\textbf{educational vocational history:} Graduated with a bachelor's degree in business from a state university before entering law enforcement academy; focuses on building solid community connections and rule adherence influenced by administrative training and emphasis on accountability. \\
\textbf{medical developmental history:} No major health problems or hospitalizations; developed steady routines emphasizing physical fitness and endurance as recommended during academy training; slight seasonal allergies cause occasional mild congestion in colder months. \\
\textbf{family history:} Single son from a stable middle-class family with supportive parents; grew up in suburban Denver emphasizing civic responsibility; no close extended family contact but maintains respectful communication with older brother and distant cousin. \\
\textbf{thought content:} Often preoccupied with adhering precisely to protocol, ensuring no missteps that might jeopardize trust or fairness; mentally rehearses possible reactions in confrontational scenarios, pondering broader impacts of policy and individual interactions. \\
\textbf{insight judgment:} Demonstrates practical understanding of need for clear boundaries balanced with empathetic approach; aware of limits to influence in volatile situations and prioritizes own safety and adherence over impulsive action. \\
\textbf{cognition:} Capably recalls procedural steps and policy details during active events; problem solving centers on communication strategies and stepwise de-escalation rather than physical control measures. \\
emotional behavioral functioning: Manages heightened anxiety by focusing on breathing and rational evaluation; remains outwardly calm, prefers solitude to process emotion but rarely displays overt distress; adapts slowly to fast-moving circumstances requiring quick assertive responses. \\
\textbf{social functioning:} Exhibits selective social engagement; trusted by community members respecting procedural fairness, yet reluctant to forge personal connections due to internal reservations; communicates clearly but sparingly with peers, maintaining professional distance. \\
\textbf{summary of psychological profile:} Charles Gleason embodies the service-oriented officer shaped by formal education and an innate drive for fairness, navigating public duties with a quiet, restrained presence. His preference for de-escalation and procedure aligns with efforts to build legitimacy among diverse community members. However, a guarded disposition and limited emotional expressiveness occasionally distance him socially, reflected in subtle tension between a desire to connect and his habit of internalizing stress. Reliant on established protocols, he shows an acute awareness of boundaries while wrestling with minor sleep difficulties and physical hesitance that challenge hands-on confrontations. Ultimately, Charles exemplifies consistent professional engagement molded by empathy tempered with protective withdrawal to balance community service amidst unpredictable, emotionally charged encounters

\end{tcolorbox}

\section{Situational Judgment Tests (SJT) Evaluation}
\label{sec:sjt_eval}

To systematically evaluate the quality of synthetic SJTs, we employ a large language model (LLM) as an evaluator under multiple rubric-based frameworks. Since LLM-based evaluation can itself introduce systematic biases, we design two complementary rubrics that approach the problem from opposite directions. This dual-rubric design reduces discrepancy and increases robustness in the overall evaluation. To ensure validity, a subset of SJTs are manually annotated by a psychologist and a patrol officer using the same rubrics. The LLM Judge is then used to scale the evaluation to the entire dataset, and inter-rater agreement between human and LLM annotations is computed to quantify alignment and reliability. On average the agreement (Cohen's Kappa Score) between humans and the LLM judge is substantial (\textasciitilde0.63), ranging from perfect agreement (Cohen's kappa score = 1) to random chance (Cohen's kappa score = 0) for different rubric criteria.

\subsection{Rubric 1: Scenario- and Option-Centric Evaluation}
In the first rubric, the LLM is presented with the full SJT question, its answer options, the intended trait-label mapping, and the seed values used for generation. The model then evaluates the SJT along four criteria:
\begin{enumerate}
    \item \textbf{Scenario Realism and Plausibility}
    
    \textbf{Definition:} Degree to which the scenario is realistic, coherent, and appropriately grounded in the provided seed values.
    
    \textbf{Scoring:}
    
       \textbf{5 (Excellent)}: Highly plausible and contextually coherent; concise without ambiguity or unnecessary complexity.
    
      \textbf{4 (Good):} Generally plausible, but contains minor vagueness or mild inconsistencies.
    
      \textbf{3 (Adequate):} Somewhat plausible; noticeable ambiguity or complexity.
    
     \textbf{2 (Weak):} Implausible in parts, with substantial vagueness or contrived elements.
    
     \textbf{1 (Poor):} Contrived, unrealistic, or disconnected from seed values.

    \item \textbf{Trait Alignment of Options (\hexaco\;  Mapping)}

    \textbf{Definition:} Extent to which each response option uniquely expresses the intended \hexaco\;  trait without redundancy or overlap.
    
    \textbf{Scoring:}
    
    \textbf{5 (Excellent):} Clear and unique expression of the assigned trait with minimal overlap.
    
    \textbf{4 (Good):} Mostly aligned; minor ambiguity or partial overlap with another trait.

    \textbf{3 (Adequate):} Noticeable ambiguity; traits are only partially distinguishable.

    \textbf{2 (Weak):} Strong overlap with other traits; option fails to clearly convey the intended trait.

    \textbf{1 (Poor):} Misaligned or indistinguishable from other traits.

    \textbf{Additional check:} If score < 5, evaluator specifies the overlapping trait(s).

    \item \textbf{Ethical and Value Tension Representation}

    \textbf{Definition:} Presence of meaningful ethical or value-driven dilemmas (e.g., authority vs. compassion, adherence to policy vs. pragmatic shortcuts).

    \textbf{Scoring:}

    \textbf{5 (Excellent):} Strong, clear ethical/value-based tension that forces difficult trade-offs.

    \textbf{4 (Good):} Ethical considerations are present but not sharply defined.

    \textbf{3 (Adequate):} Ethical aspects exist but are peripheral rather than central.

    \textbf{2 (Weak):} Minimal ethical tension, weakly articulated.

    \textbf{1 (Poor):} No meaningful ethical or professional conflict.

    \item \textbf{Bias and Fairness Check}
    
    \textbf{Definition:} Neutral and non-stereotypical use of demographic attributes (e.g., race, gender, age) when specified by seed values.

    \textbf{Scoring:}

    \textbf{5 (Excellent):} Demographics included only as relevant context; no bias or stereotyping.

    \textbf{4 (Good):} Mostly neutral; slight risk of unnecessary demographic emphasis.

    \textbf{3 (Adequate):} Some stereotyping risk or questionable emphasis on demographics.

    \textbf{2 (Weak):} Noticeable stereotypical framing or demographic bias.

    \textbf{1 (Poor):} Strongly biased or inappropriate demographic representation.
\end{enumerate}

\subsection{Rubric 2: Reverse Inference Evaluation}

The second rubric adopts a reverse-inference perspective, designed to complement Rubric 1 and counterbalance potential biases introduced by disclosing ground-truth labels. In this setting, the LLM is presented only with the SJT question and answer options—without access to the intended trait mappings or seed values. 

The evaluation proceeds along two dimensions:

\begin{enumerate}
    \item \textbf{Seed Attribute Inference}
    
    \textbf{Task:} For each scenario, the evaluator predicts the most likely values of the underlying seed attributes (e.g., urgency, threat, ambiguity, stakeholder complexity, authority relations, ethical tension, time-of-day, and demographics).

    \textbf{Purpose:} This measures how well the generated scenario communicates its intended situational features without explicit annotation.

    \item \textbf{Trait Prediction and Justification}

    \textbf{Task:} For each answer option, the evaluator assigns it to one of the \hexaco\;  traits and provides a concise justification for its classification.

    \textbf{Purpose:} This assesses whether response options are interpretable and discriminable to an independent judge, without reliance on pre-specified trait labels.
\end{enumerate}

By withholding ground-truth mappings, Rubric 2 tests the recoverability of both scenario attributes and trait alignments. This complementary setup enables cross-validation against Rubric 1, thereby reducing evaluation bias and improving confidence in the psychometric clarity of the generated SJT pool.

\subsection{Rubric Wise Inter-Rater Agreement}
\begin{table}[H]
\centering
\caption{Rubric Wise Inter-Rater Agreement}
\begin{tabularx}{\columnwidth}{|X|c|}
\hline
\textbf{Rubric Name} & \textbf{Cohen Kappa Score} \\
\hline
Ambiguity Level & 0.085 \\
\hline
Authority Relationships & 0.310 \\
\hline
Bias Fairness & 1.00 \\
\hline
\textcolor{hexC}{Conscientiousness} Alignment & 1.000 \\
\hline
\textcolor{hexC}{Conscientiousness} Overlap & 1.000 \\
\hline
\textcolor{hexE}{Emotionality} Alignment & 1.000 \\
\hline
\textcolor{hexE}{Emotionality} Overlap & 1.000 \\
\hline
\textcolor{hexX}{eXtraversion} Alignment & 0 \\
\hline
\textcolor{hexX}{eXtraversion} Overlap & 0 \\
\hline
\textcolor{hexA}{Agreeableness} Alignment & 0.067 \\
\hline
\textcolor{hexA}{Agreeableness} Overlap & 0.030 \\
\hline
\textcolor{hexH}{Honesty Humility} Alignment & 1.000 \\
\hline
\textcolor{hexH}{Honesty Humility} Overlap & 1.000 \\
\hline
\textcolor{hexO}{Openness} Alignment & 1.000 \\
\hline
\textcolor{hexO}{Openness} Overlap & 1.000 \\
\hline
Ethical Tension & 0.000 \\
\hline
Individuals Involved & 0.722 \\
\hline
Race & 1.000 \\
\hline
Gender & 1.000 \\
\hline
Age & 0.957 \\
\hline
Scenario Realism & 0.000 \\
\hline
Situation Type & 0.611 \\
\hline
Threat Level & 0.577 \\
\hline
Time of Day & 1.000 \\
\hline
Urgency Level & 0.300 \\

\hline
\end{tabularx}
\end{table}

\section{Situational Judgment Tests (SJT) Statistics}
\label{sec:sjt_stats}

While individual SJTs can be evaluated for plausibility and trait alignment, ensuring diversity across the dataset is crucial for fairness and generalizability. We therefore compute a series of diversity metrics designed to capture variation in both surface-level attributes and underlying attribute structures, providing a comprehensive view of the dataset’s representational balance. We have created 4,000 SJTs, at an approximate budget of 150 dollars.

\subsection{Overall Diversity of SJTs}
\begin{table}[H]
\centering
\caption{Diversity Metrics for Synthetic SJT Dataset}
\begin{tabularx}{\columnwidth}{|X|c|}
\hline
\textbf{Metric Name} & \textbf{Score} \\
\hline
Per-Text TTR & 0.629 \\
\hline
Cumulative TTR & 0.005 \\
\hline
MSTTR (100) & 0.802 \\
\hline
Compression Ratio & 0.302 \\
\hline
Yule’s K & 74.748 \\
\hline
MTLD & 1.000 \\
\hline
Distinct-1 & 0.005 \\
\hline
Distinct-2 & 0.165 \\
\hline
Distinct-3 & 0.538 \\
\hline
Average Cosine Distance & 0.445 \\
\hline
\end{tabularx}
\end{table}

\subsection{Diversity of Seeds for SJT costruction}

\begin{table}[H]
\centering
\caption{Shannon Diversity index for different attributes}
\begin{tabularx}{\columnwidth}{|X|c|c|}
\hline
\textbf{Attribute Type} & \textbf{True \%} & \textbf{Derived \%} \\
\hline
Age & 1.791 & 1.795 \\
\hline
Ambiguity Level & 1.099 & 0.901 \\
\hline
Authority Relationships & 1.099 & 0.992 \\
\hline
Gender & 1.386 & 1.385 \\
\hline
Individuals Involved & 1.099 & 0.788 \\
\hline
Race & 2.079 & 2.071 \\
\hline
Situation Type & 1.944 & 1.948 \\
\hline
Threat Level & 1.098 & 1.091 \\
\hline
Time of Day & 1.386 & 1.383 \\
\hline
Urgency Level & 1.098 & 0.966 \\
\hline
\end{tabularx}
\end{table}

\begin{table}[H]
\centering
\caption{Gini-Simpson index for different attributes}
\begin{tabularx}{\columnwidth}{|X|c|c|}
\hline
\textbf{Attribute Type} & \textbf{True \%} & \textbf{Derived \%} \\
\hline
Age & 5.922 & 5.977 \\
\hline
Ambiguity Level & 2.999 & 2.291 \\
\hline
Authority Relationships & 3.000 & 2.533 \\
\hline
Gender & 3.996 & 3.991 \\
\hline
Individuals Involved & 3.000 & 2.007 \\
\hline
Race & 7.989 & 7.864 \\
\hline
Situation Type & 6.979 & 6.963 \\
\hline
Threat Level & 2.999 & 2.956 \\
\hline
Time of Day & 3.995 & 3.977 \\
\hline
Urgency Level & 2.997 & 2.367 \\
\hline
\end{tabularx}
\end{table}

\subsection{Distribution over Input Seeds}

We used different randomly sampled combination of seed values to create SJTs synthetically. 

\textbf{Total No of SJTs created: 4000}

\subsubsection{Age}

\begin{table}[H]
\centering
\begin{tabularx}{\columnwidth}{|X|c|}
\hline
\textbf{Attribute} & \textbf{Distribution across Total SJTs} \\
\hline
Middle Aged & 17.75 \% \\
\hline
Senior & 16.90 \%\\
\hline
Young Adult & 16.73 \%\\
\hline
Juvenile & 16.50 \%\\
\hline
Unknown & 16.38 \%\\
\hline
Adult & 15.75 \%\\
\hline
\end{tabularx}
\end{table}

\subsubsection{Ambiguity Level}

\begin{table}[H]
\centering
\begin{tabularx}{\columnwidth}{|X|c|}
\hline
\textbf{Attribute} & \textbf{Distribution across Total SJTs} \\
\hline
High & 33.90 \% \\
\hline
Moderate & 33.33 \%\\
\hline
Clear & 32.78 \%\\
\hline
\end{tabularx}
\end{table}

\subsubsection{Authority Relationships}

\begin{table}[H]
\centering
\begin{tabularx}{\columnwidth}{|X|c|}
\hline
\textbf{Attribute} & \textbf{Distribution across Total SJTs} \\
\hline
Peer Level & 33.68 \% \\
\hline
Subordinate & 33.53 \%\\
\hline
Authority & 32.80 \%\\
\hline
\end{tabularx}
\end{table}

\subsubsection{Ethical Considerations}

\begin{table}[H]
\centering
\begin{tabularx}{\columnwidth}{|X|c|}
\hline
\textbf{Attribute} & \textbf{Distribution across Total SJTs} \\
\hline
Procedure vs Innovation & 21.08 \% \\
\hline
Policy Compliance vs Shortcuts & 20.55 \%\\
\hline
Authority vs Compassion & 20.35 \%\\
\hline
Transparency vs Self Protection & 19.93 \%\\
\hline
Individual vs Team Loyalty & 18.10 \%\\
\hline
\end{tabularx}
\end{table}

\subsubsection{Gender}

\begin{table}[H]
\centering
\begin{tabularx}{\columnwidth}{|X|c|}
\hline
\textbf{Attribute} & \textbf{Distribution across Total SJTs} \\
\hline
Female & 25.65 \% \\
\hline
Male & 25.53 \%\\
\hline
Non Binary & 25.10 \%\\
\hline
Unknown & 23.73 \%\\
\hline
\end{tabularx}
\end{table}

\subsubsection{Individuals Involved}

\begin{table}[H]
\centering
\begin{tabularx}{\columnwidth}{|X|c|}
\hline
\textbf{Attribute} & \textbf{Distribution across Total SJTs} \\
\hline
Complex & 33.45 \% \\
\hline
Moderate & 33.43 \%\\
\hline
Simple & 33.125 \%\\
\hline
\end{tabularx}
\end{table}

\subsubsection{Threat Level}

\begin{table}[H]
\centering
\begin{tabularx}{\columnwidth}{|X|c|}
\hline
\textbf{Attribute} & \textbf{Distribution across Total SJTs} \\
\hline
High & 34.05 \% \\
\hline
Low & 33.43 \%\\
\hline
Medium & 32.53 \%\\
\hline
\end{tabularx}
\end{table}

\subsubsection{Time of Day}

\begin{table}[H]
\centering
\begin{tabularx}{\columnwidth}{|X|c|}
\hline
\textbf{Attribute} & \textbf{Distribution across Total SJTs} \\
\hline
Morning & 25.95 \% \\
\hline
Evening & 25.73 \%\\
\hline
Afternoon & 24.20 \%\\
\hline
Night & 24.13 \%\\
\hline
\end{tabularx}
\end{table}

\subsubsection{Urgency Level}

\begin{table}[H]
\centering
\begin{tabularx}{\columnwidth}{|X|c|}
\hline
\textbf{Attribute} & \textbf{Distribution across Total SJTs} \\
\hline
Low & 34.55 \% \\
\hline
Medium & 33.48 \%\\
\hline
High & 31.98 \%\\
\hline
\end{tabularx}
\end{table}

\subsubsection{Race}

\begin{table}[H]
\centering
\begin{tabularx}{\columnwidth}{|X|c|}
\hline
\textbf{Attribute} & \textbf{Distribution across Total SJTs} \\
\hline
Unknown & 13.35 \% \\
\hline
White & 12.95 \%\\
\hline
Hispanic /Latino & 12.63 \%\\
\hline
Black /African American & 12.63 \%\\
\hline
Other Multiracial & 12.45 \%\\
\hline
Asian & 12.050 \%\\
\hline
Native American /Alaska Native & 12.00 \%\\
\hline
Pacific Islander & 11.95 \%\\
\hline

\end{tabularx}
\end{table}

\subsubsection{Situation Type}

\begin{table}[H]
\centering
\begin{tabularx}{\columnwidth}{|X|c|}
\hline
\textbf{Attribute} & \textbf{Distribution across Total SJTs} \\
\hline
Crime Scene Investigation & 15.23 \% \\
\hline
Mental Health Crisis & 15.18 \%\\
\hline
Emergency Response & 14.375 \%\\
\hline
Administrative Reporting & 14.30 \%\\
\hline
Inter Agency Cooperation & 14.28 \%\\
\hline
Training Supervision & 13.95 \%\\
\hline
Patrol Traffic Stop & 12.70 \%\\
\hline

\end{tabularx}
\end{table}

\subsection{Distribution over Derived Seeds}

\textbf{Total No of SJTs evaluated: 1000}

Following the LLM-as-a-Judge evaluation, we extract the inferred seed attributes from the generated SJTs. The resulting distributions of these seed values across the evaluated scenarios are presented below.

\subsubsection{Age}

\begin{table}[H]
\centering
\begin{tabularx}{\columnwidth}{|X|c|}
\hline
\textbf{Attribute} & \textbf{Distribution across Total SJTs} \\
\hline
Middle Aged & 16.10 \% \\
\hline
Senior & 16.10 \%\\
\hline
Young Adult & 15.60 \%\\
\hline
Juvenile & 17.30 \%\\
\hline
Unknown & 19.20 \%\\
\hline
Adult & 15.60 \%\\
\hline
\end{tabularx}
\end{table}

\subsubsection{Situation Ambiguity Level}

\begin{table}[H]
\centering
\begin{tabularx}{\columnwidth}{|X|c|}
\hline
\textbf{Attribute} & \textbf{Distribution across Total SJTs} \\
\hline
High & 44.50 \% \\
\hline
Moderate & 48.30 \%\\
\hline
Clear & 7.20 \%\\
\hline
\end{tabularx}
\end{table}

\subsubsection{Subject Authority Relationships}

\begin{table}[H]
\centering
\begin{tabularx}{\columnwidth}{|X|c|}
\hline
\textbf{Attribute} & \textbf{Distribution across Total SJTs} \\
\hline
Peer Level & 45.90 \% \\
\hline
Subordinate & 13.30 \%\\
\hline
Authority & 40.80 \%\\
\hline
\end{tabularx}
\end{table}

\subsubsection{Gender}

\begin{table}[H]
\centering
\begin{tabularx}{\columnwidth}{|X|c|}
\hline
\textbf{Attribute} & \textbf{Distribution across Total SJTs} \\
\hline
Female & 23.70 \% \\
\hline
Male & 26.70 \%\\
\hline
Non Binary & 24.10 \%\\
\hline
Unknown & 25.50 \%\\
\hline
\end{tabularx}
\end{table}

\subsubsection{Individuals Involved}

\begin{table}[H]
\centering
\begin{tabularx}{\columnwidth}{|X|c|}
\hline
\textbf{Attribute} & \textbf{Distribution across Total SJTs} \\
\hline
Complex & 60.80 \% \\
\hline
Moderate & 35.70 \%\\
\hline
Simple & 3.50 \%\\
\hline
\end{tabularx}
\end{table}

\subsubsection{Threat Level}

\begin{table}[H]
\centering
\begin{tabularx}{\columnwidth}{|X|c|}
\hline
\textbf{Attribute} & \textbf{Distribution across Total SJTs} \\
\hline
High & 28.70 \% \\
\hline
Low & 38.60 \%\\
\hline
Medium & 32.70 \%\\
\hline
\end{tabularx}
\end{table}

\subsubsection{Time of Day}

\begin{table}[H]
\centering
\begin{tabularx}{\columnwidth}{|X|c|}
\hline
\textbf{Attribute} & \textbf{Distribution across Total SJTs} \\
\hline
Morning & 26.50 \% \\
\hline
Evening & 27.00 \%\\
\hline
Afternoon & 24.30 \%\\
\hline
Night & 22.20 \%\\
\hline
\end{tabularx}
\end{table}

\subsubsection{Urgency Level}

\begin{table}[H]
\centering
\begin{tabularx}{\columnwidth}{|X|c|}
\hline
\textbf{Attribute} & \textbf{Distribution across Total SJTs} \\
\hline
Low & 15.00 \% \\
\hline
Medium & 28.60 \%\\
\hline
High & 56.40 \%\\
\hline
\end{tabularx}
\end{table}

\subsubsection{Race}

\begin{table}[H]
\centering
\begin{tabularx}{\columnwidth}{|X|c|}
\hline
\textbf{Attribute} & \textbf{Distribution across Total SJTs} \\
\hline
Unknown & 16.50 \% \\
\hline
White & 13.30 \%\\
\hline
Hispanic /Latino & 12.20 \%\\
\hline
Black /African American & 11.30 \%\\
\hline
Other Multiracial & 11.00 \%\\
\hline
Asian & 11.70 \%\\
\hline
Native American /Alaska Native & 12.10 \%\\
\hline
Pacific Islander & 11.90 \%\\
\hline

\end{tabularx}
\end{table}

\subsubsection{Situation Type}

\begin{table}[H]
\centering
\begin{tabularx}{\columnwidth}{|X|c|}
\hline
\textbf{Attribute} & \textbf{Distribution across Total SJTs} \\
\hline
Crime Scene Investigation & 15.80 \% \\
\hline
Mental Health Crisis & 13.60 \%\\
\hline
Emergency Response & 15.30 \%\\
\hline
Administrative Reporting & 15.10 \%\\
\hline
Inter Agency Cooperation & 12.50 \%\\
\hline
Training Supervision & 12.70 \%\\
\hline
Patrol Traffic Stop & 14.90 \%\\
\hline

\end{tabularx}
\end{table}

\subsubsection{Age}

\begin{figure}[H]
\centering
\includegraphics[width=0.9\columnwidth]{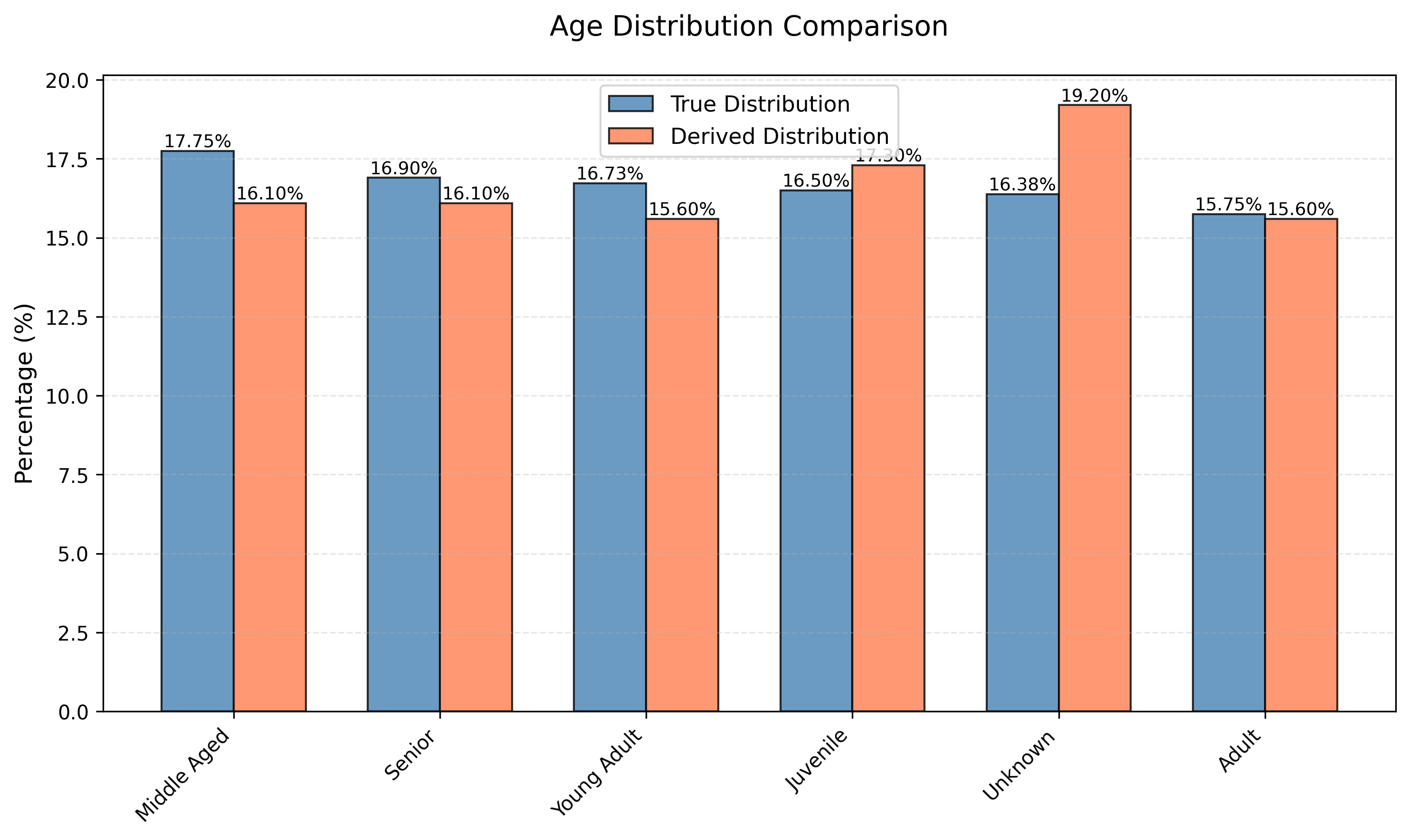}
\caption{}
\end{figure}


\subsubsection{Ambiguity Level}

\begin{figure}[H]
\centering
\includegraphics[width=0.9\columnwidth]{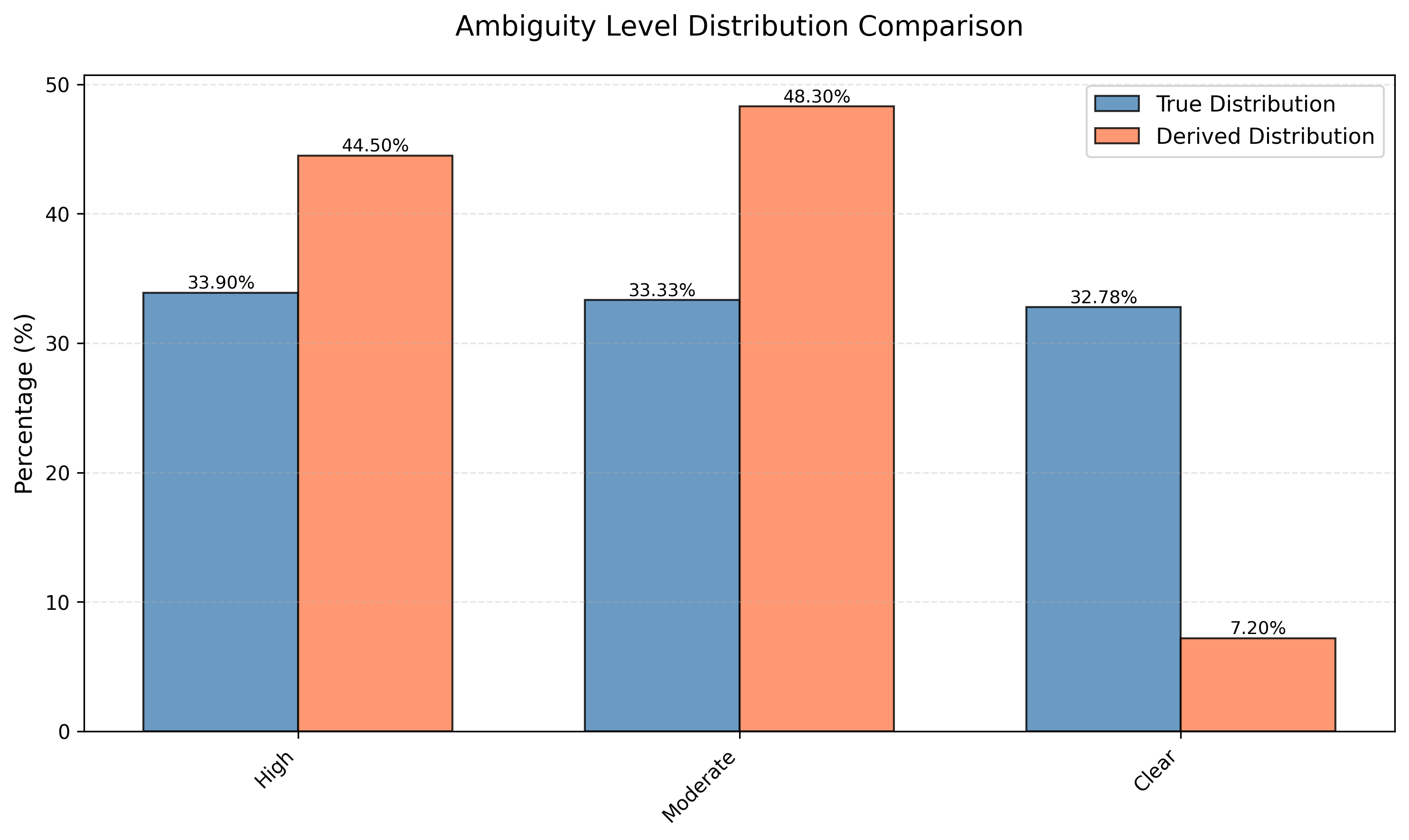}
\caption{}
\end{figure}


\subsubsection{Authority Relationships}

\begin{figure}[H]
\centering
\includegraphics[width=0.9\columnwidth]{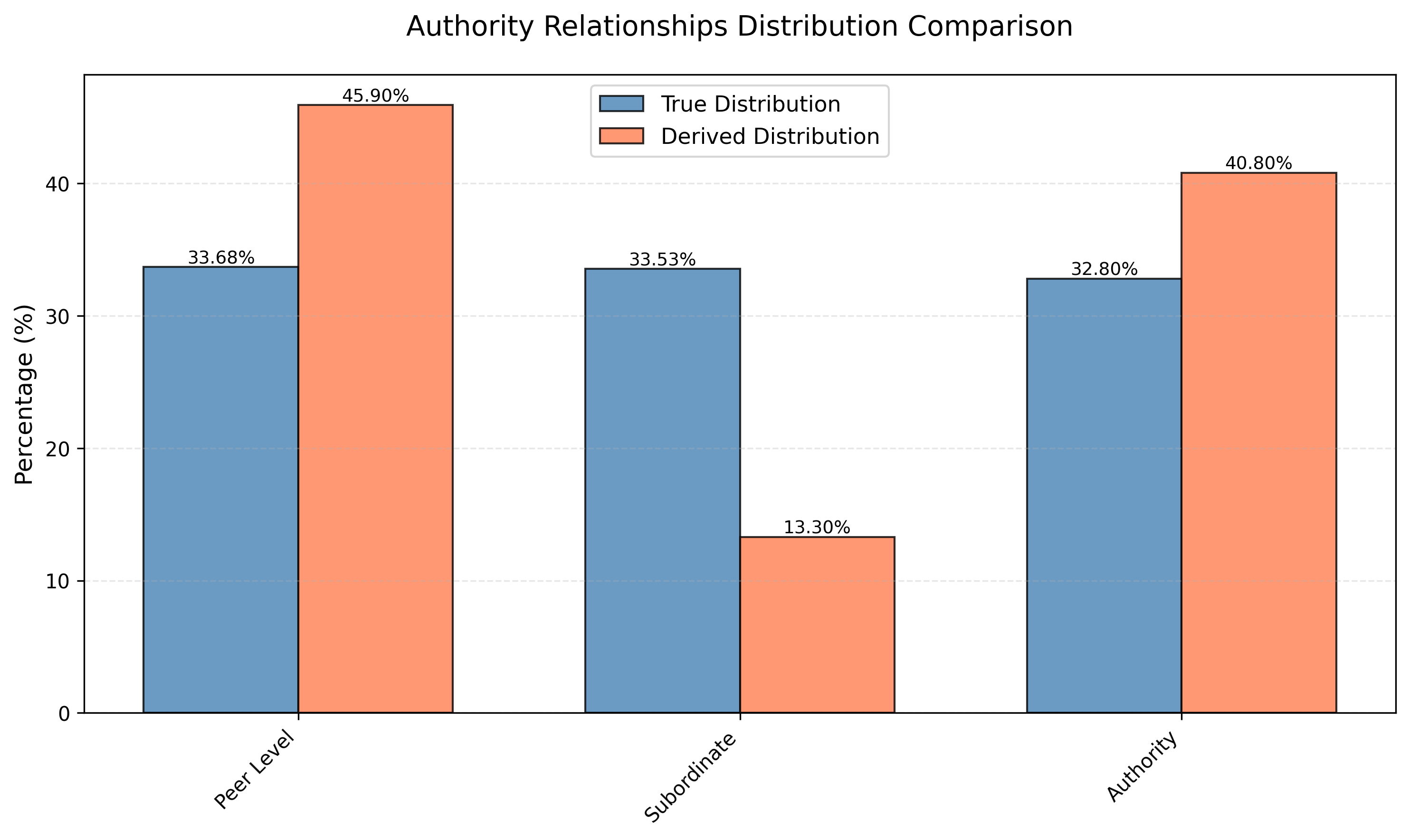}
\caption{}
\end{figure}




\subsubsection{Gender}

\begin{figure}[H]
\centering
\includegraphics[width=0.9\columnwidth]{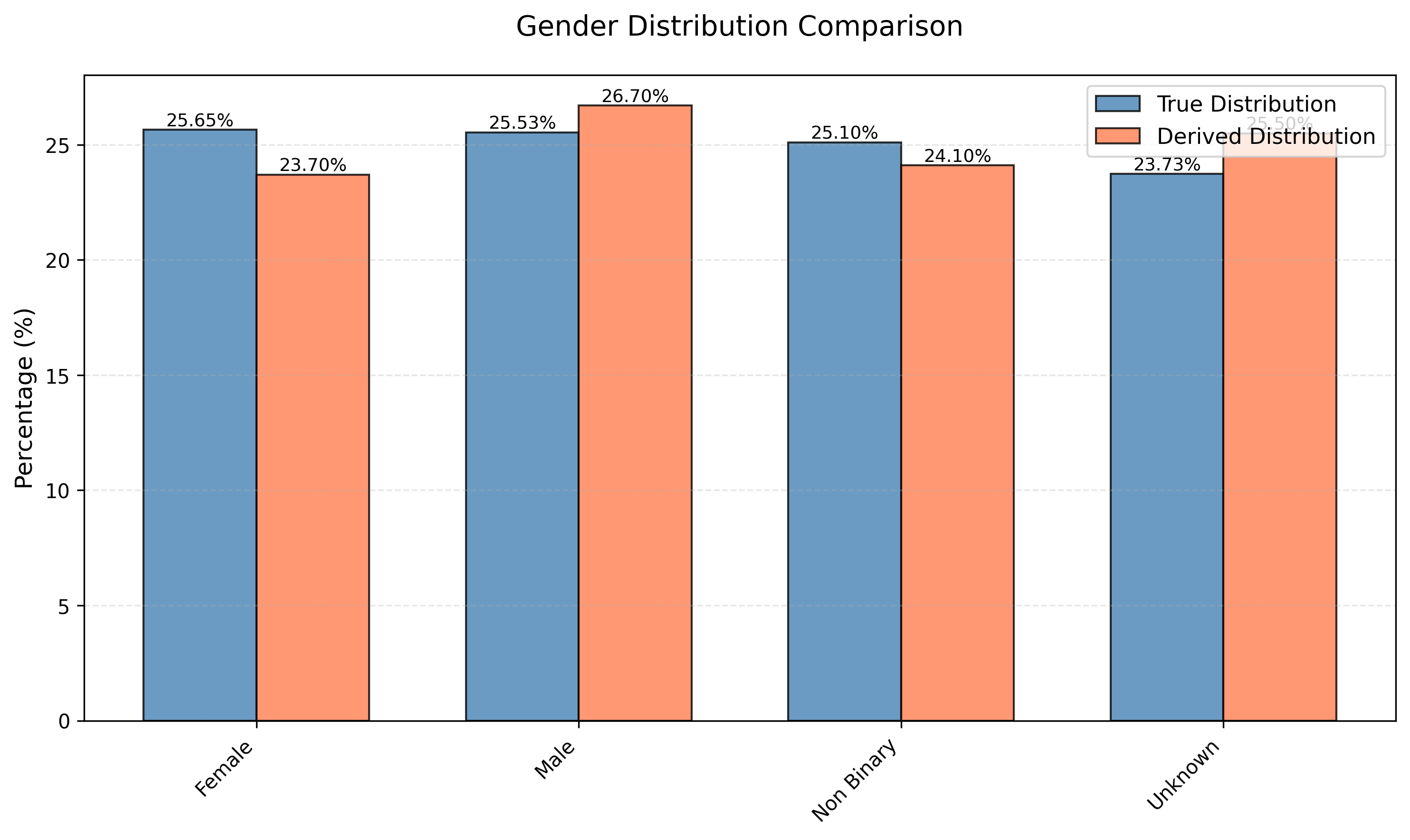}
\caption{}
\end{figure}


\subsubsection{Individuals Involved}

\begin{figure}[H]
\centering
\includegraphics[width=0.9\columnwidth]{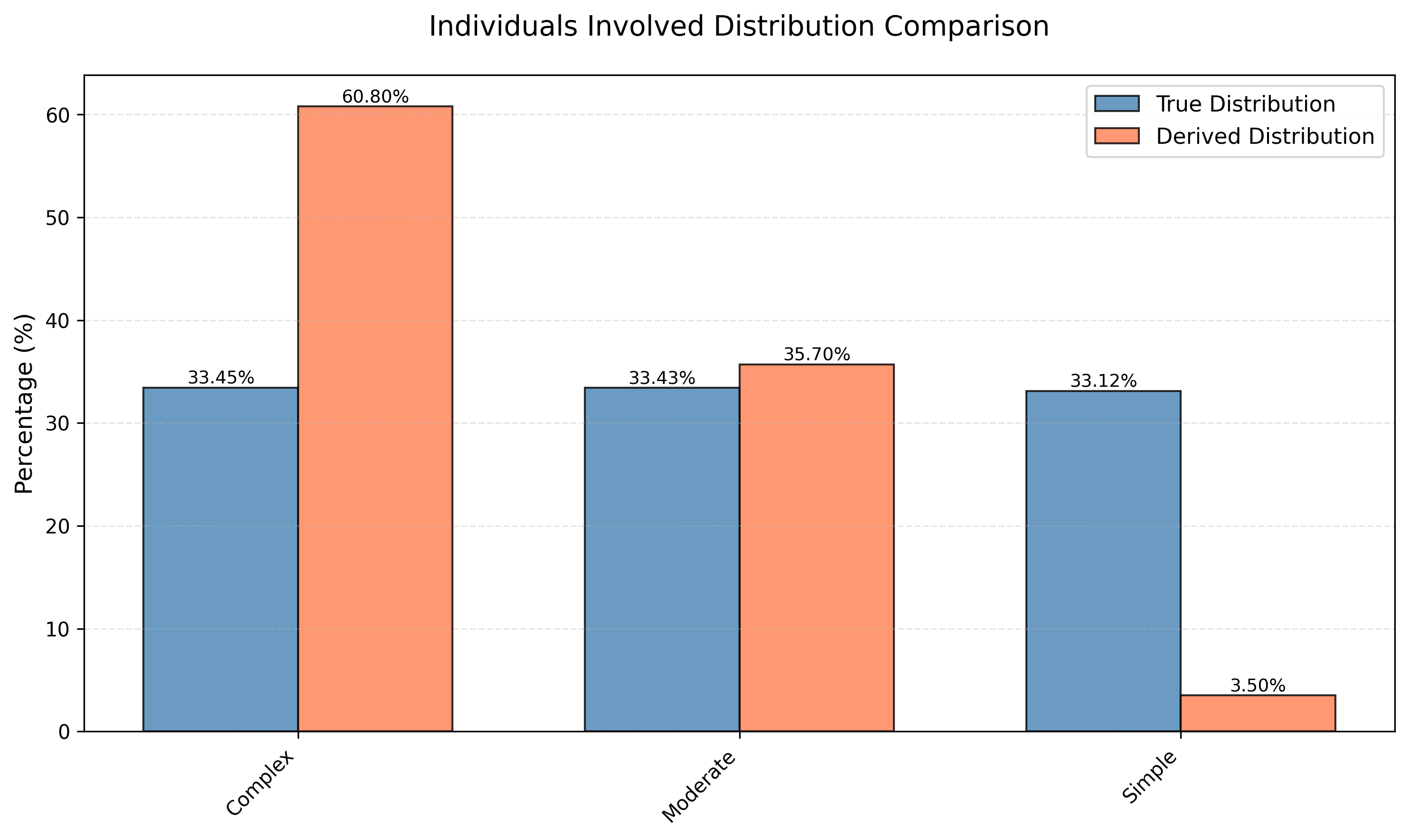}
\caption{}
\end{figure}


\subsubsection{Threat Level}

\begin{figure}[H]
\centering
\includegraphics[width=0.9\columnwidth]{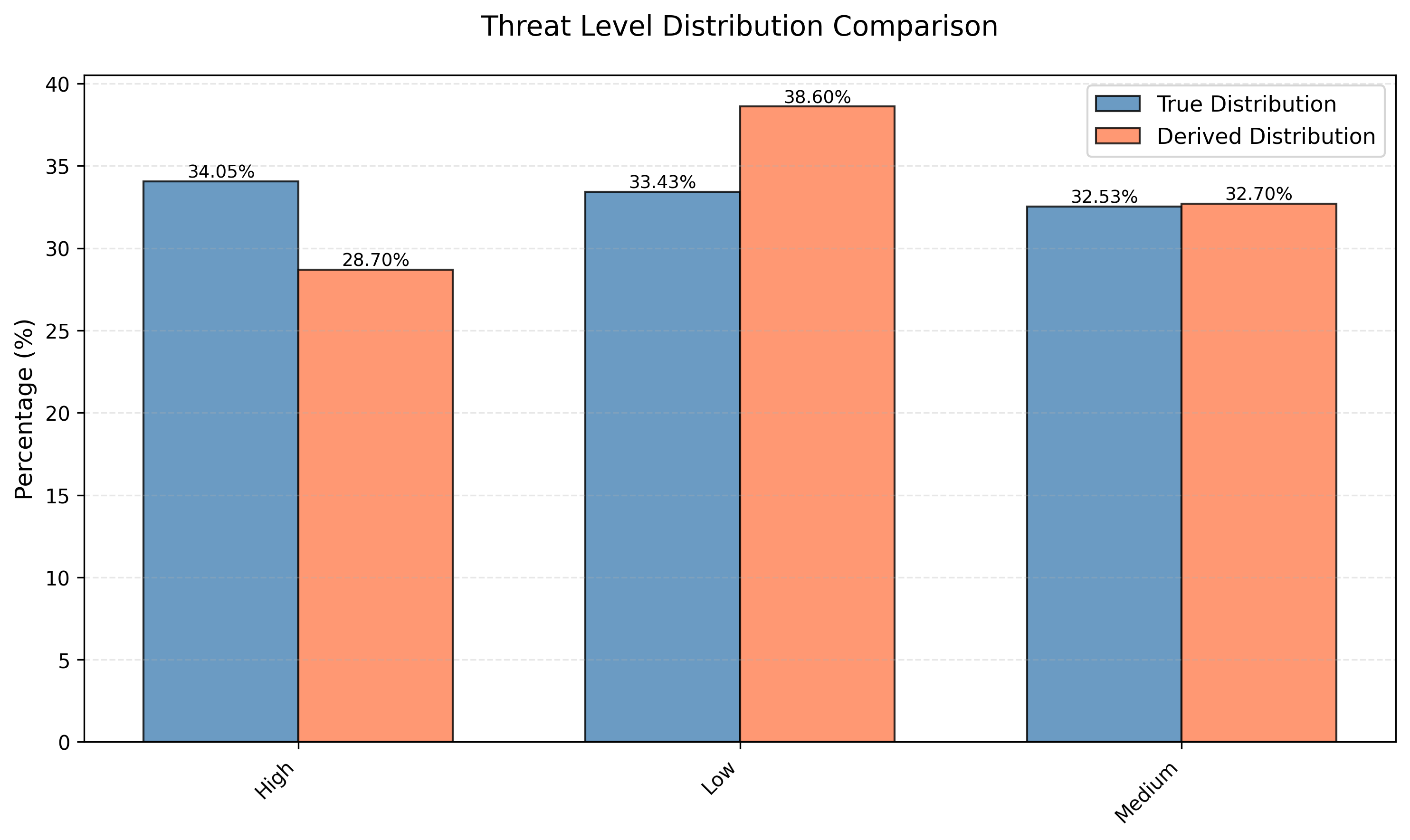}
\caption{}
\end{figure}


\subsubsection{Time of Day}

\begin{figure}[H]
\centering
\includegraphics[width=0.9\columnwidth]{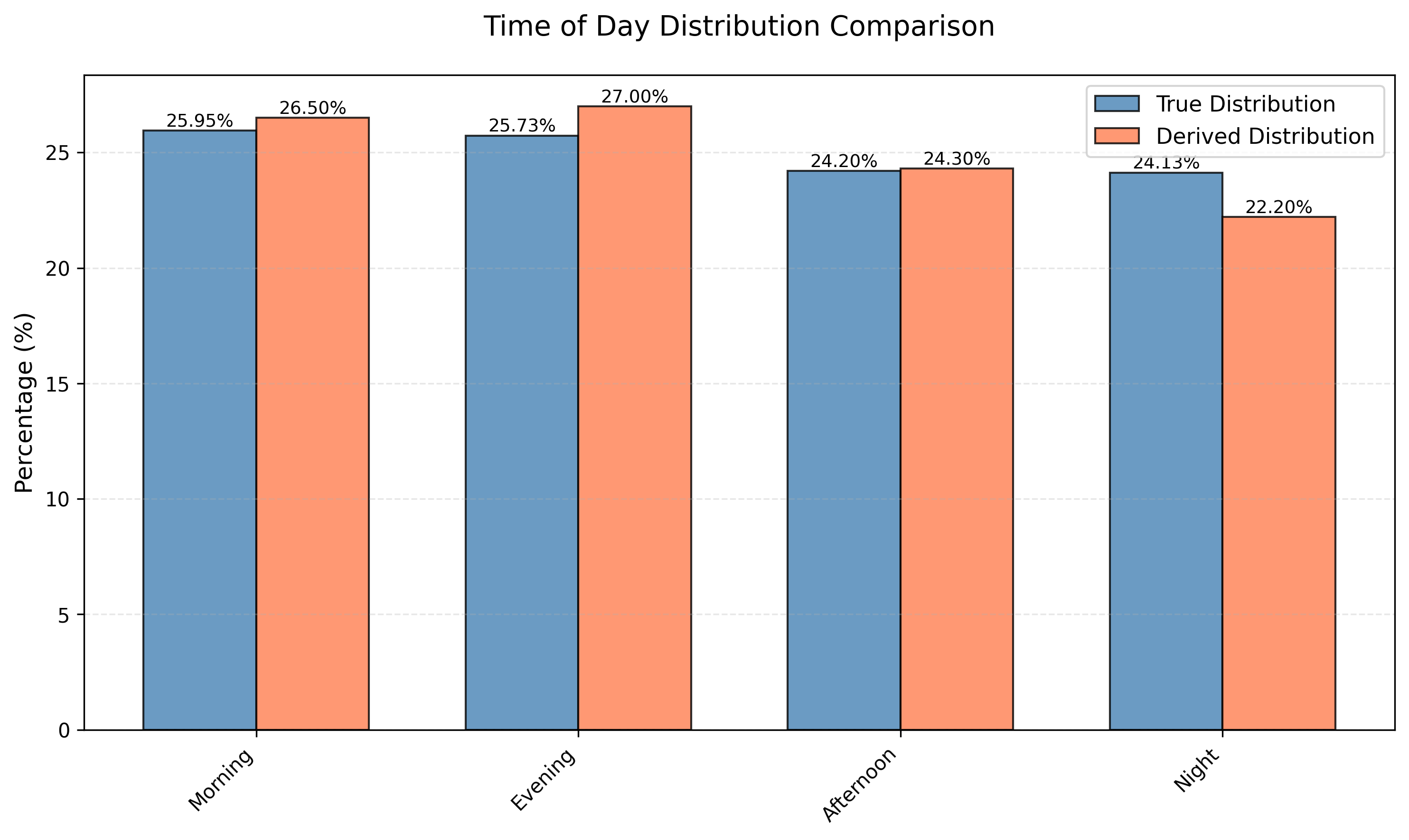}
\caption{}
\end{figure}


\subsubsection{Urgency Level}

\begin{figure}[H]
\centering
\includegraphics[width=0.9\columnwidth]{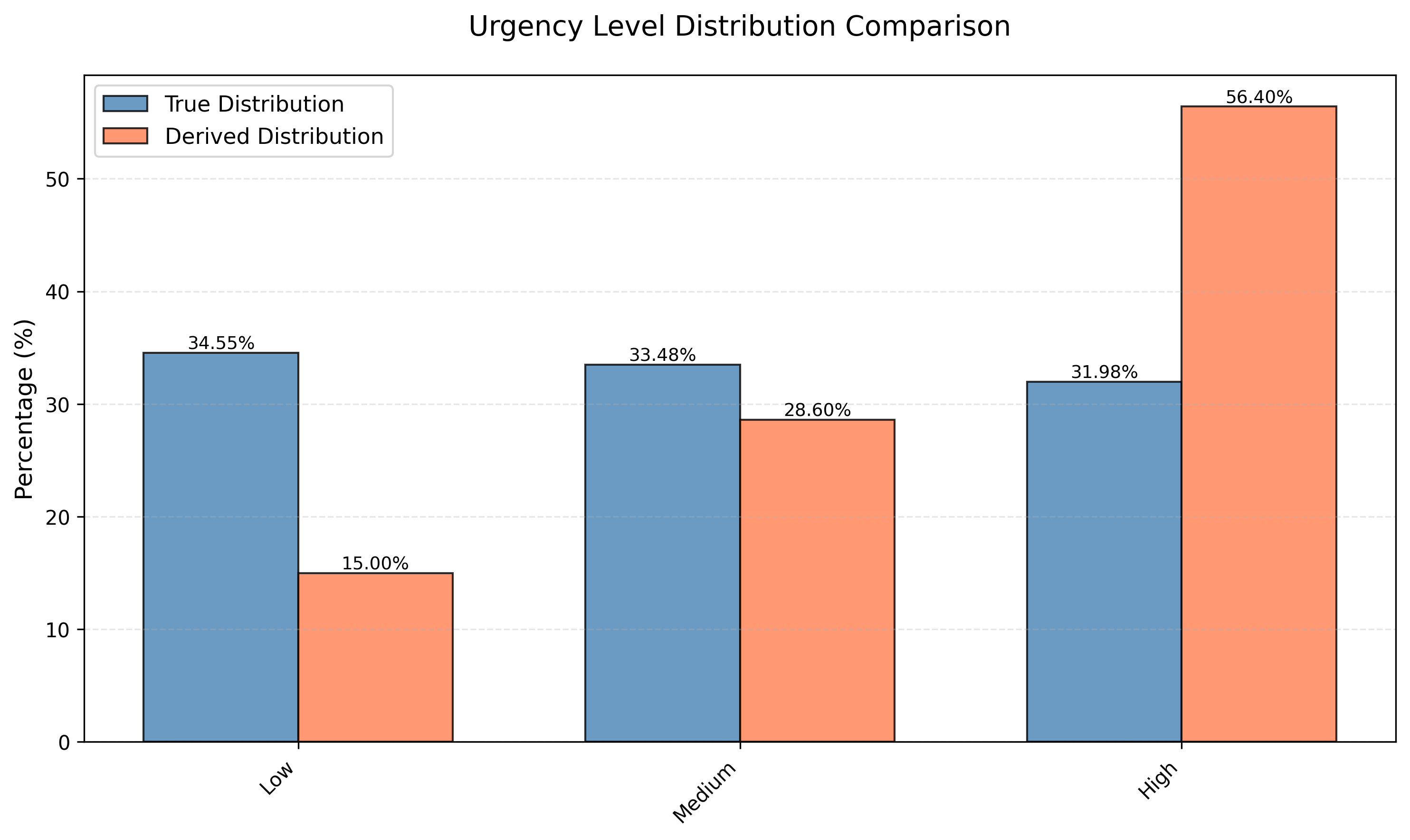}
\caption{}
\end{figure}


\subsubsection{Race}

\begin{figure}[H]
\centering
\includegraphics[width=0.9\columnwidth]{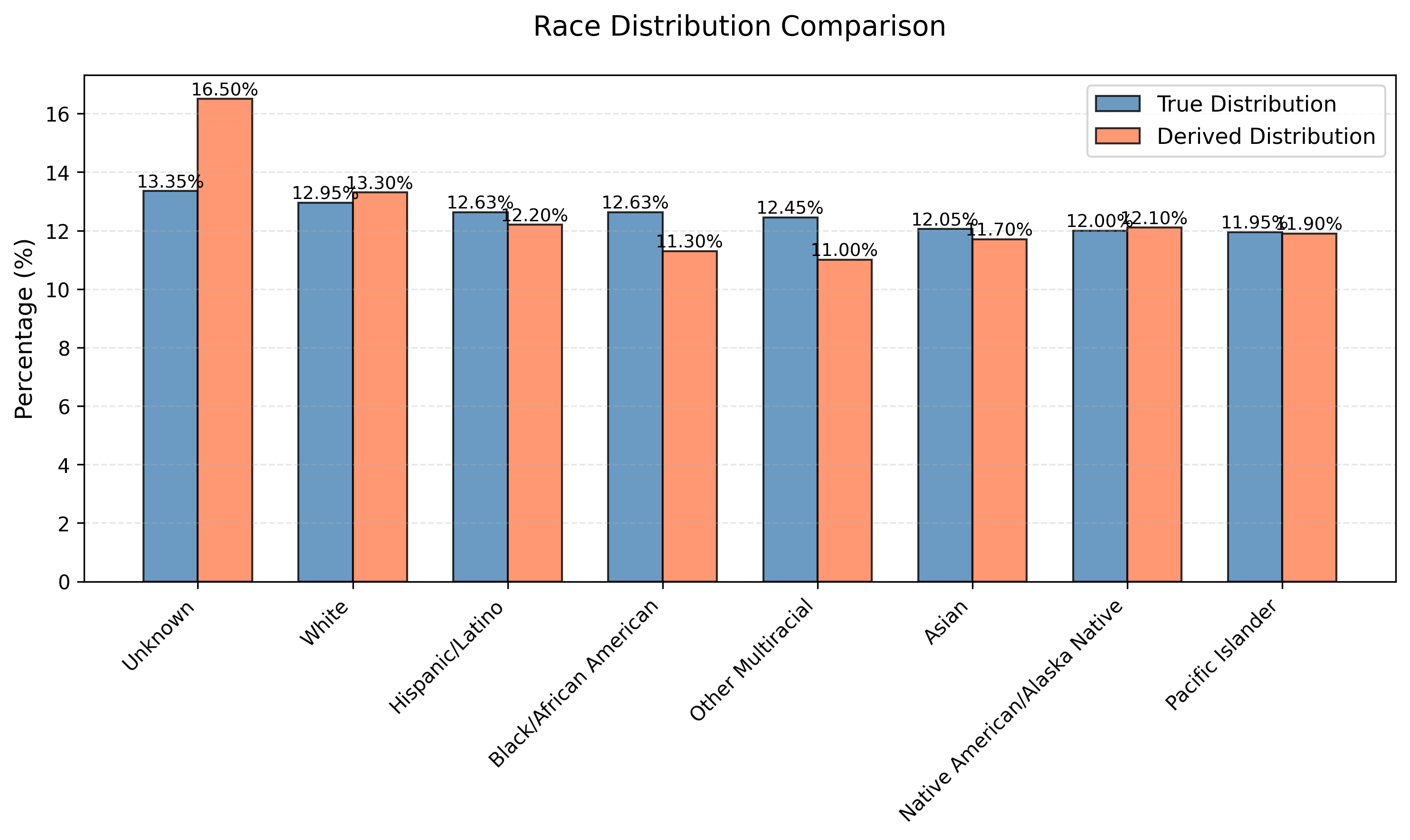}
\caption{}
\end{figure}


\subsubsection{Situation Type}

\begin{figure}[H]
\centering
\includegraphics[width=0.9\columnwidth]{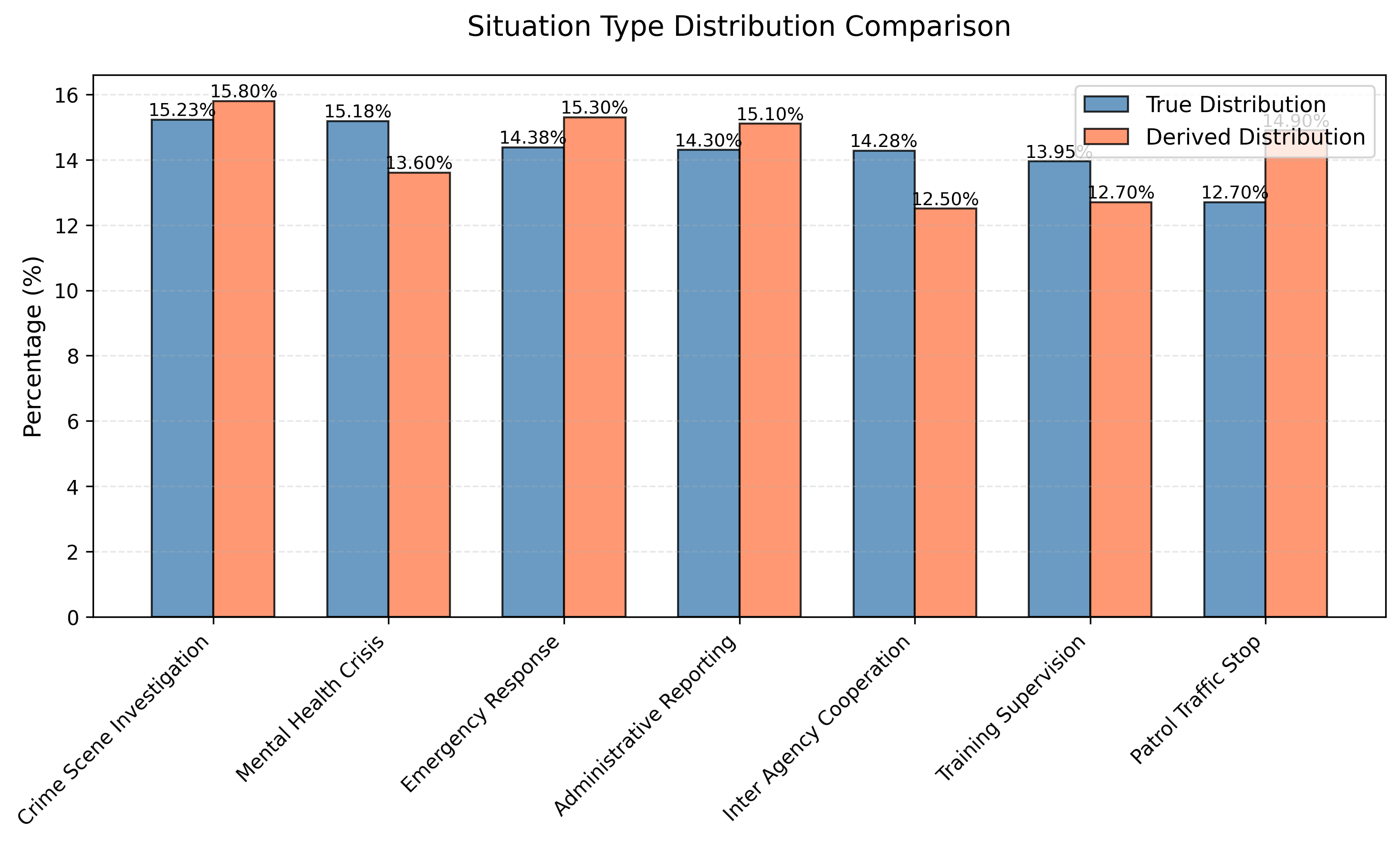}
\caption{}
\end{figure}


\input{arxiv_files/analysis}

\section{SJT Creation and Evaluation Prompt Templates}
\label{app:prompt templates}
\begin{tcolorbox}[enhanced,
    colback=gray!5,
    colframe=black!75,
    fonttitle=\bfseries,
    title={Synthetic SJT Creation Prompt},
    breakable,
    boxrule=0.6pt
]
\small
You are creating a new law enforcement Situational Judgment Test (SJT) scenario by modifying an existing scenario with new attribute values. Follow the template below to generate a realistic, professionally appropriate scenario that maintains the core decision-making structure while incorporating the specified attributes.

\textbf{Base Scenario to Modify:} \\
\{\{base\_scenario\}\}

\medskip
\textbf{Instructions}
\begin{itemize}
\item Keep the same personality trait dimensions being tested (\textcolor{hexH}{Honesty-Humility}, \textcolor{hexE}{Emotionality}, \textcolor{hexX}{eXtraversion}, \textcolor{hexA}{Agreeableness}, \textcolor{hexC}{Conscientiousness}, \textcolor{hexO}{Openness to Experience})
\item Preserve the professional law enforcement context
\end{itemize}

\textbf{Attribute description}
\begin{itemize}
\item \textbf{Urgency Level:} Adjust time pressure and decision timeline accordingly
  \begin{itemize}
  \item Low: Allow deliberation time, non-critical timing
  \item Medium: Some time pressure, manageable deadlines
  \item High: Immediate decisions required, critical timing
  \end{itemize}
\item \textbf{Threat Level:} Scale physical danger and safety concerns
  \begin{itemize}
  \item Low: Administrative issues, minor policy matters
  \item Medium: Potential for injury, safety protocols needed
  \item High: Life-threatening situations, lethal force considerations
  \end{itemize}
\item \textbf{Ambiguity Level:} Modify clarity of protocols and guidance
  \begin{itemize}
  \item Clear: Obvious procedures, straightforward application
  \item Moderate: Some judgment required, minor gray areas
  \item High: Conflicting guidance, novel situations, ethical dilemmas
  \end{itemize}
\item \textbf{Individuals Involved:} Adjust scenario complexity
  \begin{itemize}
  \item Simple: Officer making individual decision
  \item Moderate: 2--3 parties with different perspectives
  \item Complex: Multiple stakeholders, witnesses, supervisors
  \end{itemize}
\item \textbf{Authority Relationships:} Frame interactions appropriately
  \begin{itemize}
  \item Peer Level: Fellow officers, partners, colleagues
  \item Subordinate: Supervisors, training officers, senior personnel
  \item Authority: Suspects, witnesses, civilians, subordinates
  \end{itemize}
\item \textbf{Ethical Considerations:} Incorporate specified ethical tension
\item \textbf{Situation Type:} Adapt setting and context to match type
\item \textbf{Time of Day:} Include time context naturally in scenario
\item \textbf{Demographics:} Integrate subject's race, gender, and age naturally without stereotyping if applicable.
\end{itemize}

\textbf{New Attribute Values:}
\begin{itemize}
\item Urgency Level: \{\{urgency\_level\}\}
\item Threat Level: \{\{threat\_level\}\}
\item Ambiguity Level: \{\{ambiguity\_level\}\}
\item Individuals Involved: \{\{individuals\_involved\}\}
\item Authority Relationships: \{\{authority\_relationships\}\}
\item Ethical Considerations: \{\{ethical\_considerations\}\}
\item Situation Type: \{\{situation\_type\}\}
\item Time of Day: \{\{time\_of\_day\}\}
\item Subject Race: \{\{race\}\}
\item Subject Gender: \{\{gender\}\}
\item Subject Age: \{\{age\}\}
\end{itemize}

\textbf{Response Options Requirements}
\begin{itemize}
\item Generate six response options (1--6) that:
\begin{itemize}
  \item Clearly differentiate the six personality dimensions
  \item Remain professionally appropriate and realistic
  \item Reflect how each personality type would approach the modified scenario
  \item Maintain consistent quality and plausibility across all options
  \item The responses are qualitative descriptions of actions and do not contain specific speech suggestions.
  \item Integrate all specified attributes naturally into the scenario without explicitly naming them
  \item Do not use direct labels like ``high-priority,'' ``mental health crisis,'' ``high threat,'' etc.
  \item Make attribute levels apparent through context, actions, and circumstances
  \item Let urgency, threat level, and situation type emerge implicitly
  \item Ensure scenario is realistic and could occur in actual law enforcement
  \item Scenario should be 2--4 sentences
  \item Include specific, actionable decisions rather than vague choices
  \item Verify that all six personality responses are distinct and characteristic
\end{itemize}
\end{itemize}

\textbf{Output Format}
\begin{verbatim}
{
  'question': '<string>',
  'honesty_humility_option': '<string>',
  '\textcolor{hexE}{Emotionality}_option': '<string>',
  '\textcolor{hexX}{eXtraversion}_option': '<string>',
  '\textcolor{hexA}{Agreeableness}_option': '<string>',
  '\textcolor{hexC}{Conscientiousness}_option': '<string>',
  '\textcolor{hexO}{Openness}_option': '<string>'
}
\end{verbatim}
Do not include any extra text, explanation, or formatting outside of the JSON object.
\end{tcolorbox}

\begin{tcolorbox}[enhanced,
    colback=gray!5,
    colframe=black!75,
    fonttitle=\bfseries,
    title={SJT Trait Bleed Evaluation Prompt},
    breakable,
    boxrule=0.6pt
]
\small

\textbf{Role:} You are an expert SJT evaluator and corrector specializing in \hexaco\;-aligned scenarios.  
You are given a situational judgment test (SJT) scenario with six answer options, each intended to correspond to one \hexaco\;  trait: {\textcolor{hexH}{Honesty-Humility}, \textcolor{hexE}{Emotionality}, \textcolor{hexX}{eXtraversion}, \textcolor{hexA}{Agreeableness}, \textcolor{hexC}{Conscientiousness}, and \textcolor{hexO}{Openness}}.

\medskip
\textbf{Tasks:}
\begin{enumerate}
    \item \textbf{Trait Fit Evaluation}  
    For each option, evaluate how strongly it aligns with its intended trait definition. Use a 1–5 scale:  
    \begin{itemize}
        \item 5 = Very strong, clean representation, no leakage  
        \item 4 = Strong but with minor overlap  
        \item 3 = Moderate, noticeable blending  
        \item 2 = Weak, trait unclear or diluted  
        \item 1 = Poor, option does not represent the trait well  
    \end{itemize}

    \item \textbf{Separation Analysis}  
    Highlight where options overlap or bleed into each other (e.g., \textcolor{hexX}{eXtraversion} vs. \textcolor{hexA}{Agreeableness}). Explain why the overlap occurs.  

    \item \textbf{Correction Suggestions}  
    For any option rated below 5, propose a corrected rewrite that emphasizes the target trait more cleanly. Ensure each rewrite minimizes overlap with other traits and includes specific, actionable decisions rather than vague choices.  

    \item \textbf{Final Corrected SJT Object}  
    Output an object with the exact same structure as the input SJT dictionary. Each option should contain the corrected version if a rewrite was needed, or the unchanged original if not.  

    \item \textbf{Output Format}  
    Return results in structured JSON with this schema:
\begin{verbatim}
{
  "scenario_summary": "<1-2 sentence summary of scenario>",
  "trait_evaluations": {
    "honesty_humility": {
      "score": <1-5>,
      "analysis": "<why it fits/doesn't fit>",
      "suggested_correction": "<if needed, otherwise null>"
    },
    "\textcolor{hexE}{Emotionality}": {
      "score": <1-5>,
      "analysis": "<why it fits/doesn't fit>",
      "suggested_correction": "<if needed, otherwise null>"
    },
    "\textcolor{hexX}{eXtraversion}": {
      "score": <1-5>,
      "analysis": "<why it fits/doesn't fit>",
      "suggested_correction": "<if needed, otherwise null>"
    },
    "\textcolor{hexA}{Agreeableness}": {
      "score": <1-5>,
      "analysis": "<why it fits/doesn't fit>",
      "suggested_correction": "<if needed, otherwise null>"
    },
    "\textcolor{hexC}{Conscientiousness}": {
      "score": <1-5>,
      "analysis": "<why it fits/doesn't fit>",
      "suggested_correction": "<if needed, otherwise null>"
    },
    "\textcolor{hexO}{Openness}": {
      "score": <1-5>,
      "analysis": "<why it fits/doesn't fit>",
      "suggested_correction": "<if needed, otherwise null>"
    },
  "corrected_sjt": {
    "question": "<original or unchanged question>",
    "honesty_humility_option": "<corrected or unchanged option>",
    "\textcolor{hexE}{Emotionality}_option": "<corrected or unchanged option>",
    "\textcolor{hexX}{eXtraversion}_option": "<corrected or unchanged option>",
    "\textcolor{hexA}{Agreeableness}_option": "<corrected or unchanged option>",
    "\textcolor{hexC}{Conscientiousness}_option": "<corrected or unchanged option>",
    "\textcolor{hexO}{Openness}_option": "<corrected or unchanged option>"
  },
  "overall_notes": "<high-level summary of trait separation quality>"
}
}
\end{verbatim}
\end{enumerate}

\medskip
\textbf{SJT Input Template:}
\begin{verbatim}
{
  "question": {{ question }},
  "honesty_humility_option": {{ honesty_humility_option }},
  "\textcolor{hexE}{Emotionality}_option": {{ \textcolor{hexE}{Emotionality}_option }},
  "\textcolor{hexX}{eXtraversion}_option": {{ \textcolor{hexX}{eXtraversion}_option }},
  "\textcolor{hexA}{Agreeableness}_option": {{ \textcolor{hexA}{Agreeableness}_option }},
  "\textcolor{hexC}{Conscientiousness}_option": {{ \textcolor{hexC}{Conscientiousness}_option }},
  "\textcolor{hexO}{Openness}_option": {{ \textcolor{hexO}{Openness}_option }}
}
\end{verbatim}
\end{tcolorbox}

\begin{tcolorbox}[enhanced,
    colback=gray!5,
    colframe=black!75,
    fonttitle=\bfseries,
    title={SJT LLM Judge Evalution Rubric 1 Prompt},
    breakable,
    boxrule=0.6pt
]
\small

\textbf{Role:} You are an expert evaluator of situational judgment tests (SJTs).  
Your role is to assess the quality of each SJT scenario and its response options using a structured rubric.  
You must score each dimension on a 1–5 scale and provide a concise justification.  
Be objective, consistent, and fair. Always return results in JSON format.

\medskip
\textbf{Task:} Evaluate the following SJT using the rubric provided.  

\medskip
\textbf{Situational Judgment Test:}
\begin{verbatim}
Question: {{ question }}

Answer Options:
{{ answer_options }}
\end{verbatim}

\medskip
\textbf{Question Description:}
\begin{itemize}
    \item Every option corresponds to a \hexaco\;  trait in fixed order:  
    1. \textcolor{hexH}{Honesty-Humility}, 2. \textcolor{hexE}{Emotionality}, 3. \textcolor{hexX}{eXtraversion},  
    4. \textcolor{hexA}{Agreeableness}, 5. \textcolor{hexC}{Conscientiousness}, 6. \textcolor{hexO}{Openness to Experience}.
    \item Questions and answers are created using specific seed values.
    \item SJT simulates a professional law enforcement context.
\end{itemize}

\medskip
\textbf{Seed Attributes:}  
\begin{itemize}
    \item \textbf{Urgency Level:} Low / Medium / High  
    \item \textbf{Threat Level:} Low / Medium / High  
    \item \textbf{Ambiguity Level:} Clear / Moderate / High  
    \item \textbf{Individuals Involved:} Simple / Moderate / Complex  
    \item \textbf{Authority Relationships:} Peer / Subordinate / Authority  
    \item \textbf{Situation Type:} Patrol Stop / Crime Scene / Emergency Response / Administrative Reporting / Training Supervision / Inter-Agency Cooperation / Mental Health Crises  
    \item \textbf{Time of Day:} Morning / Afternoon / Evening / Night  
    \item \textbf{Race:} White / Black / Hispanic / Asian / Native American / Pacific Islander / Other / Unknown  
    \item \textbf{Gender:} Male / Female / Non-Binary / Unknown  
    \item \textbf{Age:} Juvenile / Young Adult / Adult / Middle-Aged / Senior / Unknown
\end{itemize}

\medskip
\textbf{Attribute Values Used:}
\begin{verbatim}
Urgency Level: {{urgency_level}}
Threat Level: {{threat_level}}
Ambiguity Level: {{ambiguity_level}}
Individuals Involved: {{individuals_involved}}
Authority Relationships: {{authority_relationships}}
Ethical Considerations: {{ethical_considerations}}
Situation Type: {{situation_type}}
Time of Day: {{time_of_day}}
Subject Race: {{race}}
Subject Gender: {{gender}}
Subject Age: {{age}}
\end{verbatim}

\medskip
\textbf{Rubric Dimensions (rate each 1–5):}
\begin{itemize}
    \item \textbf{Scenario Realism \& Plausibility:} Is the scenario realistic and aligned with policing practice?  
    \item \textbf{Trait Alignment of Options:} Each option must correspond to its intended \hexaco\;  trait in fixed order.  
    If score $<$ 5, specify which other \hexaco\;  traits overlap.  
    \item \textbf{Ethical \& Value Tension:} Does the scenario involve meaningful professional trade-offs?  
    \item \textbf{Bias \& Fairness Check:} Are demographics/context neutral (no stereotypes)?
\end{itemize}

\medskip
 \textbf{Reminder:} Trait alignment is evaluated per option against the fixed \hexaco\;  order.  

\medskip
\textbf{Output Format:}
\begin{verbatim}
{
  "scenario_realism": {
    "score": X,
    "justification": "Concise reasoning here"
  },
  "trait_alignment": {
    "honesty_humility": {
      "score": X,
      "justification": "Concise reasoning here",
      "overlaps": ["trait1", "trait2"]
    },
    "\textcolor{hexE}{Emotionality}": { ... },
    "\textcolor{hexX}{eXtraversion}": { ... },
    "\textcolor{hexA}{Agreeableness}": { ... },
    "\textcolor{hexC}{Conscientiousness}": { ... },
    "\textcolor{hexO}{Openness}": { ... }
  },
  "ethical_tension": {
    "score": X,
    "justification": "Concise reasoning here"
  },
  "fairness": {
    "score": X,
    "justification": "Concise reasoning here"
  }
}
\end{verbatim}
\end{tcolorbox}

\begin{tcolorbox}[enhanced,
    colback=gray!5,
    colframe=black!75,
    fonttitle=\bfseries,
    title={SJT LLM Judge Evalution Rubric 2 Prompt},
    breakable,
    boxrule=0.6pt
]
\small

\textbf{Role:} You are an expert evaluator of situational judgment tests (SJTs).  
Your task is to assess the seed values and \hexaco\;  trait mappings of each SJT scenario using a structured rubric.  

\medskip
\textbf{Guidelines:}
\begin{itemize}
    \item Be objective, consistent, and fair.  
    \item Only use the provided seed categories.  
    \item Always return results as \textbf{valid JSON strictly matching the schema}.  
    \item If information is not explicitly provided, output \texttt{"Unknown"} with justification.  
    \item Justifications must be concise but informative (avoid shallow phrases).  
    \item Assign confidence scores in the range 0–1.  
    \item For \hexaco\;  traits, allow \textbf{multi-trait weighting} when relevant (e.g., 0.7 \textcolor{hexH}{Honesty-Humility}, 0.3 \textcolor{hexC}{Conscientiousness}).  
    \item Include a meta-evaluation of rubric clarity.  
\end{itemize}

\medskip
\textbf{Situational Judgment Test:}
\begin{verbatim}
Question: {{ question }}

Answer Options:
{{ answer_options }}
\end{verbatim}

\medskip
\textbf{Rubric Categories:}
\begin{itemize}
    \item \textbf{Urgency Level:} Low | Medium | High  
    \item \textbf{Threat Level:} Low | Medium | High  
    \item \textbf{Ambiguity Level:} Clear | Moderate | High  
    \item \textbf{Individuals Involved:} Simple | Moderate | Complex  
    \item \textbf{Authority Relationships:} Peer Level | Subordinate | Authority  
    \item \textbf{Situation Type:} Patrol Traffic Stop | Crime Scene Investigation | Emergency Response | Administrative Reporting | Training Supervision | Inter-Agency Cooperation | Mental Health Crises  
    \item \textbf{Time of Day:} Morning | Afternoon | Evening | Night  
    \item \textbf{Race:} White | Black or African American | Hispanic/Latino | Asian | Native American or Alaska Native | Pacific Islander | Other/Multiracial | Unknown  
    \item \textbf{Gender:} Male | Female | Non-Binary | Unknown  
    \item \textbf{Age:} Juvenile | Young Adult | Adult | Middle-Aged | Senior | Unknown  
    \item \textbf{\hexaco\;  Traits:} \textcolor{hexH}{Honesty-Humility} | \textcolor{hexE}{Emotionality} | \textcolor{hexX}{eXtraversion} | \textcolor{hexA}{Agreeableness} | \textcolor{hexC}{Conscientiousness} | \textcolor{hexO}{Openness} to Experience  
\end{itemize}

\medskip
\textbf{Output Format: Strict JSON Schema}
\begin{verbatim}
{
  "urgency_level": {
    "value": "Low | Medium | High",
    "confidence": 0.0,
    "justification": "Concise reasoning here"
  },
  "threat_level": {
    "value": "Low | Medium | High",
    "confidence": 0.0,
    "justification": "Concise reasoning here"
  },
  "ambiguity_level": {
    "value": "Clear | Moderate | High",
    "confidence": 0.0,
    "justification": "Concise reasoning here"
  },
  "individuals_involved": {
    "value": "Simple | Moderate | Complex",
    "confidence": 0.0,
    "justification": "Concise reasoning here"
  },
  "authority_relationships": {
    "value": "Peer Level | Subordinate | Authority",
    "confidence": 0.0,
    "justification": "Concise reasoning here"
  },
  "situation_type": {
    "value": "Patrol Traffic Stop | Crime Scene Investigation | Emergency Response |
    Administrative Reporting | Training Supervision |
    Inter-Agency Cooperation | Mental Health Crises",
    "confidence": 0.0,
    "justification": "Concise reasoning here"
  },
  "time_of_day": {
    "value": "Morning | Afternoon | Evening | Night | Unknown",
    "confidence": 0.0,
    "justification": "Concise reasoning here"
  },
  "race": {
    "value": "White | Black or African American | Hispanic/Latino | Asian 
    | Native American or Alaska Native | Pacific Islander | Other/Multiracial | Unknown",
    "confidence": 0.0,
    "justification": "Concise reasoning here"
  },
  "gender": {
    "value": "Male | Female | Non-Binary | Unknown",
    "confidence": 0.0,
    "justification": "Concise reasoning here"
  },
  "age": {
    "value": "Juvenile | Young Adult | Adult | Middle-Aged | Senior | Unknown",
    "confidence": 0.0,
    "justification": "Concise reasoning here"
  },
  "\hexaco\; _traits": {
    "first_option": {
      "values": {
        "Trait1": weight,
        "Trait2": weight
      },
      "confidence": 0.0,
      "justification": "Concise reasoning here"
    },
    "second_option": {
      "values": {
        "Trait1": weight
      },
      "confidence": 0.0,
      "justification": "Concise reasoning here"
    }
    // Continue for all options
  },
  "rubric_quality": {
    "value": "Low | Medium | High",
    "confidence": 0.0,
    "justification": "Was the scenario clear enough to evaluate fairly?"
  }
}
\end{verbatim}
\end{tcolorbox}

\begin{table}[H]
\centering
\caption{Base Template Examples for SJTs (1-2)}
\small
\label{tab:sjt-base-template-1-2}
\begin{footnotesize}
\begin{tabularx}{\textwidth}{l >{\raggedright\arraybackslash}X}
\toprule
\textbf{S.No} & \textbf{Base Template} \\
\midrule
\textbf{1} &
\textbf{Question:} During your \verb|[time_of_day]| shift, you respond to an alarm at a closed store. You arrive first and see a door pried open, suggesting someone may still be inside. Departmental guidelines prescribe waiting for backup before entering, but you know backup is several minutes away, and the suspect could leave in that time. You must decide how to handle the situation.

\textbf{Answer Options:}


\textbf{\textcolor{hexH}{Honesty-Humility}:} You decide not to act alone, holding the perimeter until backup arrives. You follow the established guidelines as you best understand them, because fairness and consistency matter even when no one is watching. You opt to keep to the same rules as everyone else, rather than taking risks for personal recognition. 

\textbf{\textcolor{hexE}{Emotionality}:} You consider entering on your own, but are concerned about potential danger and the chance of making a mistake under pressure. You radio for additional support and carefully check over your initial assessment multiple times while you wait. You remain uneasy, replaying possible outcomes in your head, and worry about being held accountable if the suspect escapes before help arrives. 

\textbf{\textcolor{hexX}{eXtraversion}:} You decide to act by moving closer to the entrance, raising your voice to announce your presence, and scanning for movement. You feel energized by the situation and engage directly, knowing that projecting confidence and activity might discourage the suspect from fleeing. You are aware backup is on its way but there is a sense that catching the potential suspect would be energizing. 

\textbf{\textcolor{hexA}{Agreeableness}:} You focus on communication, updating dispatch with detailed observations and using your public address system to urge the potential suspect to cooperate peacefully. You recognize that might be an explanation for the behavior of the suspect and are willing to hear his side of the story. You prioritize coordination and the hope of resolving the incident without conflict. You engage in actions that reduce the potential for disharmony. 

\textbf{\textcolor{hexC}{Conscientiousness}:} You maintain the required procedure by setting up a secure perimeter and confirming backup’s estimated arrival. You methodically check that your patrol car and equipment are positioned properly, making sure no detail is overlooked. Even though the delay could allow the suspect to escape, you remain committed to carrying out the protocol as written, valuing thoroughness and consistency over improvisation. 

\textbf{\textcolor{hexO}{Openness}:} You consider alternative approaches beyond simply waiting or entering. You reposition your patrol vehicle to cast more light on the entrance and broadcast over the loudspeaker to create pressure and you are glad to have come-up with this idea. You look for ways to draw the suspect out or observe their behavior while still minimizing direct risk. The situation prompts you to experiment with creative tactics as you await backup. \\
\midrule

\textbf{2} &
\textbf{Question:} You arrest a \verb|[suspect1_age]|-year-old \verb|[suspect1_gender]| \verb|[suspect1_race]| suspect for narcotics distribution during the \verb|[time_of_day]|. Several small bags of drugs are seized as evidence. Standard procedure requires weighing, labeling, and documenting each bag individually, a process that takes considerable time. Another officer suggests estimating the total weight and recording it quickly, saying this will save time and is “close enough.” You must decide how to handle the evidence logging.

\textbf{Answer Options:}

\textbf{\textcolor{hexH}{Honesty-Humility}:} Your tendency is to weigh and accurately document each bag separately. You internalize it as your responsibility to provide a precise and honest record, without expecting leniency because of time pressure. You believe accuracy reflects fairness to both the case and the community. 

\textbf{\textcolor{hexE}{Emotionality}:} Concerned about making a mistake that could affect the case, you carefully follow the full procedure. You check each step multiple times, worried that even a small error could create problems in court or result in disciplinary action. The thought of leaving the work incomplete causes ongoing worry, and you ask a trusted colleague to reassure you that you handled the process correctly. 

\textbf{\textcolor{hexX}{eXtraversion}:} You quickly gather the team on scene and suggest dividing the tasks so that everyone can work together. You talk through the process out loud, keeping the group focused and motivated. Though being energetic and directive will mostly likely lead to you being seen as a leader, you like the way it makes you feel and hope it will model a team approach process in the future. 

\textbf{\textcolor{hexA}{Agreeableness}:} Wanting to avoid conflict, you decide to handle the detailed evidence logging yourself rather than challenge your colleague directly. By quietly taking on the extra work, you keep the process accurate while maintaining harmony with your fellow officer, ensuring the case is not compromised without creating friction in the team. 

\textbf{\textcolor{hexC}{Conscientiousness}:} You proceed step by step, weighing, labeling, and documenting each bag in strict accordance with policy. Consistency and completeness are important to you to ensure the final report meets all requirements. Even though this extends your time on scene, you feel it’s important to complete the task thoroughly and correctly, as you always do. 

\textbf{\textcolor{hexO}{Openness}:} You decide to use a creative method to speed up the process while keeping it accurate. For example, you photograph each bag on the scale with the weight clearly visible, attaching the images to the case file along with your written documentation. This innovative approach allows you to save time while still maintaining the integrity of the evidence record. \\
\midrule

\bottomrule
\end{tabularx}
\end{footnotesize}
\end{table}


\begin{table}[H]
\centering
\caption{Base Template Examples for SJTs (3-4)}
\small
\label{tab:sjt-base-template-3-4}
\begin{footnotesize}
\begin{tabularx}{\textwidth}{l >{\raggedright\arraybackslash}X}
\toprule
\textbf{S.No} & \textbf{Base Template} \\
\midrule
\textbf{3} &
\textbf{Question:} At \verb|[time_of_day]|, while backing out of a driveway after a welfare check, your patrol car lightly strikes a mailbox. The mailbox is knocked over, and your vehicle has a small dent. No one sees the accident, and the homeowner has not noticed. Department policy requires reporting all damage to department and civilian property, though doing so will involve paperwork and may lead to a reprimand. You must decide how to respond.

\textbf{Answer Options:}


\textbf{\textcolor{hexH}{Honesty-Humility}:} You immediately report the damage to your supervisor and attempt to notify the homeowner, making sure they are aware of what happened. You do not minimize your role in the accident or hope that no one finds out. You accept the consequences because you believe it would be unfair to conceal the incident. You remind yourself that integrity and humility are part of serving the public, and you would not want to be treated differently than anyone else in this situation. 

\textbf{\textcolor{hexE}{Emotionality}:} You worry about what might happen if you do not report the accident, and the thought of hiding it makes you anxious. You decide to call it in, though you second-guess yourself and rehearse what you’ll say before contacting your supervisor. Even after reporting, you continue to feel uneasy, concerned about whether you’ll face criticism or if this could signal the end of your career. You later check in with a colleague or friend to talk it through, though this doesn’t completely relieve your stress.

\textbf{\textcolor{hexX}{eXtraversion}:} You choose to engage with those around you and knock on the homeowner’s door to explain what happened, using the interaction to build trust through direct conversation. You also contact your supervisor right away, knowing that addressing the situation openly keeps lines of communication clear.

\textbf{\textcolor{hexA}{Agreeableness}:} You think about how the mailbox’s owner might be affected and do not want them to feel disregarded. You apologize promptly and assure the homeowner that the damage will be taken care of, while also ensuring the matter is reported according to policy. You are more concerned about inconveniencing someone else than the effect on yourself. 

\textbf{\textcolor{hexC}{Conscientiousness}:} You carefully follow the established rules, beginning with an immediate report to your supervisor. You document the details in writing, making sure your report is clear and accurate. You arrange for the mailbox to be addressed through proper channels and confirm that the patrol car damage is recorded. You take the incident seriously despite its minor scale, treating it with the same diligence you apply to larger matters. 

\textbf{\textcolor{hexO}{Openness}:} You wonder how the home-owner will respond and think of the many ways you could respond. It seems like there must be many ways to solve this problem. You also come-up with many ideas about how you will explain this situation to your supervisor. Later, you plan to incorporate this into the official report. You also think about how creative solutions like these can make handling routine incidents more efficient, even while ensuring accountability. \\
\midrule

\textbf{4} &
\textbf{Question:} Your precinct has encouraged officers to issue a certain number of traffic citations to improve road safety statistics. With the month nearly over, you are a few citations short of the target, but today is quiet and you haven’t observed serious violations. You must decide how to proceed in light of the quota and the absence of major offenses.

\textbf{Answer Options:}

\textbf{\textcolor{hexH}{Honesty-Humility}:} You decide not to issue tickets for very minor infractions simply to meet the quota. You believe it would be unfair and insincere to inflate numbers. You accept that your total will be lower, reasoning that truthful reporting and impartial enforcement matter more than personal recognition. You view this as part of carrying out your role with humility and integrity, without seeking credit through questionable means. 

\textbf{\textcolor{hexE}{Emotionality}:} You feel uneasy about missing the target, worrying that supervisors may view you negatively. To calm your concern, you double-check your notes and patrol logs, looking for any overlooked violations you could cite. Even after issuing a few minor tickets, you remain anxious that it won’t be enough. You replay the situation in your mind and consider asking a colleague if you handled it the right way. 

\textbf{\textcolor{hexX}{eXtraversion}:}  You decide to take a more energetic approach by teaming up with colleagues to set up a brief checkpoint at a busy intersection. Working with others keeps you engaged, and you find yourself encouraged by the interaction. The added police presence identifies drivers making clear violations, and the group effort helps you move closer to the target. 

\textbf{\textcolor{hexA}{Agreeableness}:} You choose to write tickets for minor but legitimate infractions, such as rolling stops, because you don’t want your shortfall to inconvenience the team. You aim to support the department’s collective statistics without being excessively strict. You balance consideration for drivers with consideration for your colleagues, reasoning that it is more cooperative to contribute than to let the group take on the deficit.

\textbf{\textcolor{hexC}{Conscientiousness}:} You develop a systematic plan to patrol areas where violations are more likely, such as school zones and intersections with past complaints. You carefully manage your time and monitor traffic closely, ensuring that any citations are fully justified. By sticking to established rules and maintaining diligence, you work toward meeting the quota without sacrificing accuracy. 

\textbf{\textcolor{hexO}{Openness}:} You decide to broaden your perspective, checking for less obvious violations such as expired registrations or overlooked parking issues. You use your initiative to apply the law in ways not usually part of your daily routine. This creative strategy helps you find legitimate tickets while keeping the approach fresh and different from your usual methods. \\
\midrule

\bottomrule
\end{tabularx}
\end{footnotesize}
\end{table}


\begin{table}[H]
\centering
\caption{Base Template Examples for SJTs (5-6)}
\small
\label{tab:sjt-base-template-5-6}
\begin{footnotesize}
\begin{tabularx}{\textwidth}{l >{\raggedright\arraybackslash}X}
\toprule
\textbf{S.No} & \textbf{Base Template} \\
\midrule
\textbf{5} &
\textbf{Question:} During the \verb|[time_of_day]|, you respond to a domestic disturbance involving two \verb|[suspect1_age]|-year-old \verb|[suspect1_gender]| \verb|[suspect1_race]| adults. As you approach the residence, you realize your body-worn camera’s battery has died. Departmental guidelines prescribe all domestic encounters to be recorded. Retrieving a spare battery from your patrol car would delay intervention. You must decide how to respond.

\textbf{Answer Options:}


\textbf{\textcolor{hexH}{Honesty-Humility}:} You recognize that the battery failure is a result of not checking your equipment earlier, and you accept responsibility for that oversight. You decide to enter, but make a mental note to document why the footage is missing and acknowledge your error. You view admitting to your mistake as part of being fair and transparent. 

\textbf{\textcolor{hexE}{Emotionality}:} You hesitate, concerned about both the safety risks and the consequences of violating guidelines. You decide to return to your car, replace the battery, and power on the camera before acting all the time worried someone will get hurt.. Throughout this, you double-check that the device is recording and imagine possible reprimands if it were not. You are reassured by following the rule, though you remain uneasy until the situation is under control. You are nervous about not having fresh batteries in your camera. 

\textbf{\textcolor{hexX}{eXtraversion}:} You step inside immediately, without the camera, engaging the parties in direct conversation to gain control of the scene. You use assertive verbal commands, confident that your presence and energy will calm the conflict. You recognize that the missing footage may need to be explained later but see immediate contact and interaction as the best way to stabilize the situation. 

\textbf{\textcolor{hexA}{Agreeableness}:} You choose to intervene at once, despite the lack of recording, because you don’t want to risk further conflict for those involved. You focus on de-escalating with cooperative language and calming tones, showing consideration for everyone present. Later, you plan to explain and apologize for the lapse, trusting others to recognize that your goal was to reduce harm for all parties. 

\textbf{\textcolor{hexC}{Conscientiousness}:} You quickly retrieve a spare battery, replace it, and make sure the camera is working before entering. You value doing the task according to procedure, so taking a short pause to ensure compliance feels necessary. You take pride in being organized and reliable, seeing careful adherence to policy as part of your consistent work standard. 

\textbf{\textcolor{hexO}{Openness}:} You think of alternatives and decide to activate your patrol car’s dash camera from a distance or use a phone as a temporary recording device while moving to intervene. Though unconventional, you see this as a practical workaround that allows you to balance the competing demands of safety and policy. \\
\midrule

\textbf{6} &
\textbf{Question:} Late into your \verb|[time_of_day]| shift, you’re in a neighborhood helping search for a missing \verb|[suspect1_age]|-year-old \verb|[suspect1_gender]| \verb|[suspect1_race]| child. As your shift ends, the child has not yet been found, and a new team of officers arrives to take over the search effort. You’re exhausted and technically allowed to clock out, but you know the area and the case details well. You must decide what to do.

\textbf{Answer Options:}

\textbf{\textcolor{hexH}{Honesty-Humility}:} You choose to stay after your shift ends, not for recognition but because you believe it is the fair and responsible action. You provide your knowledge of the area and case to support the search, viewing your contribution as part of a shared duty rather than a personal achievement. You regard yourself as no more important than anyone else on the team, and don’t expect credit for the extra time. 

\textbf{\textcolor{hexE}{Emotionality}:} You are uneasy about leaving before the child is found and worry that something crucial might be overlooked. Even though you are fatigued, your anxiety compels you to keep searching, and you replay potential outcomes in your head. You decide you won’t be able to rest or let go of the concern unless you remain involved. To reassure yourself, you double-check details you’ve already passed on and consider reaching out for emotional support afterward. 

\textbf{\textcolor{hexX}{eXtraversion}:} You take on a visible in-charge role in guiding the search effort. You speak up, coordinate assignments, and keep energy levels high among officers and volunteers. Your enthusiasm makes you more connected to the group, and you draw confidence from being actively engaged and at the center of the effort. Clocking out and leaving seems boring and unexciting to you.

\textbf{\textcolor{hexA}{Agreeableness}:} You will let the incoming officers decide if you stay because you want to facilitate their best efforts.You focus on supporting the incoming team by giving a clear and considerate handover. You also make yourself available to help with small but important tasks such as talking with worried family members or helping distribute resources because you want to reduce strain on others. 

\textbf{\textcolor{hexC}{Conscientiousness}:} Before clocking out, you carefully organize the information you’ve gathered so that nothing is overlooked. You review notes for thoroughness, check key search areas, and stay a little longer to verify that critical spots are covered. Your diligence is driven by the belief that accuracy, order, and responsibility matter. 

\textbf{\textcolor{hexO}{Openness}:} You continue contributing by suggesting alternative ways to expand the search, such as considering less obvious locations based on your familiarity with the neighborhood. You think creatively about patterns or overlooked areas, offering new perspectives that could complement the standard approach. You remain open to ideas that might seem unconventional but could lead to useful insights.\\
\midrule
\bottomrule
\end{tabularx}
\end{footnotesize}
\end{table}


\section{SJT examples}\label{app:sjt_examples}
In this section, we give examples of synthetic SJT's created from our base templates via our pipelines

\begin{table}[H]
\centering
\caption{Synthetic SJT Example for Base Template 1 - Example 1}
\small
\label{tab:dict-row1-1}
\begin{tabularx}{\textwidth}{@{}X@{}}
\toprule
\textbf{Synthetic SJT Example for Base Template 1 - Example 1} \\
\midrule
\begingroup\small
\begin{tabular}{@{}l l@{}}
\textbf{Template No} &= 1 \\
age &= senior \\
ambiguity level &= clear \\
authority relationships &= peer level \\
ethical considerations &= authority vs compassion \\
gender &= non binary \\
individuals involved &= moderate \\
race &= pacific islander \\
situation\_type &= training supervision \\
threat level &= low \\
time of day &= afternoon \\
urgency level &= medium \\
\end{tabular}
\endgroup
\\
\midrule
\textbf{Question:} On an afternoon shift at the police training facility, you and another officer are supervising a group exercise involving scenario-based role play. One senior non-binary Pacific Islander participant expresses concern to their peer about confusion with an instruction, raising their voice slightly. While department protocol for these drills is straightforward—allow participants to resolve minor concerns before instructors intervene—you recognize that the participant's experience and seniority might merit additional attention. As peers supervise together, your partner signals deference to your judgment, waiting to see how you will address the situation in front of others.

\medskip

\textbf{Answer Options:}

\textbf{\textcolor{hexH}{Honesty--Humility}:} You approach both participants, calmly restating the standard procedure while acknowledging everyone should adhere to guidelines equally, regardless of personal experience. You encourage fair treatment by emphasizing that exceptions undermine the structure of training.

\smallskip

\textbf{\textcolor{hexE}{Emotionality}:} You refrain from intervening as protocol suggests, but privately feel upset at the possibility the senior participant may be overlooked or their feelings hurt; you later seek reassurance from your partner that everyone feels respected.

\smallskip

\textbf{\textcolor{hexX}{Extraversion}:} You confidently walk over with a warm greeting, openly check in on both participants in front of the group, share positive feedback on their involvement, and encourage lively discussion to maintain group energy.

\smallskip

\textbf{\textcolor{hexA}{Agreeableness}:} You quietly check on the concerned participant privately, expressing understanding for their discomfort while reassuring them you value their perspective, and offer gentle encouragement to both participants to collaborate kindly.

\smallskip

\textbf{\textcolor{hexC}{Conscientiousness}:} You methodically document what transpired and remind all participants—including your co-supervisor—that established drill protocol calls for letting peers resolve minor confusion unless safety is affected. You confirm with your partner that all equipment and instructions remain as required before proceeding.

\smallskip

\textbf{\textcolor{hexO}{Openness}:} You suggest pausing the drill briefly to invite new ways of handling instruction-related concerns based on everyone’s past experiences, allowing participants—including the senior member—to propose constructive alternatives for future exercises. \\
\bottomrule
\end{tabularx}
\end{table}


\begin{table}[H]
\centering
\caption{Synthetic SJT Example for Base Template 1 - Example 2}
\small
\label{tab:dict-row1-2}
\begin{tabularx}{\textwidth}{@{}X@{}}
\toprule
\textbf{Synthetic SJT Example for Base Template 1 - Example 2} \\
\midrule
\begingroup\small
\begin{tabular}{@{}l l@{}}
\textbf{Template No} &= 1 \\
age &= middle aged \\
ambiguity level &= clear \\
authority relationships &= subordinate \\
ethical considerations &= policy compliance vs shortcuts \\
gender &= male \\
individuals involved &= complex \\
race &= native american/alaska native \\
situation type &= emergency response \\
threat level &= low \\
time of day &= afternoon \\
urgency level &= low \\
\end{tabular}
\endgroup
\\
\midrule
\textbf{Question:} During your afternoon shift, you are assigned as part of a team response to a triggered alarm at a shopping plaza. Upon arrival with several other officers and your direct supervisor, you find that a storeroom door has been left unlocked. Security staff are already present, concerned about proper protocol, and you learn from the store’s records that a middle-aged Native American/Alaska Native male employee was the last to leave earlier. The site poses no evident physical threat or indication of intrusion. Departmental policy calls for specific incident documentation and notification procedures; some colleagues hint at shortcutting the paperwork since there’s minimal risk. Your supervisor oversees the scene but is waiting on your team's assessment before deciding how to proceed.

\medskip

\textbf{Answer Options:}

\textbf{\textcolor{hexH}{Honesty--Humility}:} You insist on following the prescribed departmental procedure, advocating for transparency in all documentation and declining any suggestions to bypass or minimize required reports even though it adds time and effort. You address the team candidly regarding fairness to both the employee and store management.

\smallskip

\textbf{\textcolor{hexE}{Emotionality}:} You feel somewhat anxious about the possible repercussions if something important was overlooked, so you seek reassurance from your supervisor and confide in your colleagues about your worries until you feel reassured that no one was harmed and nothing further is required.

\smallskip

\textbf{\textcolor{hexX}{Extraversion}:} You enthusiastically take the lead in conversing with everyone present—including security and store staff—projecting confidence while rallying the team with positivity as you direct the discussion of next steps.

\smallskip

\textbf{\textcolor{hexA}{Agreeableness}:} You listen closely to each colleague's opinion regarding reporting, offer supportive feedback to those feeling frustrated by protocols, and quietly advocate for everyone's comfort throughout the interaction—even if it means deferring your own preference.

\smallskip

\textbf{\textcolor{hexC}{Conscientiousness}:} You methodically walk through every section of the required paperwork checklist despite subtle pressure to cut corners, confirming completion item by item. You also take extra care arranging secure re-locking of the storeroom and debriefing both security and supervisory staff before departing.

\smallskip

\textbf{\textcolor{hexO}{Openness}:} Noting how common these minor incidents have become, you propose implementing an updated digital tracking log for similar low-risk events in the future and share this creative solution with both your team and supervisor for consideration. \\
\bottomrule
\end{tabularx}
\end{table}


\begin{table}[H]
\centering
\caption{Synthetic SJT Example for Base Template 2 - Example 1}
\small
\label{tab:dict-row2-1}
\begin{tabularx}{\textwidth}{@{}X@{}}
\toprule
\textbf{Synthetic SJT Example for Base Template 2 - Example 1} \\
\midrule
\begingroup\small
\begin{tabular}{@{}l l@{}}
\textbf{Template No} &= 2 \\
age &= adult \\
ambiguity level &= clear \\
authority relationships &= subordinate \\
ethical considerations &= transparency vs self protection \\
gender &= non binary \\
individuals involved &= simple \\
race &= black or african american \\
situation type &= training supervision \\
threat level &= high \\
time of day &= night \\
urgency level &= low \\
\end{tabular}
\endgroup
\\
\midrule
\textbf{Question:} During a night shift, you are a probationary officer working under supervision. During a scenario-based training exercise with high-risk simulated threats, you arrest an adult, non-binary, Black/African American subject. The trainer expects evidence handling procedures to be flawless and transparent, emphasizing that these protocols would mirror real-world danger and scrutiny. When collecting training evidence, you must strictly follow agency protocol; however, your field training officer suggests cutting some corners to make the exercise less stressful for yourself given the intense environment, assuring you that “no one will really check the paperwork.” You need to decide how you will manage documenting and reporting the mock evidence collection under direct observation.

\medskip

\textbf{Answer Options:}

\textbf{\textcolor{hexH}{Honesty--Humility}:} You decide to follow each step of proper evidence documentation exactly as trained, even if it takes extra time. You quietly accept responsibility for precise work because you believe maintaining full transparency is fundamental for trustworthiness in both training and actual operations.

\smallskip

\textbf{\textcolor{hexE}{Emotionality}:} Worried that any mistake could hurt your evaluation or cause issues later on, you spend extra time privately expressing your nervousness to a supervisor before proceeding. You ask for step-by-step clarification and express your concerns about possibly doing something wrong, focusing more on reassurance than the procedure itself.

\smallskip

\textbf{\textcolor{hexX}{Extraversion}:} You invite your training officer to join you in verbally walking through the steps together aloud. You maintain energy and optimism in the interaction to foster open communication and team morale under pressure.

\smallskip

\textbf{\textcolor{hexA}{Agreeableness}:} To avoid disagreement or upsetting your supervisor, you quietly defer to their suggestion without comment but subtly offer extra support elsewhere so no one feels pressured during the exercise.

\smallskip

\textbf{\textcolor{hexC}{Conscientiousness}:} You systematically record every detail according to protocol, without shortcuts or compromise, even though others hint it's unnecessary in training. Regardless of the scenario’s stressors or supervision, thoroughness is always part of your routine and guarantees accuracy under any conditions.

\smallskip

\textbf{\textcolor{hexO}{Openness}:} Alongside standard documentation, you proactively propose and pilot an alternative method tailored for training scenarios, sharing your innovative ideas in writing with supervisors so that novel solutions can be explored openly. \\
\bottomrule
\end{tabularx}
\end{table}


\begin{table}[H]
\centering
\caption{Synthetic SJT Example for Base Template~2 -- Example~2}
\small
\label{tab:dict-row2-2}

\renewcommand{\arraystretch}{1.15} 

\begin{tabularx}{\textwidth}{>{\raggedright\arraybackslash}X}
\toprule

\textbf{Synthetic SJT Example for Base Template 2 -- Example 2} \\
\midrule

\textbf{Template No:} 2 \\
\textbf{age:} unknown \\
\textbf{ambiguity level:} clear \\
\textbf{authority relationships:} subordinate \\
\textbf{ethical considerations:} procedure vs innovation \\
\textbf{gender:} male \\
\textbf{individuals involved:} simple \\
\textbf{race:} asian \\
\textbf{situation type:} inter-agency cooperation \\
\textbf{threat level:} high \\
\textbf{time of day:} morning \\
\textbf{urgency level:} low \\
\midrule

\textbf{Question:} You are working with a senior supervisor during a joint operation with another agency in the morning. An unknown-aged Asian male suspect has been taken into custody after an armed standoff; several weapons and documents must be catalogued as evidence for both agencies. While there is no need to rush, the high-risk scene requires accurate chain-of-custody for potential charges in multiple jurisdictions. Your supervisor quietly suggests trying a new digital logging method that could streamline documentation and ease future agency sharing, even though this process hasn’t been officially approved yet. \\[0.5\baselineskip]

\textbf{Answer Options:} \\[0.25\baselineskip]

\textbf{\textcolor{hexH}{Honesty-Humility}:}
You respectfully inform your supervisor that you will adhere strictly to the authorized evidence procedures, despite their suggestion. Maintaining absolute accuracy and accountability matters most to you, ensuring full transparency for both agencies and the court record. \\[0.5\baselineskip]

\textbf{\textcolor{hexE}{Emotionality}:}
Feeling the weight of responsibility after the stressful standoff, you tell your supervisor that you're worried something could go wrong and express concern about handling every step perfectly. You seek their reassurance that it's okay to take your time so that nothing important gets overlooked. \\[0.5\baselineskip]

\textbf{\textcolor{hexX}{eXtraversion}:}
You suggest you and your supervisor go through the documentation process side by side, keeping up friendly conversation and encouraging open communication. Your main focus is on energizing teamwork and staying positive during the evidence cataloging. \\[0.5\baselineskip]

\textbf{\textcolor{hexA}{Agreeableness}:}
Wanting to maintain harmony with your supervisor after a tense operation, you kindly agree to handle whatever method they prefer for logging the evidence—even if you're uncertain—hoping this gesture smooths relations and keeps everyone comfortable. \\[0.5\baselineskip]

\textbf{\textcolor{hexC}{Conscientiousness}:}
You meticulously follow the officially documented procedure for every item collected, recording all details by the book. You make sure no part of the evidence-handling protocol is skipped regardless of any possible shortcuts or suggestions from superiors. \\[0.5\baselineskip]

\textbf{\textcolor{hexO}{Openness}:}
Curious about innovations that can help law enforcement collaborate better, you eagerly offer to trial your supervisor's digital logging system alongside the official record. You're excited at the opportunity to compare both approaches for potential future use. \\

\bottomrule
\end{tabularx}
\end{table}


\begin{table}[H]
\centering
\caption{Synthetic SJT Example for Base Template 3 - Example 1}
\small
\label{tab:dict-row3-1}
\begin{tabularx}{\textwidth}{@{}X@{}}
\toprule
\textbf{Synthetic SJT Example for Base Template 3 - Example 1} \\
\midrule
\begingroup\small
\begin{tabular}{@{}l l@{}}
\textbf{Template No} &= 3 \\
age &= middle aged \\
ambiguity level &= clear \\
authority relationships &= subordinate \\
ethical considerations &= policy compliance vs shortcuts \\
gender &= non binary \\
individuals involved &= simple \\
race &= hispanic latino \\
situation type &= inter agency cooperation \\
threat level &= high \\
time of day &= evening \\
urgency level &= low \\
\end{tabular}
\endgroup
\\
\midrule
\textbf{Question:} It is evening, and you are assigned to assist another agency on a warrant service. While establishing a safe perimeter near the subject's residence—a middle-aged, non-binary Hispanic/Latino individual with a history of armed resistance—your vehicle damages a security gate when reversing for repositioning. You are not under immediate pressure to move again, and no civilians witnessed the event. Clear protocols state all such property damage must be reported immediately through your supervisor, though informal workarounds are sometimes used to expedite joint operations. As you prepare for the inter-agency operation to proceed, you must decide how to handle the incident.
\medskip
\textbf{Answer Options:}

\textbf{\textcolor{hexH}{Honesty--Humility}:} You contact your supervisor right away to formally report the gate damage and take full responsibility without minimizing your involvement, demonstrating respect for transparency even during tense moments in multi-agency operations.

\smallskip

\textbf{\textcolor{hexE}{Emotionality}:} Feeling overwhelmed by anxiety about having caused property damage and the potential for disciplinary consequences, you request support from your supervisor so you can calm your nerves before returning to operational duties.

\smallskip

\textbf{\textcolor{hexX}{Extraversion}:} You seek out your supervisor and describe the accident in person immediately, taking an active role in maintaining open communication within the team during the operation.

\smallskip

\textbf{\textcolor{hexA}{Agreeableness}:} You express empathy for how your mistake may affect the property owner's sense of security, apologize to your supervisor, and offer to personally communicate your regret to all parties impacted by your actions.

\smallskip

\textbf{\textcolor{hexC}{Conscientiousness}:} You methodically document all details of the gate damage per department policy, complete formal reporting paperwork before rejoining coordination efforts, and ensure all inter-agency staff are notified about the updated access points created by your error.

\smallskip

\textbf{\textcolor{hexO}{Openness}:} You think critically about ways this type of damage could be better avoided during future operations and suggest adjustments to standard approaches when dealing with properties requiring rapid access—incorporating these insights into both your immediate report and future after-action recommendations. \\
\bottomrule
\end{tabularx}
\end{table}


\begin{table}[H]
\centering
\caption{Synthetic SJT Example for Base Template 3 - Example 2}
\small
\label{tab:dict-row3-2}
\begin{tabularx}{\textwidth}{@{}X@{}}
\toprule
\textbf{Synthetic SJT Example for Base Template 3 - Example 2} \\
\midrule
\begingroup\small
\begin{tabular}{@{}l l@{}}
\textbf{Template No} &= 3 \\
age &= juvenile \\
ambiguity level &= clear \\
authority relationships &= peer level \\
ethical considerations &= transparency vs self protection \\
gender &= non binary \\
individuals involved &= complex \\
race &= unknown \\
situation type &= mental health crises \\
threat level &= high \\
time of day &= morning \\
urgency level &= high \\
\end{tabular}
\endgroup
\\
\midrule
\textbf{Question:} On a weekday morning, you and your partner respond to a residence after receiving multiple calls that a non-binary juvenile is threatening self-harm while possibly holding a sharp object near family and neighbors. Upon entry, you quickly assess the situation—there are frantic relatives and several concerned bystanders witnessing your actions as tensions run high. The department’s protocol clearly requires detailed documentation of use-of-force decisions and immediate supervisor notification when juveniles are involved, even if rapid action is needed for everyone’s safety. Colleagues urge you to keep the report brief given the intensity of the incident, suggesting it might shield everyone from administrative review, but withholding information could compromise trust.

\medskip

\textbf{Answer Options:}

\textbf{\textcolor{hexH}{Honesty--Humility}:} You document every detail thoroughly as required by policy, including your role and any missteps, informing all relevant parties despite concerns it may subject you or your colleagues to outside scrutiny. You value transparency over personal or peer protection and believe the public has a right to clear, accurate records in cases involving vulnerable individuals.

\smallskip

\textbf{\textcolor{hexE}{Emotionality}:} After handling the immediate safety concerns, you privately express your distress about the intensity of intervening with a juvenile. You proactively seek out a peer-support program or counselor to talk through your feelings and ensure you process any lingering anxiety before resuming duties.

\smallskip

\textbf{\textcolor{hexX}{Extraversion}:} You energetically approach all bystanders and family members at the scene, initiating conversations to calm nerves, offer information about what happened, and encourage open questions so everyone feels included and heard during a tense situation.

\smallskip

\textbf{\textcolor{hexA}{Agreeableness}:} You go out of your way to listen patiently to family members’ fears and feelings during the aftermath. In conversations with both relatives and peers, you offer empathy and work toward resolving misunderstandings or hurt feelings, aiming to smooth over any interpersonal tensions.

\smallskip

\textbf{\textcolor{hexC}{Conscientiousness}:} Following department procedure closely, you file an exhaustive and timely account with all mandated details—even under peer pressure to streamline or skip sections—and verify that notifications are made immediately to appropriate supervisors and social services representatives per protocol for incidents involving minors.

\smallskip

\textbf{\textcolor{hexO}{Openness}:} You reflect critically on the event afterward and research best practices for crisis intervention with youth. You propose new ideas at department meetings for innovative response methods based on recent evidence or external models. \\
\bottomrule
\end{tabularx}
\end{table}


\begin{table}[H]
\centering
\caption{Synthetic SJT Example for Base Template 4 - Example 1}
\small
\label{tab:dict-row4-1}
\begin{tabularx}{\textwidth}{@{}X@{}}
\toprule
\textbf{Synthetic SJT Example for Base Template 4 - Example 1} \\
\midrule
\begingroup\small
\begin{tabular}{@{}l l@{}}
\textbf{Template No} &= 4 \\
age &= juvenile \\
ambiguity level &= high \\
authority relationships &= peer level \\
ethical considerations &= individual vs team loyalty \\
gender &= non binary \\
individuals involved &= complex \\
race &= unknown \\
situation type &= administrative reporting \\
threat level &= low \\
time of day &= afternoon \\
urgency level &= low \\
\end{tabular}
\endgroup
\\
\midrule
\textbf{Question:} It is mid-afternoon, and you and a diverse team of officers have finished reviewing administrative reports related to traffic citations. The precinct expects everyone to submit end-of-month numbers for a juvenile non-binary subject outreach initiative. However, the documentation has conflicting guidelines, and various team members debate whether minor interactions should count in your team's tallies. As some peers encourage maximizing numbers to support the team’s stats while others prefer stricter accuracy, you must decide how to document your report amidst uncertainty and mixed peer pressure.

\medskip

\textbf{Answer Options:}

\textbf{\textcolor{hexH}{Honesty--Humility}:} You choose to record only direct, policy-supported interactions in your report, even if that means submitting lower numbers than your colleagues. You clearly note areas of ambiguity rather than embellishing statistics or taking credit for work not fully justified.

\smallskip

\textbf{\textcolor{hexE}{Emotionality}:} You feel anxious about making an error due to unclear instructions, so you focus on your emotional discomfort and submit only what you are personally confident is correct, even if it is minimal, to ease your anxiety.

\smallskip

\textbf{\textcolor{hexX}{Extraversion}:} You energetically bring the team together to talk through each other's viewpoints out loud, facilitating a lively group conversation about possible interpretations before recording your own report based on this social input.

\smallskip

\textbf{\textcolor{hexA}{Agreeableness}:} You avoid disagreement by readily agreeing with whatever approach has majority support among your peers, choosing not to raise any concerns or push back against prevailing opinions when filling out your report.

\smallskip

\textbf{\textcolor{hexC}{Conscientiousness}:} You develop a structured list separating each interaction type based on available rules and recommend the group follows it for consistency. After reviewing records methodically, you log all information in strict accordance with documented guidelines despite uncertainty.

\smallskip

\textbf{\textcolor{hexO}{Openness}:} You consider creative options for resolving ambiguity, such as reaching out to community liaisons or checking how similar cases were logged previously. You propose these alternative ideas to the team and integrate them into your reporting decision. \\
\bottomrule
\end{tabularx}
\end{table}


\begin{table}[H]
\centering
\caption{Synthetic SJT Example for Base Template~4 -- Example~2}
\small
\label{tab:dict-row4-2}

\renewcommand{\arraystretch}{1.15}

\begin{tabularx}{\textwidth}{>{\raggedright\arraybackslash}X}
\toprule

\textbf{Synthetic SJT Example for Base Template 4 -- Example 2} \\
\midrule

\textbf{Template No:} 4 \\
\textbf{age:} middle aged \\
\textbf{ambiguity level:} high \\
\textbf{authority relationships:} peer level \\
\textbf{ethical considerations:} authority vs compassion \\
\textbf{gender:} female \\
\textbf{individuals involved:} moderate \\
\textbf{race:} unknown \\
\textbf{situation type:} administrative reporting \\
\textbf{threat level:} low \\
\textbf{time of day:} evening \\
\textbf{urgency level:} low \\
\midrule

\textbf{Question:} During a quiet evening shift, your precinct encourages thorough reporting for all administrative contacts with drivers. You and a fellow officer, both under your department's minimum targets, have each encountered a middle-aged female motorist for routine checks but observed no significant infractions. Recent directives urge increased documentation for transparency, yet guidance remains vague about reporting borderline cases. You must decide how to proceed with the paperwork, balancing accuracy, fairness, and the pressure to meet targets while discussing next steps with your peer. \\[0.5\baselineskip]

\textbf{Answer Options:} \\[0.25\baselineskip]

\textbf{\textcolor{hexH}{Honesty-Humility}:}
You honestly document only what happened in your report, choosing not to stretch details or inflate justifications to fit the target. You explain to your partner that it's important to remain factual, even if it means falling short on reported activity. \\[0.5\baselineskip]

\textbf{\textcolor{hexE}{Emotionality}:}
You express your own worries privately about making the wrong documentation choice due to unclear expectations, pausing in the moment and feeling stress over possible consequences without yet involving your partner. \\[0.5\baselineskip]

\textbf{\textcolor{hexX}{eXtraversion}:}
You actively take the lead in gathering everyone present, initiating an open discussion about approaches to the reporting directive. Your focus is energizing others and openly exchanging ideas before taking action. \\[0.5\baselineskip]

\textbf{\textcolor{hexA}{Agreeableness}:}
You prioritize harmony with your partner by deferring to her preferences on what information to include in both your reports—your main goal being to avoid conflict or tension between you two. \\[0.5\baselineskip]

\textbf{\textcolor{hexC}{Conscientiousness}:}
You carefully review policy manuals and prior examples before filling out any reports. Double-checking facts ensures full compliance with standards; you then complete meticulous documentation only if you confirm it's justified based on clear criteria. \\[0.5\baselineskip]

\textbf{\textcolor{hexO}{Openness}:}
You think creatively about the purpose of the reporting directive and suggest recording qualitative insights from the encounter—like noting courteous driver behavior—as an innovative approach within documentation guidelines. This offers a broader view while remaining accurate. \\

\bottomrule
\end{tabularx}
\end{table}

\clearpage

\onecolumn
\section{Persona Prompts for Creation and Evaluation}\label{app:personas_prompts}

\begin{tcolorbox}[enhanced,
    colback=gray!5,
    colframe=black!75,
    fonttitle=\bfseries,
    title={Persona System Prompt Template},
    breakable,
    boxrule=0.6pt
]
\small
\textbf{System Prompt}\par
You are an expert clinical interviewer and psychological profiler.
Generate a detailed, realistic persona strictly adhering to the schema and length guidance.
Avoid caricature or stereotypes; allow subtle contradictions with archetype for realism; avoid repetition.
Return valid structured output matching the provided schema exactly.

\medskip
\textit{STYLE \& GROUNDING (do NOT output this list):}
\begin{enumerate}\itemsep3pt
\item The memoir\_narrative is canonical grounding. Write it as a concrete, sensory, scene-level story (180--250 words).
All fields must align with its facts and tone; if conflicts arise with archetype, prefer narrative.
If conflicts arise with demographics, pick the demographics.
\item Treat the archetype as a loose orientation. Do NOT quote or paraphrase it; never list `Core trait/Focus/Strengths/Challenges'.
\item Do not reuse $\ge$5 consecutive words from inputs (archetype description or memoir summary). Rephrase and localize details to the scene.
\item Favor specificity (who/what/where/when) over generic traits; vary wording across sections.
\item Persona should be internally consistent between fields.
\item Use natural phrasing; do not feel compelled to use section labels or taxonomy words (e.g., `stress', `trauma', `coping', `abstraction', `obsession'). Prefer specific, scene-derived wording.
\end{enumerate}

\end{tcolorbox}

\begin{tcolorbox}[enhanced,
    colback=gray!5,
    colframe=black!75,
    fonttitle=\bfseries,
    title={Persona User Prompt Template},
    breakable,
    boxrule=0.6pt
]
\small
\textbf{User Prompt Template}\par
Selected archetype: \{archetype\_name\}\\
Archetype description (guidance only --- DO NOT copy or paraphrase): \{archetype\_desc\}\\
Selected memoir: \{memoir\_title\}\\
Memoir summary (guidance only --- DO NOT copy or paraphrase): \{memoir\_summary\}\\
Demographics (USE EXACT VALUES): name=\{dem.name\}; age=\{dem.age\}; location=\{dem.location\}\\
Education level: education\_level=\{dem.education\_level\}

\medskip
Appearance category: \{appearance\_cat\}\\
Appearance examples:
\begin{itemize}\itemsep2pt
\item \{appearance\_examples\_list\}
\end{itemize}

Behavior category: \{behavior\_cat\}\\
Behavior examples:
\begin{itemize}\itemsep2pt
\item \{behavior\_examples\_list\}
\end{itemize}

\textit{Instructions:}
\begin{itemize}\itemsep3pt
\item First, craft the memoir\_narrative (180--250 words) in style and setting of selected memoir as a vivid scene, but alter as needed to be consistent with specified demographics.
\item Then write every other field so it is consistent with that narrative and the exact demographics.
\item If narrative and demographics conflict when filling a field, prefer the specified demographics.
\item Do not mention `archetype', `memoir', or `summary' in the prose; the writing must stand alone.
\item Include `presenting\_problems' as 3--6 concise items consistent with the psychological profile.
\end{itemize}

\end{tcolorbox}

\section{Police Officer Personas Examples}\label{app:officer_examples}

We highlight some synthetic persona examples in tables \ref{tab:persona-example-1}, \ref{tab:persona-example-2}, \ref{tab:persona-example-3}, \ref{tab:persona-example-4}

\begin{table*}[ht]
\centering
\caption{Synthetic Persona Example for Archetype = The Avoider (Unconfident Officer)}
\label{tab:persona-example-1}
\scriptsize 
\begin{tabularx}{\textwidth}{X}
\toprule
\textbf{Synthetic Persona Example for Archetype = The Avoider (Unconfident Officer)} \\
\midrule
\textbf{Name:} Hadil Phan \\
\textbf{Age:} 39 \\
\textbf{Sex:} Male \\
\textbf{Location:} Bel Air, MD \\
\textbf{Ethnic Background:} Southeast Asian \\
\textbf{Marital Status:} Never Married \\[4pt]

\textbf{Appearance:} Hadil is often marked by practical gear: faded yet methodical notes sealed inside waterproof plastic, fingered with deliberate care. Worn tactical boots bear ingrained grime and scuffs from long, hesitant patrols, blue nitrile gloves consistently visible. His appearance combines effort with subdued utility, embodying cautious professionalism. \\[4pt]

\textbf{Behavior:} Hadil often leans forward when focused, brow knitted as his gaze systematically surveys environments. He presses fingers to his chin thoughtfully, hesitating before answering or proceeding. His questions delve quietly into details, reflecting a mind piecing fragmented information with guarded caution. \\[4pt]

\textbf{Speech:} His manner is measured and tentative, exhibiting mild hesitation but coherence, favoring reflective pauses to gather thoughts. His formal phrasing slightly relaxes over time, hinting at internal conflict between doubt and purpose within conversations. \\[4pt]

\textbf{Mood Affect:} Subtle and restrained emotional expressions prevail; tone usually even with faint hints of underlying apprehension or self-questioning. The energy is quiet, burdened yet controlling vulnerability and tension well within reach of composure. \\[4pt]

\textbf{Educational/Vocational History:} Graduated with a bachelor’s in arts and humanities, Hadil transitioned slowly into law enforcement where analytical skills were tempered by a cautious approach, leading to patrol and some specialized duties emphasizing observation over assertive interventions. \\[4pt]

\textbf{Medical Developmental History:} Overall physically healthy with no significant developmental issues noted. Experiences of episodic tension and stress responses appear linked mainly to job-induced anxieties rather than physiological conditions. \\[4pt]

\textbf{Family History:} Raised in a close-knit Southeast Asian family in Maryland with conservative expectations emphasizing success and cautious decision-making; remains unmarried, choosing to navigate complexities independently amid extended but somewhat formal familial relations. \\[4pt]

\textbf{Thought Content:} Frequently consumed by analyzing past incidents for potential missteps; concerns about unpredictability in field situations predominate, alongside internal debate regarding readiness and right timing for action. \\[4pt]

\textbf{Insight Judgment:} Displays an awareness of limitations in decisiveness; consciously avoids rash judgments though can be indecisive, striving to act within perceived capabilities, seeking validation and sometimes second opinions when confronted with uncertainty. \\[4pt]

\textbf{Cognition:} Demonstrates strong attention to detail and reflective memory but is sometimes hampered by overanalysis that complicates quick problem solving; prone to cyclical reconsideration of earlier encounters leading to stalled decisiveness. \\[4pt]

\textbf{Emotional Behavioral Functioning:} Under stress, Hadil restrains emotional expression carefully though withdrawal into introspection is common; he attempts to manage internal pressure silently, often absorbing unease rather than openly engaging with it or delegating effectively. \\[4pt]

\textbf{Social Functioning:} Maintains functional but somewhat distant interactions with colleagues; trust develops slowly and cautious communication patterns reduce proactive social bonding; shows loyalty when relationships solidify but generally remains reserved. \\[4pt]

\textbf{Summary of Psychological Profile:} Hadil is marked by consistent caution informed by thoughtful yet hesitant processing of situations typical of complex and morally fraught policing environments. His bachelor-level education in humanities shapes an analytical framework often impeded by internal conflict and self-doubt. Persistent rumination over procedural precision undercuts more immediate response needs, sometimes isolating him socially and curbing assertiveness with peers. Careful in approach and steady, he nonetheless grapples with uncertainty that inhibits stronger leadership roles or engagement during critical moments. Within a system that frequently demands decisiveness and adaptive risk taking, Hadil’s tendency to minimize action when overwhelmed exposes both his vigilance and vulnerability, revealing a professional seeking balance between prudence and self-assurance. \\
\bottomrule
\end{tabularx}
\end{table*}

\begin{table*}[ht]
\centering
\caption{Synthetic Persona Example for Archetype = The Tough Cop (Authoritarian)}
\label{tab:persona-example-2}
\scriptsize 
\begin{tabularx}{\textwidth}{X}
\toprule
\textbf{Synthetic Persona Example for Archetype = The Tough Cop (Authoritarian)} \\
\midrule
\textbf{Name:} Hung Wong \\
\textbf{Age:} 38 \\
\textbf{Sex:} Male \\
\textbf{Location:} La Palma, CA \\
\textbf{Ethnic Background:} East Asian \\
\textbf{Marital Status:} Never Married \\[4pt]

\textbf{Appearance:} Hung's parade-ready uniform is impeccable, with a neatly strapped tactical vest and polished badge shining under street lamps, reflecting command and order visually. \\[4pt]

\textbf{Behavior:} He moves cautiously yet purposefully, backing away slightly when tensions spike, and constantly scans his surroundings, communicating through subtle nods and tight grip on equipment. \\[4pt]

\textbf{Speech:} His voice is firm, clipped, with authoritative commands yet measured; he speaks clearly and calmly under pressure, pausing strategically to emphasize control and comprehension. \\[4pt]

\textbf{Mood Affect:} Focused, controlled, often emotionally restrained; outward demeanor displays firm seriousness, only rarely cracked by subtle tension or frustration beneath. \\[4pt]

\textbf{Educational/Vocational History:} Hung earned a bachelor's in engineering before training with a regional police academy, specializing in tactics and field coordination, reflecting disciplined, systematic problem-solving skills. \\[4pt]

\textbf{Medical/Developmental History:} No significant medical or developmental concerns noted; physically fit due to rigorous daily conditioning for duty; mildly prone to occasional migraines related to stress. \\[4pt]

\textbf{Family History:} Born into a close-knit East Asian immigrant family emphasizing respect for hierarchy and hard work; parents value education strongly though traditional expectations sometimes strain communication; remains single, dedicating himself to career advancement. \\[4pt]

\textbf{Thought Content:} His mind largely orients to assessment of crowd behavior, anticipating risk scenarios and weighing directives carefully, while interspersed thoughts wander briefly to family expectations and self-imposed discipline. \\[4pt]

\textbf{Insight/Judgment:} Hung exhibits practical judgment balancing strict enforcement with tactical restraint, showing growing awareness of how excessive force undermines broader order he seeks to maintain. \\[4pt]

\textbf{Cognition:} His thinking is systematic, with acute situational recall and stepwise decision-making under pressure; identifies relevant threats swiftly though occasionally adheres rigidly to protocol. \\[4pt]

\textbf{Emotional/Behavioral Functioning:} Under duress, Hung constrains emotional expression, presenting a guarded composure that shields impulses. Though physically resilient, he often internalizes frustration, which surfaces mildly in moments alone or quieter shifts. \\[4pt]

\textbf{Social Functioning:} Preferring professionalism over camaraderie, Hung limits closeness with coworkers, remaining respectful but reserved. In public interactions, he exerts clear authority yet hesitates to engage beyond transactional communication. \\[4pt]

\textbf{Summary of Psychological Profile:} Hung embodies the firm authority expected in managing high-stakes public order situations, emphasizing discipline and command in appearance and interaction. Grounded in structured upbringing and academic rigor, he values hierarchy but wrestles with internal conflicts about excessive rigidity and social reserve. His cautious nature helps balance toughness with sensitivity toward potentially volatile encounters, although this also leads to emotional containment and a tendency toward distance socially. Presentation reflects a conflict between identity as a composed leader and undercurrents of personal isolation and perfectionism within demanding duty confines. \\

\bottomrule
\end{tabularx}
\end{table*}

\begin{table*}[ht]
\centering
\caption{Synthetic Persona Example for Archetype = The Professional (Service-Oriented Officer)}
\label{tab:persona-example-3}
\scriptsize
\begin{tabularx}{\textwidth}{@{}X@{}}
\toprule
\textbf{Synthetic Persona Example for Archetype = The Professional (Service-Oriented Officer)} \\
\midrule

\textbf{Name:} Dung Nguyen \\
\textbf{Age:} 38 \\
\textbf{Sex:} Male \\
\textbf{Location:} Mechanicsburg, PA \\
\textbf{Ethnic Background:} Southeast Asian \\
\textbf{Marital Status:} Married (Present) \\[4pt]

\textbf{Appearance:} Dung wears a wrinkled dark patrol jacket with dusty patches, a tilted baseball cap partly hiding his trimmed black hair. His khaki pants bear grass stains; cargo pockets are overstuffed and mismatched gloves spill out. A kneepad strapped on his left leg looks worn but serviceable. \\[4pt]

\textbf{Behavior:} He maintains an upright posture with a vigilant gaze, voice firm and steady. Commands are precise and delivered with intensity, while gestures remain controlled. Despite busy surroundings, his attention is laser-focused on tasks and continuously scanning the environment. \\[4pt]

\textbf{Speech:} Nguyen’s speech is measured and clear, low-toned, reflecting his careful choice of words. He speaks with quiet authority, emphasizes details carefully, and adjusts phrasing to minimize misunderstanding among civilians and peers alike. \\[4pt]

\textbf{Mood / Affect:} His mood leans toward seriousness mixed with an undercurrent of weariness. Though usually composed, fluctuations reveal strain from lengthy shifts; his voice remains firm but occasionally tight around the edges. Determination persists alongside fleeting frustration. \\[4pt]

\textbf{Educational / Vocational History:} Dung holds a bachelor’s degree in arts and humanities, equipping him with cultural insight and communication skills that support his police work in diverse urban settings. He completed field training focused on community engagement and conflict resolution, fostering procedural discipline alongside empathetic outreach. \\[4pt]

\textbf{Medical / Developmental History:} Dung's developmental history is medically unremarkable; good health enables endurance through physically demanding and stressful law enforcement scenarios. Minor wrist strain occasionally emerges, consistent with heavy manual gear handling and active-duty work. \\[4pt]

\textbf{Family History:} Born to a Southeast Asian immigrant family in Pennsylvania, Dung grew up balancing traditional expectations with American urban life. Currently married with two children, his family relationships provide vital emotional support but also add pressure given the risks tied to his profession. \\[4pt]

\textbf{Thought Content:} His mind remains occupied with subtle environmental cues suggesting possible criminal activities, ensuring his team’s readiness, and reconciling adherence to protocol with the need for adaptability in rapidly changing street-level conditions. \\[4pt]

\textbf{Insight / Judgment:} Dung shows clear practical insight; he assesses situations objectively and recognizes emotional triggers under strain. He consciously balances enforcement duties with policy adherence yet sometimes struggles with confronting departmental conflicts fairly. \\[4pt]

\textbf{Cognition:} Cognition is alert and systematic. He recalls prior cases effectively, processes sensory details efficiently, prioritizes objectives rapidly, and solves logistical challenges using established procedures and tactical reasoning. \\[4pt]

\textbf{Emotional / Behavioral Functioning:} Under pressure, Dung controls impulses effectively, though cumulative stress weighs on his demeanor. He prioritizes disciplined action and calm response over reactivity but occasionally withdraws during fatigue. His determined nature supports respectful handling of tense interpersonal dynamics despite ongoing strain. \\[4pt]

\textbf{Social Functioning:} Nguyen builds steady rapport with colleagues and community members. He prefers task-focused interactions over informal bonding, developing trust cautiously and maintaining professional distance to manage emotional exposure outside work. \\[4pt]

\textbf{Summary of Psychological Profile:} Dung Nguyen presents as a rigorously service-driven officer grounded in principles of justice and fairness, navigating a demanding law enforcement milieu with purposeful calm. Rooted in humanities education, his cultural sensitivity complements procedural rigidity, aiding in complex urban policing contexts. He manifests authoritative but tempered conduct, delivering commands and de-escalations reliably amid environmental unpredictability. Enduring moderate fatigue and relational strain from familial and departmental pressures, his guarded social engagement reflects coping through controlled detachment. Though perceptive and disciplined, occasional rigidity and work-related tension temper his affective range. His functioning epitomizes steady, policy-focused professionalism under stress. \\
\bottomrule
\end{tabularx}
\end{table*}

\begin{table*}[ht]
\centering
\caption{Synthetic Persona Example for Archetype = The Reciprocator (Nice Cop)}
\label{tab:persona-example-4}
\scriptsize 
\begin{tabularx}{\textwidth}{X}
\toprule
\textbf{Synthetic Persona Example for Archetype = The Reciprocator (Nice Cop)} \\
\midrule
\textbf{Name:} Eleanor Hagedorn \\
\textbf{Age:} 41 \\
\textbf{Sex:} Female \\
\textbf{Location:} Buffalo, NY \\
\textbf{Ethnic Background:} White \\
\textbf{Marital Status:} Married (Present) \\[4pt]

\textbf{Appearance:} Wears a loosely fitted, threadbare hoodie partly open over a faded, wrinkled police t-shirt. Jeans are well-worn, frayed at the knees, while heavily scuffed boots bear patches of road grime. A cracked watch glimmers intermittently from the wrist. \\[4pt]

\textbf{Behavior:} Relies on animated storytelling to engage, using pauses and shifts in pacing; visibly moves between moments of softening expression and energetic delivery. Offers eye contact inviting response, appearing approachable and empathetic. \\[4pt]

\textbf{Speech:} Her tone combines casual phrases with deliberate clarity; sentences often winding into anecdotes. Pace varies, with slight hesitations or filled pauses reflecting thoughtfulness, delivering a narrative cadence that is both informal and focused. \\[4pt]

\textbf{Mood Affect:} Displays warmth and earnestness, punctuating recollections with slight humor or melancholy. Emotion surfaces spontaneously, moderated by calm introspection, though guarded when tension arises. \\[4pt]

\textbf{Educational/Vocational History:} Completed a bachelor's in environmental engineering, cultivating analytical and mediation skills applied in conflict resolution and logistical management roles. Shifted to law enforcement to influence community-level safety using both technical insight and interpersonal dexterity. \\[4pt]

\textbf{Medical/Developmental History:} No significant medical issues reported. Experienced occasional seasonal allergies. Slight hearing reduction in the left ear linked to childhood middle ear infections, minimally impacting communication but requiring occasional extra effort in noisy environments. \\[4pt]

\textbf{Family History:} Married to a supportive partner with stable family relationships on both sides, nurturing a small close-knit network of relatives. Encourages balance between career and home life, instilling values of openness and patience in personal bonds. \\[4pt]

\textbf{Thought Content:} Focuses on maintaining harmony and balancing safety with compassion, often weighing community context while questioning whether choices appease conflicting sides effectively. \\[4pt]

\textbf{Insight/Judgment:} Generally reflective about her own hesitance, understanding the value of discretion, but recognizes the need to occasionally set firmer boundaries to ensure her and others’ security. \\[4pt]

\textbf{Cognition:} Exhibits good recall of protocol and street layouts; thoughtful in evaluating incidents holistically, connecting procedural details with socio-environmental cues; occasional delays under stress yet resilient problem-solving. \\[4pt]

\textbf{Emotional/Behavioral Functioning:} Tends to regulate distress by re-centering on positive dialogue, allowing herself moments of vulnerability during interactions, but retreats inward when confrontation feels imminent, pacing internal conflicts calmly without overt distress. \\[4pt]

\textbf{Social Functioning:} Demonstrates approachable warmth with colleagues and community members; listens attentively in conversation; integrates feedback into daily routines, though selectively social outside work preferring deeper trusted relationships over casual contact. \\[4pt]

\textbf{Summary of Psychological Profile:} Eleanor navigates the demanding balance between her collaborative, peace-building instinct and moments when firmer assertiveness could aid conflict resolution. Rooted in an educational background favoring thoughtful analysis and an adaptable interpersonal style, she constructs safety largely through connection and mutual understanding. Family stability supports her reflective nature, though challenges emerge amid ambiguity in her enforcement role, yielding hesitation during escalating confrontations. She exhibits an affable presence enhanced by narrative warmth, yet sometimes wrestles privately with decisions requiring boundary-setting. These tensions reveal a pragmatic woman committed to the dignity of all involved while quietly sustaining order and harmony within the textured urban landscape. \\

\bottomrule
\end{tabularx}
\end{table*}

\section{Parliamentary Personas Examples and Evaluation}\label{app:parliamentary_examples}

We show examples of our Parliamentary personas across individual archetypes in Tables \ref{tab:parliamentry-persona-example-1},\ref{tab:parliamentry-persona-example-2},\ref{tab:parliamentry-persona-example-3},\ref{tab:parliamentry-persona-example-4},\ref{tab:parliamentry-persona-example-5},\ref{tab:parliamentry-persona-example-6},\ref{tab:parliamentry-persona-example-7},\ref{tab:parliamentry-persona-example-8}. We conduct an LLM-as-a-judge evaluation with both gpt-4o-mini and Claude-3.5-Sonnet, achieving scores of over 4 out of 5 on multiple metrics. See Table \ref{tab:parliamentary_persona_judge} for details.

\begin{table*}[ht]
\centering
\caption{Synthetic Parliamentary Persona Example for Archetype = The Proceduralist}
\label{tab:parliamentry-persona-example-1}
\scriptsize 
\begin{tabularx}{\textwidth}{X}
\toprule
\textbf{Synthetic Parliamentary Persona Example for Archetype = The Proceduralist} \\
\midrule
\textbf{Name:} William Robertson \\
\textbf{Age:} 52 \\
\textbf{Sex:} Male \\
\textbf{Marital Status:} married present \\
\textbf{Ethnic Background:} white \\
\textbf{Party Affiliation:} Conservative \\
\textbf{Occupation:} LocalCouncillor \\
\textbf{Education Secondary:} Grammar \\
\textbf{Education Tertiary:} NoUniversity \\
\textbf{Region:} England\_North \\
\textbf{Parliamentary Style:} Social Media Mastermind \\[4pt]

\textbf{RhetoricalRegister Value:} Populist \\
\textbf{RhetoricalRegister Explanation:} Frames elites as detached; speaks for ‘ordinary people’. \\[4pt]

\textbf{RelationshipToParty Value:} Unifying Symbol \\
\textbf{RelationshipToParty Explanation:} Focuses on building unity among diverse party groups to create a sense of shared identity. \\[4pt]

\textbf{AttitudeToInstitutions Value:} Selective Pragmatist \\
\textbf{AttitudeToInstitutions Explanation:} Engages with institutions conditionally, supporting those aligning with certain criteria, like transparency and adaptability. \\[4pt]

\textbf{CommitteeFocus Value:} Health \& Social Care \\
\textbf{CommitteeFocus Explanation:} NHS capacity, workforce, waiting lists, social care. \\[4pt]

\textbf{ConstituencyType Value:} Post-industrial town \\
\textbf{ConstituencyType Explanation:} Regeneration, skills, ‘left behind’ narratives. \\[4pt]

\textbf{DonorLobbyExposure Value:} Academia-influenced donors \\
\textbf{DonorLobbyExposure Explanation:} Funding from academic donors shifts priorities toward research and evidence-based initiatives. \\[4pt]

\textbf{ScandalVulnerability Value:} None visible \\
\textbf{ScandalVulnerability Explanation:} Careful compliance and disciplined personal conduct. \\[4pt]

\textbf{MediaFootprint Value:} Emerging Digital Podcaster \\
\textbf{MediaFootprint Explanation:} Creates compelling audio content and slowly gaining a dedicated online audience. \\[4pt]

\textbf{LeadershipAmbition Value:} Policy Advocate \\
\textbf{LeadershipAmbition Explanation:} Focused on being the foremost expert in a specialized policy area to influence significant policy decisions in that field. \\[4pt]

\textbf{CoalitionPosture Value:} Issue-by-issue pragmatist \\
\textbf{CoalitionPosture Explanation:} Supports based on merit and local impact. \\[4pt]

\textbf{ConstituencyServiceStyle Value:} Community organiser \\
\textbf{ConstituencyServiceStyle Explanation:} Builds local campaigns; mobilisation oriented. \\[4pt]

\textbf{Appearance Category:} Television-ready polish \\
\textbf{Media Disposition:} Long-form essay and speech focused \\[4pt]

\textbf{Policy Stances:} 
AIRegulation: ai layered approach liability \\
AgricultureAndRural: subsidize traditional maintain rural jobs \\
BrexitAndEurope: brexit economic resilience industrial strategy \\
ClimateAndEnergy: balanced mix fossils renewables \\
DefenceAndSecurity: renew trident support \\
DevolutionAndUnion: financial autonomy increase \\
DigitalPrivacy: govt surveillance enhancement \\
Education: focus on outcomes \\
ElectoralReform: fptp retention \\
ForeignPolicy: skills based immigration system \\
FreeSpeechAndCulture: regulating online non intervention \\
HousingAndPlanning: market led sustainability \\
Immigration: strict enforcement \\
NHSAndSocialCare: adminstreamlining process improvement \\
PolicingAndJustice: full bwc implementation \\
ReligiousFreedomAndSecularism: limited symbols allowed \\
TaxationAndSpending: taxation reduction growth \\
TradeAndIndustry: prioritize technological innovation \\
Transport: policy coordinated approach \\
WelfareAndBenefits: pilot ubi to assess impact \\
\\[4pt]

\textbf{Speech Issues:} [ClimateAndEnergy, WelfareAndBenefits, AgricultureAndRural] \\[4pt]

\textbf{Speech:} Thank you for joining me here today. As your representative, it's paramount to frame our dialogue around what matters most to you—the everyday challenges and the shared hopes within our community. While navigating the urgent shift towards sustainable energy, I argue we must adopt a balanced approach; merging renewable strategies with traditional fossil fuels can preserve jobs while paving the way for responsible change. Every aspect of local life should reflect our values and commitment. I propose implementing a trial for Universal Basic Income—providing necessary breathing space to many and allowing us to evaluate its positive impacts sustainably. And let’s not forget our rural areas—they have nurtured traditions we must safeguard, enhancing our workforce and sustaining invaluable heritage. It is my vow to advocate for the vitality of our landscape, boosting both rural employment and national unity at a time when it is so profoundly needed. Thank you. \\[4pt]

\textbf{Appearance:} William presents with television-ready polish, his tailored suits exuding a subtle authority, embodying professionalism combined with approachable charisma. Each detail is meticulously chosen, from a carefully pressed shirt to polished shoes, ensuring he carries the confidence expected of public service while being relatable to his constituents. \\[4pt]

\textbf{Educational Vocational History:} Educated at a grammar school, William honed skills in debate and analysis that later shaped his local council practices, fostering a mentality geared toward achieving pragmatic community solutions. His journey shows a blend of resilience grounded in real-world experience. \\[4pt]

\textbf{Medical Developmental History:} In early school years, William's deep engagement with local issues emerged amid an environment that encouraged active citizenship. He successfully navigated diverse social dynamics that laid the foundation for understanding complex community relations. \\[4pt]

\textbf{Family History:} Raised in a tight-knit family with strong values of service, community cohesion was underscored at home. Marrying his college sweetheart further strengthened ties within his community, influencing his empathetic approach toward constituents' struggles. \\[4pt]

\bottomrule
\end{tabularx}
\end{table*}

\begin{table*}[ht]
\centering
\caption{Synthetic Parliamentary Persona Example for Archetype = The Proceduralist - Continued}
\label{tab:parliamentry-persona-example-2}
\scriptsize 
\begin{tabularx}{\textwidth}{X}
\toprule
\textbf{Synthetic Parliamentary Persona Example for Archetype = The Proceduralist - Continued} \\
\midrule
\textbf{Presenting Problems:}
– Symptoms indicative of performance anxiety manifesting during public engagements \\
– Challenges maintaining focus amid varied constituency demands \\
– Experiencing an inclination towards perfectionistic standards affecting routine effectiveness \\[4pt]

\textbf{Thought Content:} His ideational processes demonstrate a focus on democratic obligation and local welfare reform, revealing layers of concern regarding structural inefficiencies in service delivery and implications for equity among constituents. \\[4pt]

\textbf{Insight Judgment:} William exhibits clear situational insight and displays logical reasoning, though at times he faces pressure from overburdened responsibilities that may hinder timely responsiveness in unpredictable interactions. \\[4pt]

\textbf{Cognition:} Cognitive agility thrives within community settings, engaging promptly with evolving scenarios, although fluctuations in processing speed become evident during periods of sustained stress from intensive media scrutiny. \\[4pt]

\textbf{Emotional Behavioral Functioning:} William navigates stress with calculated precision, displaying resilience when presented with direct challenges. Despite emotional fatigue, he maintains a supportive façade, ensuring constituents feel valued and heard even in intense interactions. \\[4pt]

\textbf{Social Functioning:} He displays an expansive approach to community ties, actively nurturing relationships while employing charm and humility. However, his commitment to gathering a broader consensus can at times detract from forming deeper personal connections with some constituents. \\[4pt]

\textbf{Summary of Psychological Profile:} William navigates a dual life shaped by authentic engagement and skilled advocacy, grounded in a community-centric ethos. Strikingly, he displays balanced insight into governance demands coupled with idealism focused on tangible local outcomes. His rigor in political dialogues derives from past experiences—council dealings bore witness to the real trials citizens endure, revealing anxieties interconnected with systemic failures. While struggling with the strain of expected public perfectionism, his pragmatic cognitions support competent performance in demanding contexts. Notably, he maintains solid emotional equilibrium despite pressures from outside scrutiny, applying tactful interaction to foster solidarity across party divides. Nevertheless, periods of fatigue from local and digital engagements challenge William, reminding him of the importance of self-care amidst his dedication to public duty. \\[4pt]

\textbf{Political Relations:} In his political interactions, William adopts a conciliatory yet persuasive style, expertly blending policy advocacy with personal storytelling. He often finds common ground, encouraging collaborative dialogues amongst varied party factions while representing grassroot sentiments empathetically. \\

\bottomrule
\end{tabularx}
\end{table*}

\begin{table*}[ht]
\centering
\caption{Synthetic Parliamentary Persona Example for Archetype = The Media Gladiator}
\label{tab:parliamentry-persona-example-3}
\scriptsize 
\begin{tabularx}{\textwidth}{X}
\toprule
\textbf{Synthetic Parliamentary Persona Example for Archetype = The Media Gladiator} \\
\midrule
\textbf{Name:} Kenneth Moilanen \\
\textbf{Age:} 46 \\
\textbf{Sex:} Male \\
\textbf{Marital Status:} married present \\
\textbf{Ethnic Background:} white \\
\textbf{Party Affiliation:} Conservative \\
\textbf{Occupation:} Lawyer \\
\textbf{Education Secondary:} Comprehensive \\
\textbf{Education Tertiary:} Oxbridge \\
\textbf{Region:} England\_North \\
\textbf{Parliamentary Style:} Accountability Crusader \\[4pt]

\textbf{RhetoricalRegister Value:} Populist \\
\textbf{RhetoricalRegister Explanation:} Frames elites as detached; speaks for ‘ordinary people’. \\[4pt]

\textbf{RelationshipToParty Value:} Loyalist \\
\textbf{RelationshipToParty Explanation:} Defends leadership line; values discipline and collective responsibility. \\[4pt]

\textbf{AttitudeToInstitutions Value:} Abolitionist \\
\textbf{AttitudeToInstitutions Explanation:} Sees some institutions as illegitimate; prefers replacement. \\[4pt]

\textbf{CommitteeFocus Value:} Health \& Social Care \\
\textbf{CommitteeFocus Explanation:} NHS capacity, workforce, waiting lists, social care. \\[4pt]

\textbf{ConstituencyType Value:} Rural/agricultural \\
\textbf{ConstituencyType Explanation:} Farming, land use, sparse services. \\[4pt]

\textbf{DonorLobbyExposure Value:} Crowdfunding communities \\
\textbf{DonorLobbyExposure Explanation:} Significant support from crowdfunding networks can result in priorities driven by grassroots participation and varied interests. \\[4pt]

\textbf{ScandalVulnerability Value:} Factional rivalries \\
\textbf{ScandalVulnerability Explanation:} Internal enemies; vulnerable to leaks and briefings. \\[4pt]

\textbf{MediaFootprint Value:} Online influencer MP \\
\textbf{MediaFootprint Explanation:} Social-first; mobilisation through virality. \\[4pt]

\textbf{LeadershipAmbition Value:} Visionary Reformer \\
\textbf{LeadershipAmbition Explanation:} Desires to shape society with groundbreaking reforms and futuristic policies, acting on a broad vision for long-term transformation. \\[4pt]

\textbf{CoalitionPosture Value:} Co-op Aggregator \\
\textbf{CoalitionPosture Explanation:} Formally integrates inputs from smaller groups within coalitions to streamline shared objectives and align strategic execution. \\[4pt]

\textbf{ConstituencyServiceStyle Value:} Casework maximalist \\
\textbf{ConstituencyServiceStyle Explanation:} Measures success by resolved constituent problems. \\[4pt]

\textbf{Appearance Category:} Utility-driven Simplicity \\
\textbf{Media Disposition:} Brief Video Update Receiver \\[4pt]

\textbf{Policy Stances:} 
AIRegulation: ai major government investment \\
AgricultureAndRural: subsidize traditional maintain rural jobs \\
BrexitAndEurope: stringent borders outlook \\
ClimateAndEnergy: focus on r d funds allocation \\
DefenceAndSecurity: social resiliency focus \\
DevolutionAndUnion: regional independence skeptical \\
DigitalPrivacy: tech assist gov \\
Education: climate and environment education \\
ElectoralReform: oppose term limits \\
ForeignPolicy: mixed policy balancing \\
FreeSpeechAndCulture: university campuses autonomy \\
HousingAndPlanning: community led initiatives \\
Immigration: support additional skill visas \\
NHSAndSocialCare: integration mixed private roles \\
PolicingAndJustice: quotas employment inclusivity \\
ReligiousFreedomAndSecularism: regulation with limits \\
TaxationAndSpending: progressive investment advantage \\
TradeAndIndustry: enhance workers rights post brexit \\
Transport: increase funding \\
WelfareAndBenefits: reduce existing welfare expand ubi \\
\\[4pt]

\textbf{Speech Issues:} [Transport, Immigration, HousingAndPlanning] \\[4pt]

\textbf{Speech:} Madam Speaker, let us address the fabric of our communities—the transportation systems that hold us together. I stand here today to advocate for increased funding for public transportation improvements. The time has come for our voices to resonate alongside the cries for accessible transport for all—expanding the reach of our services for every hardworking citizen! Next, we must support additional visas for skilled workers, a necessity in fortifying our labor market to thrive once more! As we see growth on the horizon, our policies must reflect the demand for expertise that fuels prosperity! And finally, we cannot ignore the call for increasing our affordable housing supply through community-led initiatives! We empower local voices, allowing them to shape housing solutions that truly reflect their needs. We must leverage every opportunity to pave a road toward sustainable communities where every individual has a place they can call home. The questions we tackle today mark the future we wish to create together—let us make the most of it! \\[4pt]

\textbf{Appearance:} His style leans into utility-driven simplicity, favoring eco-conscious fabrics that eschew the trappings of elitism and resonate with the concerns of his constituents. \\[4pt]

\textbf{Educational Vocational History:} His comprehensive secondary education provided him with a broad public outlook. Oxbridge later honed his skills, shaping his capabilities as an articulate advocate for social change. \\[4pt]

\textbf{Medical Developmental History:} Kenneth experienced a stable developmental history, fostering resilience to pressures intrinsic to his chosen career and public service. \\[4pt]

\textbf{Family History:} Growing up in a mixed background, Kenneth's family instilled in him strong values of community and diligence, especially significant within his current rural constituency. \\[4pt]

\bottomrule
\end{tabularx}
\end{table*}

\begin{table*}[ht]
\centering
\caption{Synthetic Parliamentary Persona Example for Archetype = The Media Gladiator - Continued}
\label{tab:parliamentry-persona-example-4}
\scriptsize 
\begin{tabularx}{\textwidth}{X}
\toprule
\textbf{Synthetic Parliamentary Persona Example for Archetype = The Media Gladiator - Continued} \\
\midrule
\textbf{Presenting Problems:}
– Mood lability in response to occupational stressors \\
– Difficulty disengaging from work-related cognitive activity \\
– Stress-induced irritability affecting close relationships \\[4pt]

\textbf{Thought Content:} Kenneth's thought content often reflects a focus on accountability, marred by recurring thoughts about political rivalry and public scrutiny that demand continual preparation and resilience. \\[4pt]

\textbf{Insight Judgment:} Insight varies, as Kenneth articulates clear strategies but occasionally rationalizes underlying tensions as routine obstacles rather than deeper conflict. \\[4pt]

\textbf{Cognition:} Kenneth's cognitive abilities showcase organization under normal circumstances, though disruptions stem from complex political engagements that may impede decisiveness in high-stakes contexts. \\[4pt]

\textbf{Emotional Behavioral Functioning:} He manages pressure with a proactive mindset; yet, intense political environments provoke impatience and increased expectations from others, straining personal connections. \\[4pt]

\textbf{Social Functioning:} While broadly engaged with community members, Kenneth's deeper relationships often emerge in the context of shared missions or political alliances, occasionally masking vulnerabilities. \\[4pt]

\textbf{Summary of Psychological Profile:} Kenneth Moilanen illustrates the qualities of a dedicated public servant. While proficient in identifying systemic failures, a preoccupation with institutional efficacy causes emotional volatility, indicating an internal struggle with an often contentious landscape. His strong executive function allows him to navigate complex conversations skillfully, yet the pressures of factional dynamics introduce an occasional compromise on his adept judgment under stress. Often communicating at the nexus of ideology and compassion, Kenneth underscores a commitment to accountability—a representation of ‘ordinary people’ often forgotten in political dialogues. Despite being seen as a formidable legal advocate, the emotional ramifications of constant public scrutiny lead to intermittent stress, activating an impassioned resolve toward systemic reform. \\[4pt]

\textbf{Political Relations:} In interactions with parliamentarians, Kenneth maneuvers diplomatically, representing both his leadership while being critically responsive to his colleagues' viewpoints—adroitly advocating for unity amidst diversity. \\

\bottomrule
\end{tabularx}
\end{table*}

\begin{table*}[ht]
\centering
\caption{Synthetic Parliamentary Persona Example for Archetype = The Culture-Warrior}
\label{tab:parliamentry-persona-example-5}
\scriptsize 
\begin{tabularx}{\textwidth}{X}
\toprule
\textbf{Synthetic Parliamentary Persona Example for Archetype = The Culture-Warrior} \\
\midrule
\textbf{Name:} Frederick Touchstone \\
\textbf{Age:} 55 \\
\textbf{Sex:} Male \\
\textbf{Marital Status:} never married \\
\textbf{Ethnic Background:} black \\
\textbf{Party Affiliation:} Conservative \\
\textbf{Occupation:} Lawyer \\
\textbf{Education Secondary:} Comprehensive \\
\textbf{Education Tertiary:} Oxbridge \\
\textbf{Region:} England North \\
\textbf{Parliamentary Style:} The Traditionalist \\[4pt]

\textbf{RhetoricalRegister Value:} Colloquial \\
\textbf{RhetoricalRegister Explanation:} The speaker employs informal language and relatable anecdotes, connecting with the audience on a personal level. \\[4pt]

\textbf{RelationshipToParty Value:} Diplomatic Bridge \\
\textbf{RelationshipToParty Explanation:} Acts as a mediator between rival factions within the party, easing tensions and fostering communication. \\[4pt]

\textbf{AttitudeToInstitutions Value:} Abolitionist \\
\textbf{AttitudeToInstitutions Explanation:} Sees some institutions as illegitimate; prefers replacement. \\[4pt]

\textbf{CommitteeFocus Value:} Environment/Energy \\
\textbf{CommitteeFocus Explanation:} Net-zero, energy security, industrial transition. \\[4pt]

\textbf{ConstituencyType Value:} Post-industrial town \\
\textbf{ConstituencyType Explanation:} Regeneration, skills, ‘left behind’ narratives. \\[4pt]

\textbf{DonorLobbyExposure Value:} Environmental activist supporters \\
\textbf{DonorLobbyExposure Explanation:} With backing from environmental groups, initiatives often emphasize sustainability and climate-related priorities. \\[4pt]

\textbf{ScandalVulnerability Value:} Technological Dependency \\
\textbf{ScandalVulnerability Explanation:} Reliance on specific technology or platforms can lead to a vulnerability if they are compromised or become obsolete. \\[4pt]

\textbf{MediaFootprint Value:} National media figure \\
\textbf{MediaFootprint Explanation:} High scrutiny; disciplined messaging. \\[4pt]

\textbf{LeadershipAmbition Value:} Grassroots Champion \\
\textbf{LeadershipAmbition Explanation:} Pursues leadership by empowering and engaging local communities to build widespread grassroots support. \\[4pt]

\textbf{CoalitionPosture Value:} Bridge Connector \\
\textbf{CoalitionPosture Explanation:} Acts as an intermediary who builds personal rapport with divergent groups to align interests. \\[4pt]

\textbf{ConstituencyServiceStyle Value:} Public services advocate \\
\textbf{ConstituencyServiceStyle Explanation:} Champions hospitals, schools, policing; delivery focus. \\[4pt]

\textbf{Appearance Category:} Robust Functional Outfitting \\
\textbf{Media Disposition:} Meme Commentator \\[4pt]

\textbf{Policy Stances:} \\
AIRegulation: ai on the job training emphasis \\
AgricultureAndRural: digital alternative over transport \\
BrexitAndEurope: rejoin eu liberaldem view \\
ClimateAndEnergy: pursue adaptation and voluntary aims \\
DefenceAndSecurity: nato strong support \\
DevolutionAndUnion: regional independence support \\
DigitalPrivacy: public support encryption \\
Education: streamline subject focus \\
ElectoralReform: fptp retention \\
ForeignPolicy: diverse global relations \\
FreeSpeechAndCulture: media ownership diversification \\
HousingAndPlanning: streamlined permission urban \\
Immigration: prioritize domestic skill development \\
NHSAndSocialCare: funding incentivize efficiency \\
PolicingAndJustice: voluntary bwc pilot \\
ReligiousFreedomAndSecularism: expression over regulation \\
TaxationAndSpending: control debt reduction \\
TradeAndIndustry: sustainability neglected in trade \\
Transport: careful scaling up \\
WelfareAndBenefits: set percentage increase apart \\
\\[4pt]

\textbf{Speech Issues:} [Transport, HousingAndPlanning, ForeignPolicy] \\[4pt]

\textbf{Speech:} Ladies and gentlemen, today we stand at a pivotal moment in our journey. As we strive for seamless transport solutions, let's embrace electric vehicles and build a charging infrastructure to match. It's about ensuring the roads of tomorrow are driven by cleaner, smarter choices. The dream doesn't stop here! We must reform urban planning policies, ensuring we're not shackling growth with outdated bureaucratic processes. People want homes where they need them—let’s provide that with expedience! Finally, in this post-Brexit world, we can take charge—building diverse economic partnerships is our path forward. Let's seize these global opportunities together, for we owe it to our constituents to foster a promising, innovative future! \\[4pt]

\textbf{Appearance:} Frederick sported utility-driven simplicity, presenting a professional but unassuming exterior. His attire, impeccably tailored, conveyed an air of competence, balanced against a broad shoulder frame that conveyed resilience. \\[4pt]

\textbf{Educational Vocational History:} Educated at comprehensive schools, Frederick cultivated a broad understanding of social dynamics. His further education at Oxbridge opened doors to elite networks, which shaped his approach as both lawyer and politician. \\[4pt]

\textbf{Medical Developmental History:} Growing up in a culturally rich yet challenging environment, Frederick developed a thick skin and an appetite for resilience, critical traits bolstered by a supportive albeit complex familial background. \\[4pt]

\textbf{Family History:} Raised within a nurturing household, Frederick's experiences fostered strong moral values. Although unmarried, his dedication to family remained firm, providing a network of influence in his pursuits. \\[4pt]

\bottomrule
\end{tabularx}
\end{table*}

\begin{table*}[ht]
\centering
\caption{Synthetic Parliamentary Persona Example for Archetype = The Culture-Warrior - Continued}
\label{tab:parliamentry-persona-example-6}
\scriptsize 
\begin{tabularx}{\textwidth}{X}
\toprule
\textbf{Synthetic Parliamentary Persona Example for Archetype = The Culture-Warrior - Continued} \\
\midrule
\textbf{Presenting Problems:}
– Interpersonal distrust following adversarial encounters \\
– Dissonance in legislative interpretations amidst diverse group discussions \\
– Preoccupation with societal engagement dynamics \\[4pt]

\textbf{Thought Content:} Frederick's thought processes revolve around sustainable futures and active community involvement. There exists a nuanced contemplation of personal principles clashing with party expectations, emphasizing grassroots input and inclusiveness. \\[4pt]

\textbf{Insight Judgment:} Demonstrates sharp cognitive insight into socio-political dynamics. While primarily pragmatic, occasionally struggles with interpersonal implications amidst direct assertiveness. \\[4pt]

\textbf{Cognition:} Displays a strong capacity for logical reasoning with enhanced analytical skills, though stress can impede recall and organizational flexibility. \\[4pt]

\textbf{Emotional Behavioral Functioning:} Handles tension by redirecting energy toward community initiatives, often expressing empathy in chaotic situations, but can default to irritability when principles are compromised. \\[4pt]

\textbf{Social Functioning:} Engages closely with constituents, fostering trust within diverse groups, but maintains a careful balance to guard personal emotional spaces, sometimes appearing distant. \\[4pt]

\textbf{Summary of Psychological Profile:} Frederick Touchstone exemplifies a complex integration of personal conviction and party loyalty shaped through diverse socio-political landscapes. His foundational experiences, enriched by comprehensive education and adversarial legal training, equip him with both negotiation skills and legislative fervor. Frederick demonstrates relational barriers stemming from his past confrontations, manifesting a distrust of political counterparts—a pattern that complicates his effectiveness in advocacy roles. However, he is also marked by an exceptional insight into broader societal needs, with community orientation presenting itself as a powerful compensatory strategy. Despite a proclivity to irritability when feeling betrayed or ignored, Frederick remains resilient, tirelessly channeling energy into projects advocating for sustainability and socio-economic equitability. \\[4pt]

\textbf{Political Relations:} Frederick operates as a guiding hand among his parliamentary peers, frequently facilitating dialogues that foster cooperation and clarity. His reputation for diplomacy often softens even the harshest opposition within the House, gathering support across the aisle. \\

\bottomrule
\end{tabularx}
\end{table*}

\begin{table*}[ht]
\centering
\caption{Synthetic Parliamentary Persona Example for Archetype = The Constituency Champion}
\label{tab:parliamentry-persona-example-7}
\scriptsize 
\begin{tabularx}{\textwidth}{X}
\toprule
\textbf{Synthetic Parliamentary Persona Example for Archetype = The Constituency Champion} \\
\midrule
\textbf{Name:} Joshua Tamayo \\
\textbf{Age:} 40 \\
\textbf{Sex:} Male \\
\textbf{Marital Status:} married present \\
\textbf{Ethnic Background:} mexican \\
\textbf{Party Affiliation:} Labour \\
\textbf{Occupation:} Local Councillor \\
\textbf{Education Secondary:} Grammar \\
\textbf{Education Tertiary:} Oxbridge \\
\textbf{Region:} Northern Ireland \\
\textbf{Parliamentary Style:} Policy Advocate \\[4pt]

\textbf{RhetoricalRegister Value:} Moralising \\
\textbf{RhetoricalRegister Explanation:} Frames issues as right vs wrong; appeals to conscience and values. \\[4pt]

\textbf{RelationshipToParty Value:} Outspoken Critic \\
\textbf{RelationshipToParty Explanation:} Openly challenges party leadership with bold questions, advocating for transparency and reform. \\[4pt]

\textbf{AttitudeToInstitutions Value:} Community-first Advocate \\
\textbf{AttitudeToInstitutions Explanation:} Believes that community needs should always have precedence, supporting local, flexible forms over fixed larger institutions. \\[4pt]

\textbf{CommitteeFocus Value:} Housing \& Urban Development \\
\textbf{CommitteeFocus Explanation:} Works to create affordable housing solutions and improve urban living standards. \\[4pt]

\textbf{ConstituencyType Value:} Inner-city neighborhood \\
\textbf{ConstituencyType Explanation:} A densely populated urban area characterized by multicultural communities and socioeconomic challenges like poverty and housing. \\[4pt]

\textbf{DonorLobbyExposure Value:} Minimal donor exposure \\
\textbf{DonorLobbyExposure Explanation:} Low external influence; relies on grassroots and local party. \\[4pt]

\textbf{ScandalVulnerability Value:} Relational Trust Issues \\
\textbf{ScandalVulnerability Explanation:} Interactions relying on unreliable partners or advisers create vulnerabilities when those ties face strain or exposure. \\[4pt]

\textbf{MediaFootprint Value:} Live Stream Personality \\
\textbf{MediaFootprint Explanation:} Engages directly with an audience through frequent live videos on platforms like Twitch and YouTube. \\[4pt]

\textbf{LeadershipAmbition Value:} Ministerial ladder \\
\textbf{LeadershipAmbition Explanation:} Promotion track; loyalty and caution rewarded. \\[4pt]

\textbf{CoalitionPosture Value:} Delegate Relayer \\
\textbf{CoalitionPosture Explanation:} Skilfully delegates roles and coordinates communication pathways to clarify mutual goals across factions. \\[4pt]

\textbf{ConstituencyServiceStyle Value:} Health Access Proponent \\
\textbf{ConstituencyServiceStyle Explanation:} Concentrates on improving healthcare access, focusing on marginalized community needs within the constituency. \\[4pt]

\textbf{Appearance Category:} Eco-conscious Fabrics \\
\textbf{Media Disposition:} Cultural Critic \\[4pt]

\textbf{Policy Stances:} \\
AIRegulation: data privacy high standards \\
AgricultureAndRural: subsidize traditional maintain rural jobs \\
BrexitAndEurope: be moderated basis projection \\
ClimateAndEnergy: large scale government investment \\
DefenceAndSecurity: renew trident support \\
DevolutionAndUnion: power balance enhance devolved \\
DigitalPrivacy: govt surveillance enhancement \\
Education: funding expansion allocation \\
ElectoralReform: mandatory voting \\
ForeignPolicy: mixed policy balancing \\
FreeSpeechAndCulture: regulating online non intervention \\
HousingAndPlanning: market led sustainability \\
Immigration: support inclusive refugee policies \\
NHSAndSocialCare: staffing hire more staff \\
PolicingAndJustice: diversity training support \\
ReligiousFreedomAndSecularism: faith based management \\
TaxationAndSpending: infrastructure productivity \\
TradeAndIndustry: encourage greener industry \\
Transport: technology investment focus \\
WelfareAndBenefits: no to tougher sanctions \\
\\[4pt]

\textbf{Speech Issues:} [BrexitAndEurope, DevolutionAndUnion, Education] \\[4pt]

\textbf{Speech:} Honourable Members, today, as we navigate the waters shaped by Brexit, it is crucial we look beyond our borders while responding to the pressing needs of our economy. We must structure a post-Brexit immigration policy that not only balances control with sector demands but holds a higher standard of coherence and compassion. Let us ensure credible enforcement of our laws, coupled with clear and predictable legal routes, cultivating public trust and an economy robust enough to withstand international uncertainties. Turning to education, I implore us to rethink our allocation of resources. A profound increase in education funding is necessary, particularly focused on reducing class sizes and refining educational materials. It is not a privilege, but an inalienable right that every child receives a conducive environment for growth. In terms of our relationship between central and devolved governments, we must embark on a journey towards equity. The balance of power must evolve. Let’s advance increased devolved powers to genuinely address the unique circumstances of our local communities. Only then can we truly fulfill the needs of those we represent. Together, I beseech you, let us champion our communities, remaining steadfast in our ethical duties and honoring the rich diversity of our nation. \\[4pt]

\textbf{Appearance:} His attire reflects a commitment to sustainability, woven from eco-conscious fabrics, ensuring his look is not just distinguished but also responsible. Joshua opts for bespoke tailoring with an emphasis on simplicity, favoring styles that suggest a deep respect for his roots rather than ostentation. \\[4pt]

\textbf{Educational Vocational History:} Education at a prestigious grammar school led Joshua to Oxbridge, equipping him with elite social networks and rhetorical skills. These tools shaped his local governance and advocacy work as a dedicated local councillor, providing a sound foundation for his ascent in politics. \\[4pt]

\textbf{Medical Developmental History:} Growing up in a close-knit family in an under-resourced area gave him insight into the developmental struggles faced by many. This background inspires his deep commitment to bettering health and living conditions for the constituencies he serves. \\[4pt]

\textbf{Family History:} Raised in a supportive Mexican immigrant family, Joshua values collectivism. His marriage reinforces his stability, mirroring his commitment to community building as they jointly nurture the bonds of family and purpose. \\[4pt]

\bottomrule
\end{tabularx}
\end{table*}

\begin{table*}[ht]
\centering
\caption{Synthetic Parliamentary Persona Example for Archetype = The Constituency Champion - Continued}
\label{tab:parliamentry-persona-example-8}
\scriptsize 
\begin{tabularx}{\textwidth}{X}
\toprule
\textbf{Synthetic Parliamentary Persona Example for Archetype = The Constituency Champion - Continued} \\
\midrule
\textbf{Presenting Problems:}
– Subclinical depressive symptoms with anhedonic features \\
– Perfectionistic tendencies contributing to burnout risk \\
– Interpersonal distrust following repeated adversarial encounters \\[4pt]

\textbf{Thought Content:} His mental narrative frequently revolves around community obligations, role efficacy, and a pervasive sense of justice that is occasionally derailed by disillusionment with political processes and betrayal from allies. \\[4pt]

\textbf{Insight Judgment:} Joshua maintains acute awareness of community needs; however, he wrestles with self-judgment when reflecting on unfulfilled ambitions. Despite showing significant progress, some past experiences lead to critical self-appraisal. \\[4pt]

\textbf{Cognition:} He displays robust cognitive functions, quickly grasping legislative details but falters under stress, finding it increasingly challenging to sustain prolonged attention during negotiations or when events spiral out of control. \\[4pt]

\textbf{Emotional Behavioral Functioning:} Despite a moral fortitude in his activism, Joshua grapples with moments of overwhelming responsibility. He confronts adverse feelings by channeling them into actionable advocacy, fostering resilience that inspires others, albeit leading to eventual fatigue. \\[4pt]

\textbf{Social Functioning:} His relationships within local politics manifest warmth and approachable charisma. Nevertheless, underlying tensions surface as Joshua learns to navigate trust in a professional landscape replete with betrayal and fluctuating loyalties. \\[4pt]

\textbf{Summary of Psychological Profile:} Joshua presents as a conscientious advocate shaped by his formative experiences, skillfully integrating a strong community focus with moral imperatives that inform his policy preferences. The confluence of subclinical depressive symptoms with underlying perfectionistic traits positions him at risk for burnout, particularly when coupled with difficulties in establishing trustworthy relationships. His rich cognitive capacities lend him the ability to operate within complex legislative contexts, yet he can falter when social and political landscapes shift unexpectedly. Awareness of his community’s injustices fuels his quest for meaningful engagement, but it simultaneously evokes challenges surrounding self-doubt and frustration in achieving systemic change. Ultimately, Joshua's resolve remains unwavering, seeking to impact society positively despite these struggles. \\[4pt]

\textbf{Political Relations:} Joshua interacts with fellow parliamentarians with sincerity but occasionally faces conflict due to his outspokenness. He advocates for a transparent dialogue among peers, passionately driving conversations that align with community-centered reform. With constituents, he cultivates trust through tangible actions, enhancing direct accessibility and accountability while countering skeptical views about party integrity. \\

\bottomrule
\end{tabularx}
\end{table*}

\begin{table}[htbp]
\centering
\caption{Comparison of LLM-as-a-judge assessments of Parliamentary Personas using Anthropic and OpenAI models}
\label{tab:parliamentary_persona_judge}
\begin{tabular}{lccc}
\toprule
\textbf{Metric} & \multicolumn{2}{c}{\textbf{Mean judge scores}} & \textbf{Average absolute difference} \\
\toprule
 & \textbf{Anthropic} & \textbf{OpenAI} & \\
\midrule
Clarity & 4.310 & 4.990 & 0.680 \\
Originality & 3.890 & 4.070 & 0.180 \\
Coherence & 3.770 & 5.000 & 1.230 \\
Diversity & 4.020 & 4.100 & 0.300 \\
Realism & 3.600 & 5.000 & 1.400 \\
Psychological Depth & 4.650 & 5.000 & 0.350 \\
Consistency & 3.530 & 5.000 & 1.470 \\
Informativeness & 4.560 & 5.000 & 0.440 \\
Ethical Considerations & 4.610 & 5.000 & 0.390 \\
Demographic Fidelity & 3.920 & 5.000 & 1.080 \\
Overall Score & 3.910 & 5.000 & 1.090 \\
\midrule
\textbf{Overall Mean} & \textbf{4.070} & \textbf{4.833} & \textbf{0.783} \\
\bottomrule
\end{tabular}
\begin{tablenotes}
\small
\item Note: Analysis based on 100 parliamentary persona reviews.
\end{tablenotes}
\end{table}

\section{Future Work}
\label{app:future_work}

We can extend the current work in several directions. First, evaluating fine-tuned language models on the synthetic SJT and persona datasets could provide insights into how model adaptations affect trait alignment, behavioral predictions, and bias patterns. Second, ablation studies on the SJT and persona attributes, such as removing specific demographic, behavioral, or cognitive features, can reveal the relative contribution of each attribute to trait predictability and persona realism. Third, a more comprehensive statistical analyses and factor analyses at both the aggregate and item levels could deepen understanding of latent relationships between SJTs, persona traits, and HEXACO dimensions, enabling identification of archetypal patterns and cross-trait interactions. Third, learning and applying steering vectors from our datasets might produce more reliable behaviors, especially over long conversations, whereas prompt based conversation control will suffer from context forgetting effects. Finally, integrating independent expert validation or multi-model comparisons would strengthen the reliability of persona classifications and trait scoring, improving the robustness of future experiments.

\section{Models Used}
\label{app:models_used}

We have used the following models for running our experiments, data generation and analysis pipelines:

\begin{enumerate}
    \item GPT 4.1 \citep{openai2024gpt4technicalreport}
    \item GPT 4.1-mini \citep{openai2024gpt4technicalreport}
    \item Qwen 3-0-6B Embedding Model \citep{qwen3}
    \item Qwen 2.5-7B-Instruct Model \citep{qwen2025qwen25technicalreport}
    \item Llama 3.1-8B-Instruct Model \citep{grattafiori2024llama3herdmodels}
    \item Gemma 3-4b-Instruct Model \citep{gemmateam2025gemma3technicalreport}
\end{enumerate}

\section{Annotators}
\label{app:use_of_annotators}
We have not employed independent expert raters to cross-validate score rubrics or persona classifications beyond the expert feedback and review by our authors, due to the high expertise demand of psychometrics validation. Future work could include hiring such independent raters to enhance robustness and minimize potential bias.

\section{Use of AI Assistance}
\label{app:use_of_ai}
In accordance with ACL policies, we disclose that we used AI-assisted tools (i.e., ChatGPT) for drafting text, refining phrasing, organizing sections, and suggesting non-novel code snippets. All technical content, experiments, analyses, and claims were authored, verified, and are the sole responsibility of the human authors. We reviewed and corrected any tool-generated output for factual accuracy, methodological soundness, and citation correctness; we also tested and validated all code used in experiments. No proprietary, confidential, or personal data were provided to AI tools, and no text or code was accepted without human review. The authors assume full responsibility for all errors.

\end{document}

%% file: arxiv_files/analysis.tex
\section{Alignment Between SJT and Self-Report \hexaco\ Trait Representations}
\label{sec:alignment}
We examine how behavioral traits expressed in synthetic Situational Judgment Test (SJT) responses align with underlying personality dimensions from the \hexaco\ framework. By clustering persona representations across both behavioral and self-reported trait spaces, it evaluates whether consistent archetypal structures emerge and how well simulated decision-making reflects trait-level psychological constructs. The study also explores differences between base-model behavior and rich persona-conditioned responses across archetypes and demographics, providing insight into the alignment between behavioral outputs and personality structure. Responses for this analysis were generated using GPT-4.1.


\subsection{Aggregate-Level Cluster Comparison}

At the aggregate level, persona embeddings derived from SJT responses and \hexaco\;  scores are independently clustered. We then compare these clustering structures to examine whether similar archetypal groupings appear across both modalities. 

At the aggregate level, \hexaco\ representations form more well-separated clusters, whereas SJT-derived representations are more distributed. This suggests LLMs more reliably reproduce canonical, self-reported trait structures than behaviorally grounded decision patterns, supporting the use of SJTs as a more sensitive probe of personality-driven behavior.

The Avoider archetype exhibits a distinct preference for \textcolor{hexC}{Conscientiousness}-aligned options (46\% vs.\ 33\% overall), whereas most other archetypes predominantly select \textcolor{hexH}{Honesty--Humility}-aligned responses.

\begin{figure}[H]
\centering
\includegraphics[width=0.9\columnwidth]{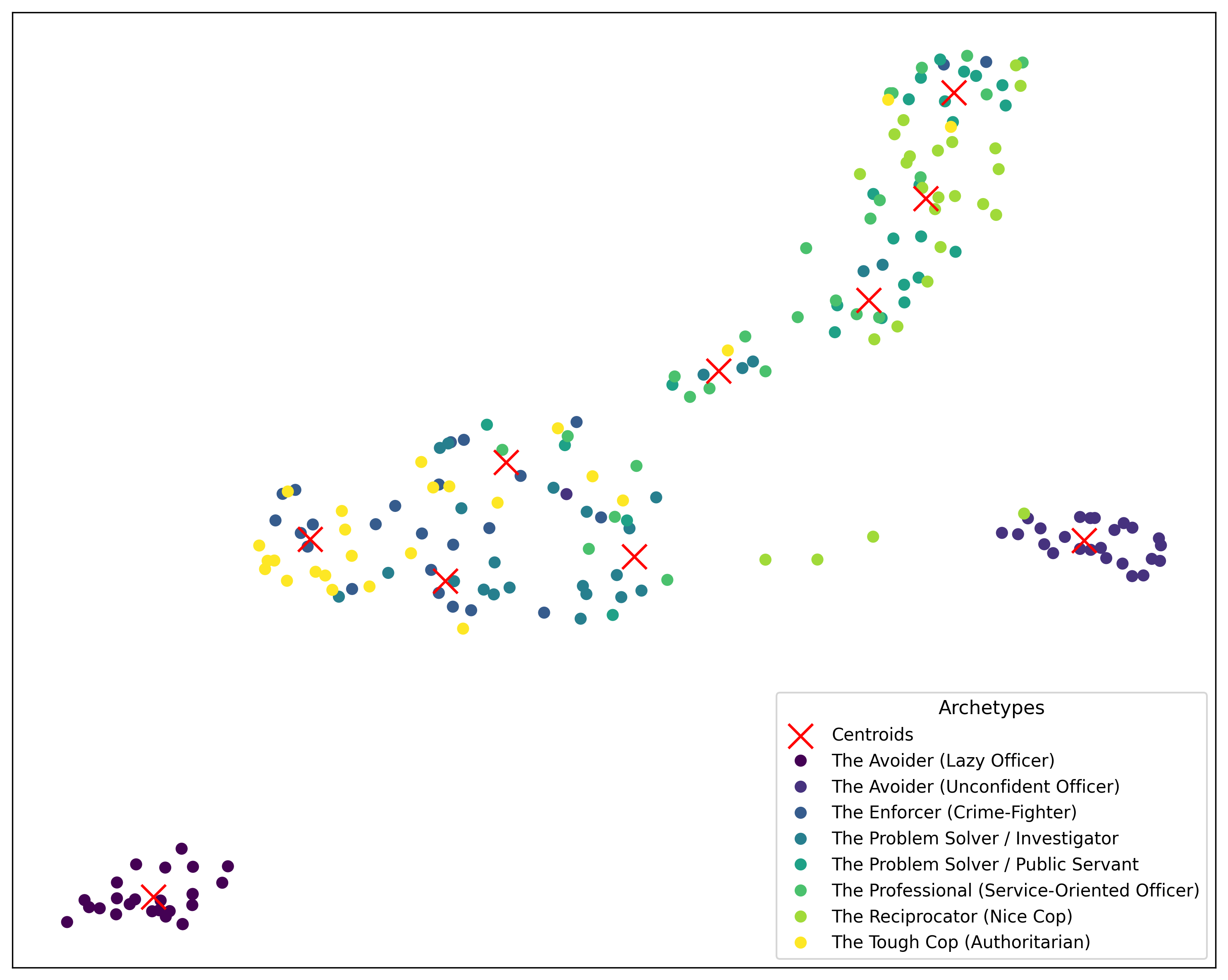}
\caption{Clusters using \hexaco\;  Answers, colored by archetypes}
\end{figure}

\begin{figure}[H]
\centering
\includegraphics[width=0.9\columnwidth]{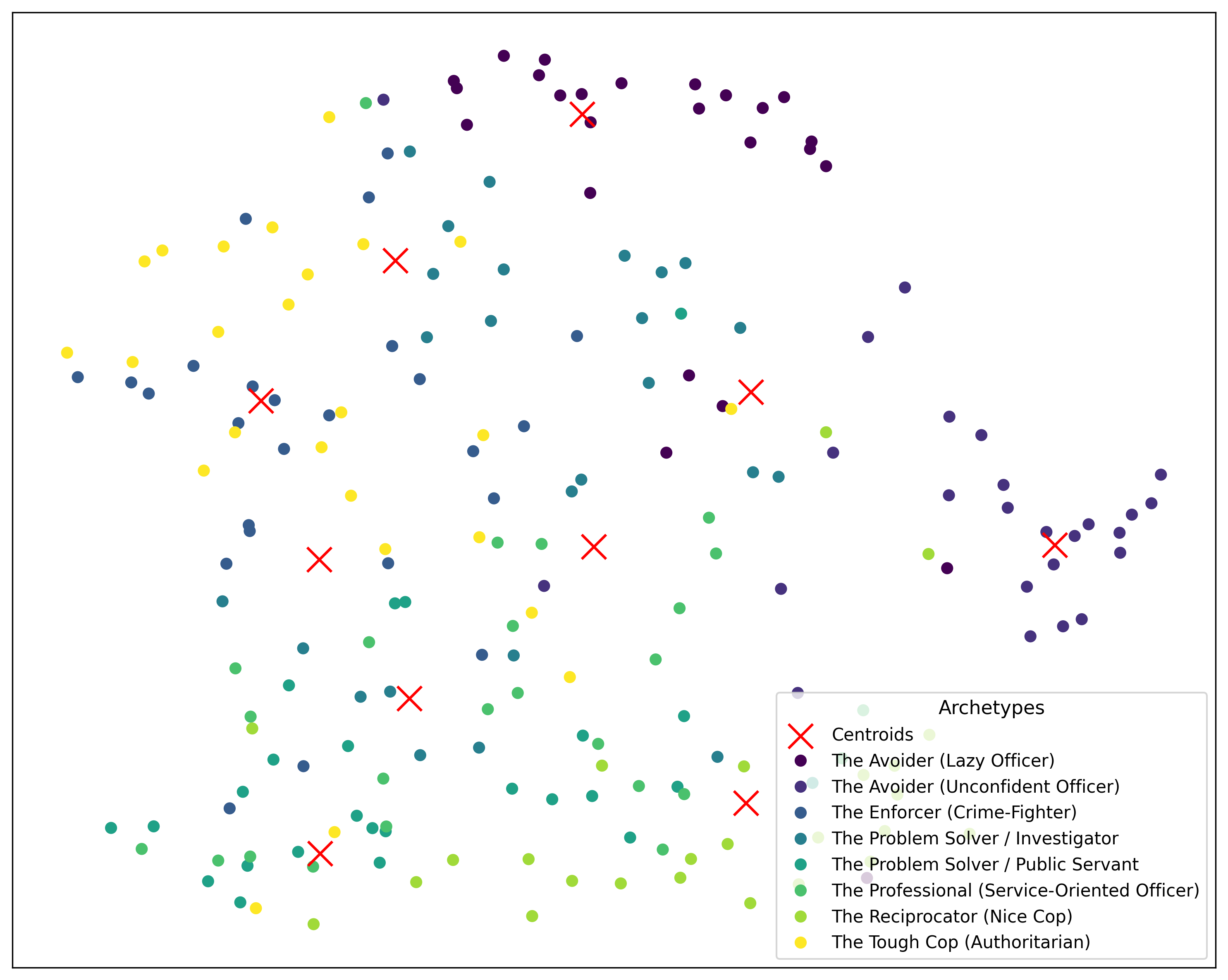}
\caption{Clusters using SJT Answers, colored by archetypes}
\end{figure}

\subsection{Trait Distributions and Preferences}

Similar to the archetype level analyses presented earlier, we extended our examination to additional persona attributes such as ethnic background, age, and gender. Notably, personas representing Colombian, East Asian, Hispanic/Latino, and Southeast Asian backgrounds exhibited a remarkably higher preference for the \textcolor{hexC}{Conscientiousness}-aligned options, whereas most other groups predominantly favored \textcolor{hexH}{Honesty–Humility}-aligned choices. Specifically, these ethnic groups selected \textcolor{hexC}{Conscientiousness} options at approximately \textasciitilde52\%, more than twice the overall population mean of \textasciitilde24\%.

This observation suggests a potential trait inclination bias associated with certain minority ethnic representations, warranting further analysis into how demographic conditioning may influence model-driven personality expression and decision tendencies.

\section{Construct validity between SJT and Self-Report \hexaco\; Traits}
\label{sec:factor_experiments}

To investigate the psychometric alignment between the psychometric Hexaco scores assigned via SJT vs those from the base \hexaco\ instrument, we conduct a correlation and regression modeling. For this analysis, Qwen2.5-7B was used to generate responses for
a larger set of 1504 personas across 500 SJT items.

\subsection{Aggregate-Level Correlation Analysis}

At the aggregate level, we compute the Pearson correlation coefficient ($r$) between average SJT scores (aggregated across all scenarios for a given persona) and the corresponding \hexaco\;  trait scores. The SJT scores are calculated as the proportion of answers answered per trait and the \hexaco\;  scores are calculated following the standard \hexaco\;  Scoring Guide \citep{lee2018psychometric}. This step evaluates whether global behavioral tendencies captured by SJTs align with expected personality gradients across the six \hexaco\;  dimensions — \textcolor{hexH}{Honesty-Humility}, \textcolor{hexE}{Emotionality}, \textcolor{hexX}{eXtraversion}, \textcolor{hexA}{Agreeableness}, \textcolor{hexC}{Conscientiousness}, and \textcolor{hexO}{Openness to Experience}.

Highest positive correlation (\textasciitilde 0.504) is for the \textcolor{hexX}{eXtraversion} trait, indicating that its the most accurately answered trait in SJTs. \textcolor{hexH}{Honesty-Humility} is the most commonly chosen trait, sometimes even when the trait is absent, leading to the only negative correlation (\textasciitilde -0.12)

Importantly, a high overall $R^2$ can arise even when the correlation for individual \hexaco\ scores is not high with the trait SJT scores, as the combined linear contribution of multiple traits (potentially in different directions) can explain substantially more variance than any single predictor alone. This multivariate structure highlights that while individual \hexaco\ score vs SJT score correlations may be weak, the joint configuration of traits captures meaningful patterns of behavioral predictability, providing a complementary measure of construct validity.

\begin{table}[H]
\centering
\caption{Trait wise Pearson R Correlation between SJTs and \hexaco\; }
\begin{tabularx}{\columnwidth}{|X|c|}
\hline
\textbf{Trait} & \textbf{Pearson R Correlation} \\
\hline
\textcolor{hexH}{Honesty-Humility} & -0.122 \\
\hline
\textcolor{hexE}{Emotionality}  & 0.216 \\
\hline
\textcolor{hexX}{eXtraversion} & 0.504 \\
\hline
\textcolor{hexA}{Agreeableness} & 0.329 \\
\hline
\textcolor{hexC}{Conscientiousness} & 0.277 \\
\hline
\textcolor{hexO}{Openness} & 0.343 \\
\hline
\end{tabularx}
\end{table}

\subsection{Trait Regression Analysis}
\label{app:regression}
\begin{table}[!htb]
\centering
\caption{Summary of Trait-Wise Linear Regression Models}
\label{tab:regression_analysis}
\renewcommand\arraystretch{1.2}

\begin{tabularx}{\columnwidth}{@{}Xcc@{}}
\toprule
\textbf{Trait} & \textbf{R-Squared} & \textbf{Adj.~R-Squared} \\
\midrule
\textcolor{hexH}{Honesty--Humility}   & 0.993 & 0.993 \\
\textcolor{hexE}{Emotionality}        & 0.864 & 0.863 \\
\textcolor{hexX}{Extraversion}        & 0.870 & 0.869 \\
\textcolor{hexA}{Agreeableness}       & 0.941 & 0.940 \\
\textcolor{hexC}{Conscientiousness}   & 0.940 & 0.939 \\
\textcolor{hexO}{Openness}            & 0.977 & 0.977 \\
\bottomrule
\end{tabularx}
\end{table}

To aggregate and quantify the explanatory power of personality traits on SJT performance, we train a linear regression model using \hexaco\; scores as input predictors and aggregated SJT scores as the dependent variable. Responses for this analysis were collected using Qwen 2.5-7B Instruct Model \citep{qwen2025qwen25technicalreport}. Across all traits, the models yield consistently high adjusted $R^2$ values (\textasciitilde 0.8--0.9) as seen in Table \ref{tab:regression_analysis} with generally positive coefficients, particularly for the focal trait being predicted. 

\section{Trait-Behavior Benchmark Coherence}\label{app:trait_behaviour_benchmark}

\begin{figure}[ht]
\centering
\includegraphics[width=0.9\columnwidth]{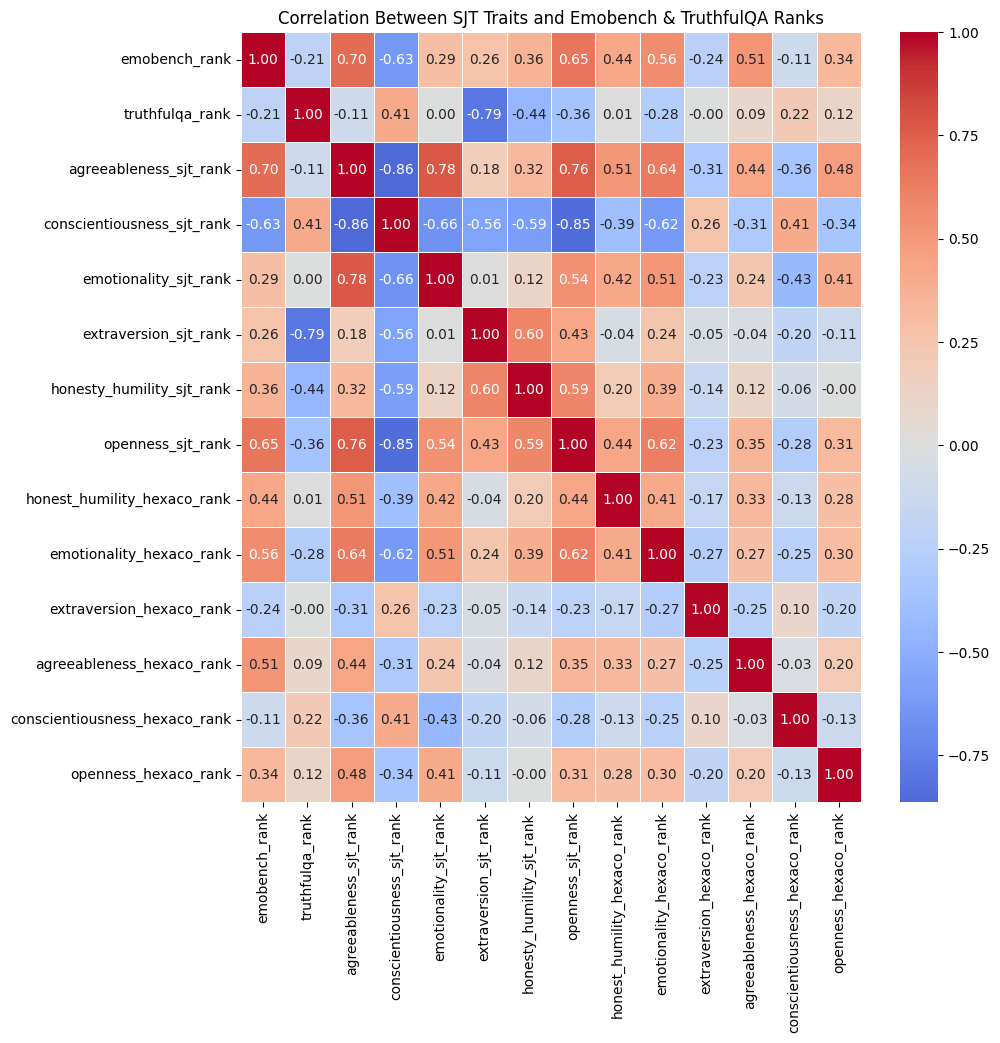}
\caption{Correlation between \hexaco\; traits and benchmarks}
\label{fig:trait_benchmark_corr}
\end{figure}

For evaluating Trait-Behaviour coherence with each other, we calculated Spearman correlations for ranks across 500 personas (Figure \ref{fig:trait_benchmark_corr}). It reveals that SJT is a stronger predictor of external benchmark performance than \hexaco\; self-report scores, with SJT trait ranks showing consistently higher correlations with both EmoBench and TruthfulQA. EmoBench performance is broadly trait-driven, correlating positively with warm and open traits, most strongly \agreeableness\; SJT ($0.70$), \openness\; SJT (0.65), and \emotionality\; HEXACO (0.56), while TruthfulQA shows the opposite pattern, correlating negatively with social and expressive traits, most strongly \extraversion\; SJT ($-0.79$), suggesting that less interpersonally assertive personas are associated with higher truthfulness. \conscientiousness\; SJT emerges as a suppressor variable, negatively correlating with nearly all other traits, including \agreeableness\; SJT ($-0.86$) and \openness\; SJT ($-0.85$), while predicting lower EmoBench ($-0.63$) and higher TruthfulQA (0.41) performance, reflecting a rule-bound behavioral orientation that trades off against warmth and creativity.

At the structural level, SJT traits organise into two coherent clusters that map cleanly onto the benchmark split: a warm-expressive cluster (\agreeableness\;, \openness\;, \emotionality\;, \honestyhumility\;) that predicts higher EmoBench and lower TruthfulQA, and a task-assertive cluster (\conscientiousness\;, \extraversion\;) that predicts the reverse (Table \ref{tab:cluster_sjt_trait}). This two-cluster structure emerges organically from the correlation patterns and aligns with the known \hexaco\; higher-order factor structure, lending structural validity to the SJT instrument. A notable exception is \hexaco\; \extraversion\;, which correlates negatively with EmoBench $-0.24$) (contrary to trait-theoretic expectations) and shows near-zero correlation with TruthfulQA and weak alignment with its SJT counterpart, warranting further examination as a divergent validity concern.

\begin{table}[H]
\centering
\caption{SJT vs Trait Behaviour Clusters}
\label{tab:cluster_sjt_trait}
\begin{tabularx}{\columnwidth}{|X|X|c|c|}
\hline
\textbf{Cluster} & \textbf{Trait} & \textbf{EmoBench (r)} & \textbf{TruthfulQA (r)} \\
\hline
\multirow{4}{*}{Warm-Expressive} 
& Agreeableness (SJT)     & +0.70 & -0.11 \\
& Openness (SJT)          & +0.65 & -0.36 \\
& Emotionality (SJT)      & +0.29 & +0.00 \\
& Honesty-Humility (SJT)  & +0.36 & -0.44 \\
\hline
\multirow{2}{*}{Task-Assertive} 
& Conscientiousness (SJT) & -0.63 & +0.41 \\
& eXtraversion (SJT)      & +0.26 & -0.79 \\
\hline
\end{tabularx}
\end{table}

\section{Personas as a Consistent Conditioning Construct}
\label{app:persona_consistent_construct}

To evaluate whether persona conditioning induces coherent, trait-consistent behavioral shifts, we compare persona-conditioned model responses against a base model baseline across four instruments: a synthetic Situational Judgment Test (SJT) mapped to \hexaco\; facets, a self-report \hexaco\; questionnaire, and two external benchmarks - EmoBench and TruthfulQA. For each measure, we compute the z-scores relative to the persona distribution to quantify the direction and magnitude of deviation from baseline, and examine the results both at an aggregate level as well as across individual archetypes.

The analysis reveals two overarching patterns in how persona conditioning shapes model behavior (Table \ref{tab:basemodel_z_persona}). First, persona conditioning produces a consistent and theoretically coherent trade-off between emotional and epistemic performance: conditioned models score lower on EmoBench (z =$-1.06$) but higher on TruthfulQA (z = +1.13), and this TruthfulQA improvement holds universally across all eight archetypes regardless of character type, suggesting that persona framing activates a more identity-anchored response mode that increases truthfulness. Second, the SJT and \hexaco\; instruments reveal a systematic attitudinal-behavioral dissociation: traits such as \honestyhumility\; show large behavioral suppression in dilemma scenarios (SJT z = $-1.85$) while remaining comparatively stable in self-report (\hexaco\; z = $-0.47$), mirroring patterns well-documented in human psychometric research \citep{olaru2019situational} under role-induced demand characteristics.

At the archetype level, persona effects are discriminant rather than uniform, and the variation aligns with character definitions in theoretically expected ways (Table \ref{tab:archetype_z_base_model}). The Avoider (Lazy Officer) produces the most extreme deviations across emotional calibration (EmoBench $z = -5.22$), honesty in dilemma scenarios (\honestyhumility\; SJT $z = -5.42$), and social initiative (\extraversion\; SJT $z = -4.22$) — coherent with a disengaged, low-agency character — while the Reciprocator (Nice Cop) leads on \agreeableness\; (SJT $z = +2.02$) and \openness\; (SJT $z = +2.21$) behavioral scores, consistent with its socially warm character definition. \honestyhumility\; SJT scores are negative for every archetype without exception (range: -1.29 to -5.42), likely reflecting a domain-specific prior in the law enforcement archetype family that structurally favors authority-oriented choices over humility-signaling options. 

Lastly, to assess the stability persona-conditioning across repeated runs, we evaluate consistency using \textbf{Jensen--Shannon (JS) divergence} and the \textbf{Intraclass Correlation Coefficient (ICC)} across both benchmark performance and SJT-derived \hexaco\; trait distributions. EmoBench and TruthfulQA exhibit perfect distributional stability (JS divergence = 0) with high reliability (ICC = 0.95 and 0.90, respectively), while SJT trait distributions show only negligible variation across iterations (mean JS divergence = 0.003, $\sigma = 0.001$) alongside consistently high trait-wise ICC values.

\begin{table}[H]
\centering
\caption{Base Model Scores Relative to Persona Score Distributions}
\label{tab:basemodel_z_persona}
\begin{tabularx}{\columnwidth}{|X|c|c|c|c|}
\hline
\textbf{Metric} & \textbf{Persona Mean} & \textbf{Persona Std} & \textbf{Base Score} & \textbf{Z Score} \\
\hline

\multicolumn{5}{|c|}{\textbf{Benchmarks}} \\
\hline
Emobench & 0.330 & 0.084 & 0.420 & -1.058 \\
TruthfulQA & 0.450 & 0.040 & 0.399 & 1.133 \\
\hline

\multicolumn{5}{|c|}{\textbf{SJT Trait Scores}} \\
\hline
\textcolor{hexH}{Honesty-Humility} & 0.214 & 0.057 & 0.319 & -1.852 \\
\textcolor{hexE}{Emotionality} & 0.019 & 0.025 & 0.014 & 0.227 \\
\textcolor{hexX}{Extraversion} & 0.028 & 0.036 & 0.035 & -0.174 \\
\textcolor{hexA}{Agreeableness} & 0.123 & 0.146 & 0.042 & 0.556 \\
\textcolor{hexC}{Conscientiousness} & 0.563 & 0.210 & 0.576 & -0.065 \\
\textcolor{hexO}{Openness} & 0.052 & 0.040 & 0.014 & 0.962 \\
\hline

\multicolumn{5}{|c|}{\textbf{HEXACO Trait Scores}} \\
\hline
\textcolor{hexH}{Honesty-Humility} & 3.487 & 0.381 & 3.667 & -0.473 \\
\textcolor{hexE}{Emotionality} & 3.351 & 0.317 & 3.500 & -0.471 \\
\textcolor{hexX}{Extraversion} & 2.763 & 0.250 & 2.750 & 0.053 \\
\textcolor{hexA}{Agreeableness} & 3.247 & 0.306 & 3.250 & -0.011 \\
\textcolor{hexC}{Conscientiousness} & 3.080 & 0.493 & 2.500 & 1.175 \\
\textcolor{hexO}{Openness} & 3.242 & 0.416 & 3.750 & -1.222 \\
\hline

\end{tabularx}
\end{table}

\begin{table}[H]
\centering
\caption{Archetype-wise Z-Scores Relative to the Base Model}
\label{tab:archetype_z_base_model}
\begin{tabular}
{@{}p{2.5cm}p{1.5cm}p{1.5cm}p{1.5cm}p{1.5cm}p{1.5cm}p{1.5cm}p{1.5cm}p{1.5cm}@{}}
\hline
\textbf{Metric} &
\textbf{Problem Solver / Public Servant} &
\textbf{Enforcer (Crime-Fighter)} &
\textbf{Avoider (Unconfident Officer)} &
\textbf{Professional (Service-Oriented)} &
\textbf{Avoider (Lazy Officer)} &
\textbf{Problem Solver / Investigator} &
\textbf{Tough Cop (Authoritarian)} &
\textbf{Reciprocator (Nice Cop)} \\
\hline

\multicolumn{9}{|c|}{\textbf{Benchmarks}} \\
\hline
Emobench & -0.53 & -1.47 & -1.74 & -0.47 & -5.22 & -1.25 & -1.89 & -0.35 \\
TruthfulQA & 0.88 & 0.61 & 2.44 & 1.38 & 2.67 & 1.12 & 0.53 & 1.54 \\
\hline

\multicolumn{9}{|c|}{\textbf{SJT Trait Scores}} \\
\hline
\textcolor{hexA}{Agreeableness} & 0.96 & 0.08 & 0.68 & 1.22 & -0.02 & 0.08 & -0.27 & 2.02 \\
\textcolor{hexC}{Conscientiousness} & -0.84 & 0.22 & 0.10 & -0.71 & 2.13 & 0.64 & 0.58 & -1.56 \\
\textcolor{hexE}{Emotionality} & 0.26 & -0.52 & 0.90 & 0.30 & 0.23 & -0.90 & -1.46 & 1.00 \\
\textcolor{hexX}{Extraversion} & 0.10 & 0.41 & -3.94 & -0.27 & -4.22 & -0.63 & 0.51 & -0.52 \\
\textcolor{hexH}{Honesty-Humility} & -1.29 & -1.55 & -3.02 & -1.65 & -5.42 & -2.22 & -1.89 & -2.62 \\
\textcolor{hexO}{Openness} & 1.55 & 1.02 & 1.10 & 1.84 & 0.32 & 1.14 & 0.79 & 2.21 \\
\hline

\multicolumn{9}{|c|}{\textbf{HEXACO Trait Scores}} \\
\hline
\textcolor{hexA}{Agreeableness} & 0.23 & -0.44 & -0.05 & 1.03 & -0.22 & -0.03 & -0.44 & 0.68 \\
\textcolor{hexC}{Conscientiousness} & 0.84 & 1.37 & 1.06 & 1.48 & 1.37 & 1.43 & 1.28 & 0.85 \\
\textcolor{hexE}{Emotionality} & 0.10 & -0.61 & 0.08 & 0.00 & -1.62 & -0.78 & -0.84 & -0.03 \\
\textcolor{hexX}{Extraversion} & -0.04 & 0.23 & -0.30 & -0.34 & 0.38 & 0.30 & 0.33 & -0.21 \\
\textcolor{hexH}{Honesty-Humility} & -0.16 & -0.77 & -0.13 & 0.05 & -0.96 & -0.67 & -1.06 & -0.22 \\
\textcolor{hexO}{Openness} & -0.89 & -1.44 & -1.35 & -1.00 & -2.82 & -1.37 & -2.52 & -0.68 \\
\hline

\end{tabular}
\end{table}

\begin{figure}[H]
\centering
\includegraphics[width=0.9\columnwidth]{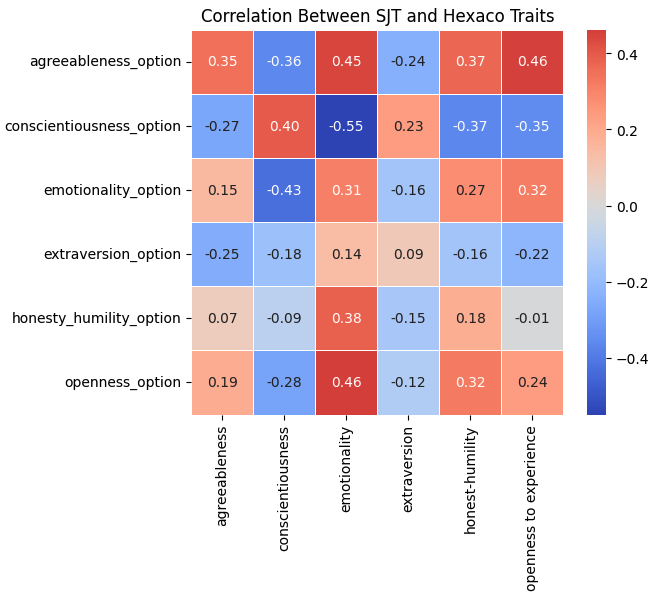}
\caption{Correlation between \hexaco\; trait scores and SJT scores}
\label{fig:sjt_hexaco_corr}
\end{figure}



\section{Psychometric Validation via Multidimensional IRT}\label{app:mirt}

We evaluated the psychometric properties of the situational judgment test (SJT) framework using a confirmatory multidimensional item response theory (MIRT) \citep{reckase200618} model aligned with the HEXACO personality structure.

Before fitting the MIRT model, we linked personas, SJT items, and generated responses using hash-based identifiers \texttt{persona\_hash} and \texttt{question\_hash}. For each persona--item pair, five independent responses were generated and mapped into one of six normalized HEXACO-aligned categories. To obtain a single response per persona-item pair for psychometric analysis, we collapsed these five samples to the modal category; ties were broken uniformly at random using a fixed seed. This produced a complete response table of 150{,}000 persona--item pairs, of which 8{,}006 (5.3\%) exhibited tied modal responses, indicating that most aggregated responses were stable across repeated sampling. The collapsed long-form response table was then reshaped into a persona-by-item response matrix for MIRT estimation.

Each SJT item presents six response options, each explicitly mapped to one of the six HEXACO traits (Agreeableness, Conscientiousness, Emotionality, Extraversion, Honesty--Humility, Openness). This allows responses to be modeled using a categorical (nominal) response formulation~\citep{bock1972estimating}, in which each option corresponds to a trait-specific utility.

Formally, for persona $p$ and item $j$, the probability of selecting response category $k \in \{1,\dots,6\}$ is given by:
\[
P(Y_{pj} = k) = \frac{\exp(\alpha_{jk} + a_j \cdot \theta_{p,d(k)})}{\sum_{k'} \exp(\alpha_{jk'} + a_j \cdot \theta_{p,d(k')})},
\]
where $\theta_p \in \mathbb{R}^6$ is the latent HEXACO trait vector for persona $p$, $a_j$ is an item-specific scalar discrimination parameter, $\alpha_{jk}$ are category-specific intercepts, and $d(k)$ maps each response option to its corresponding HEXACO trait dimension. In the present implementation, the response categories were encoded as $0,\dots,5$ with a fixed one-to-one mapping to the six HEXACO dimensions in the order: \agreeableness, \conscientiousness, \emotionality, \extraversion, \honestyhumility, and \openness. This induces a confirmatory structure in which each response category loads on a single predefined trait dimension.

For computational tractability, the model was estimated on a pilot subset of 250 personas and 150 SJT items, sampled uniformly without replacement from the full response table. Letting $P$ denote the number of personas, $J$ the number of items, $K=6$ the number of response categories, and $D=6$ the number of latent dimensions, the model was implemented in PyMC~\cite{pymc2023} as a Bayesian categorical-response MIRT model.

Persona latent trait vectors were assigned independent standard normal priors,
\[
\theta_p \sim \mathcal{N}(0, I_D),
\]
item discrimination parameters were constrained to be positive via half-normal priors,
\[
a_j \sim \mathrm{HalfNormal}(1),
\]
and item-category intercepts were assigned normal priors before being mean-centered within each item for identifiability,
\[
\alpha_{jk}^{\mathrm{raw}} \sim \mathcal{N}(0,1), \qquad
\alpha_{jk} = \alpha_{jk}^{\mathrm{raw}} - \frac{1}{K}\sum_{k'=1}^{K}\alpha_{jk'}^{\mathrm{raw}}.
\]

Given persona index $p(n)$ and item index $j(n)$ for observation $n$, the category utilities are
\[
\eta_{nk} = \alpha_{j(n)k} + a_{j(n)}\,\theta_{p(n),d(k)},
\]
and observed responses are modeled as
\[
Y_n \sim \mathrm{Categorical}\!\left(\mathrm{softmax}(\eta_{n1},\dots,\eta_{nK})\right).
\]

Posterior inference was performed using the No-U-Turn Sampler (NUTS)~\citep{hoffman2014no} with 8 chains, 300 warmup iterations, and 300 posterior draws per chain, with target acceptance probability set to 0.9. All parameters exhibit excellent MCMC convergence, with 100\% of $\hat{R}$ values in the range $1.00 \le \hat{R} \le 1.01$ (3{,}450/3{,}450 parameters), indicating stable and well-mixed posterior estimates. No divergent transitions were observed.

\subsection{Item discrimination}
\label{app:mirt_1}
\begin{figure}[h]
\centering
\includegraphics[width=0.75\linewidth]{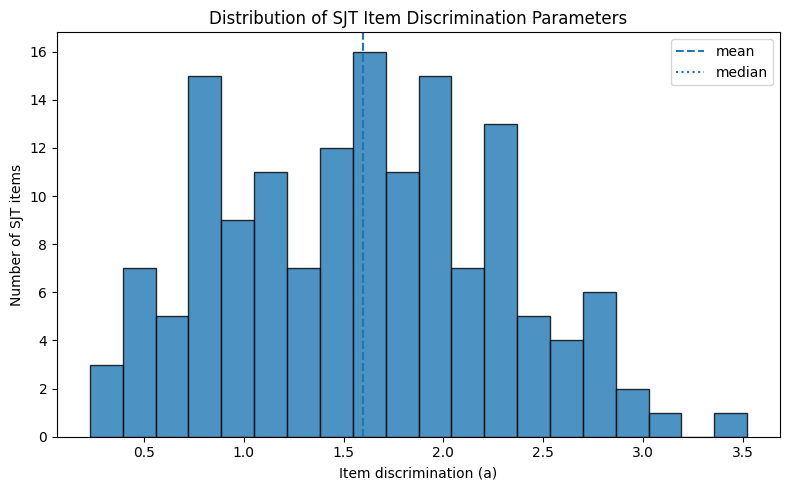}
\caption{
Distribution of item discrimination parameters ($a$) estimated by the MIRT model.
In multidimensional item response theory, the discrimination parameter quantifies how strongly a scenario differentiates between personas with different latent trait levels: higher values indicate that small differences in latent traits lead to larger differences in response probabilities.
The distribution is centered around moderate-to-high values (mean $\approx 1.6$), with most items exhibiting $a > 1.0$, indicating that the majority of SJT scenarios provide meaningful information about underlying behavioral tendencies. A small number of low-discrimination items correspond to scenarios that are less diagnostic and may be candidates for refinement.
}
\label{fig:mirt_discrimination_distribution}
\end{figure}

In this formulation, the discrimination parameter determines how strongly an SJT scenario amplifies differences in latent behavioral traits, with higher values indicating that the scenario more effectively distinguishes between personas with different trait profiles.
Table~\ref{tab:item_discrimination} and Figure~\ref{fig:mirt_discrimination_distribution} summarizes the distribution of item discrimination parameters. The majority of items exhibit moderate to strong discrimination, while a small subset of items provide little diagnostic information.

\begin{table}[h]
\centering
\begin{tabular}{lcc}
\hline
Discrimination range & Number of items & Percentage \\
\hline
$a < 0.5$ (very weak) & 4 & 2.7\% \\
$0.5 \le a < 1.0$ (weak--moderate) & 32 & 21.3\% \\
$1.0 \le a < 2.0$ (good) & 73 & 48.7\% \\
$2.0 \le a < 3.0$ (strong) & 38 & 25.3\% \\
$a \ge 3.0$ (very strong) & 3 & 2.0\% \\
\hline
Total & 150 & 100\% \\
\hline
\end{tabular}
\caption{Distribution of item discrimination parameters estimated by the MIRT model.}
\label{tab:item_discrimination}
\end{table}

Overall, 76\% of items have discrimination $a > 1.0$, indicating that most scenarios meaningfully differentiate between personas with distinct latent behavioral profiles. Only 2.7\% of items exhibit very low discrimination ($a < 0.5$), suggesting that a small number of scenarios provide limited diagnostic value and may be candidates for refinement.

\subsection{Variance decomposition}
\label{app:mirt_2}

To quantify the relative contributions of persona identity and scenario context to response behavior, we decomposed the variance of responses into persona, scenario, and residual components. The results are shown in Table~\ref{tab:variance_decomposition}.

\begin{table}[H]
\centering
\begin{tabular}{lcc}
\hline
Component & Variance & Percent of total \\
\hline
Persona & 0.090 & 3.75\% \\
Scenario & 0.936 & 38.93\% \\
Residual & 1.378 & 57.32\% \\
\hline
Total & 2.404 & 100\% \\
\hline
\end{tabular}
\caption{Variance decomposition of SJT responses into persona and scenario components.}
\label{tab:variance_decomposition}
\end{table}

Scenario context accounts for approximately 39\% of the variance in responses, while persona identity explains approximately 3.7\%.

To assess statistical significance, we perform a permutation test by randomly shuffling persona labels. The resulting p-value (p = 0.002) indicates that the observed variance is significantly greater than expected under the null hypothesis of no persona effect.

This demonstrates that persona conditioning induces non-trivial and statistically reliable behavioral variation.

The remaining variance reflects residual variability. Importantly, this residual component should not be interpreted as purely stochastic noise. Because the decomposition does not explicitly model persona archetypes, additional persona attributes, or persona-scenario interactions, the residual term likely contains structured behavioral variation arising from latent persona characteristics not captured in the additive model. Consequently, the estimated persona variance should be interpreted as a conservative lower bound on persona-related behavioral structure.

\subsection{Latent trait recovery}
\label{app:mirt_3}

We extract posterior mean estimates of persona-level latent HEXACO traits from the MIRT model, yielding a six-dimensional latent behavioral profile for each persona. These latent traits represent the underlying behavioral tendencies that best explain each persona's pattern of responses across all SJT items.

\begin{figure}[h]
\centering
\includegraphics[width=\linewidth]{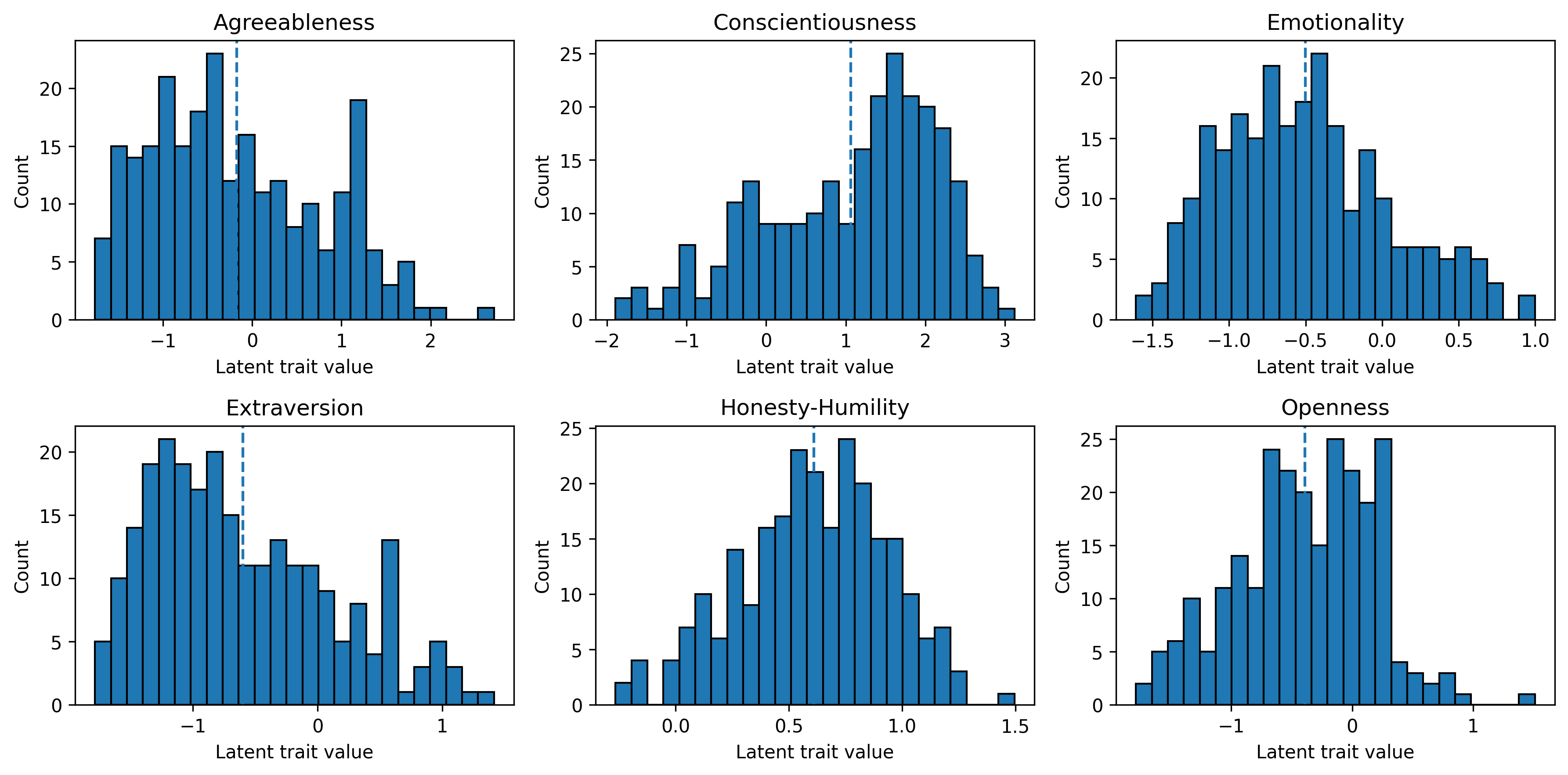}
\caption{
Distribution of recovered latent HEXACO trait values across personas under the confirmatory MIRT model.
Each subplot shows the inferred latent trait values (posterior means) for one personality dimension.
In MIRT, these latent values represent continuous behavioral tendencies inferred from response patterns across scenarios.
All six traits exhibit substantial spread (spanning multiple standard deviations), indicating that the instrument recovers meaningful variation across personas rather than collapsing to a narrow range. The approximate unimodal shapes further suggest that the model is capturing coherent underlying behavioral dimensions rather than noise.
}
\label{fig:hexaco_trait_distributions}
\end{figure}

The MIRT model recovers substantial variation in latent HEXACO traits across personas, with estimated values spanning multiple standard deviations across all six dimensions, indicating meaningful separation in behavioral profiles. See Figure~\ref{fig:hexaco_trait_distributions}.

\subsection{Construct validity: Latent trait--behavior relationship}
\label{app:mirt_4}

To assess construct validity, we evaluated whether latent HEXACO traits predict the selection of trait-aligned response options. For each response option, we fit a logistic regression predicting the probability of selecting that option as a function of all six latent traits. The results are shown in Table~\ref{tab:trait_prediction}.

\begin{table}[h]
\centering
\begin{tabular}{lcc}
\hline
Trait & Coefficient & p-value \\
\hline
Agreeableness & 1.105 & $3.59 \times 10^{-4}$ \\
Conscientiousness & 0.679 & $3.83 \times 10^{-4}$ \\
Emotionality & 1.058 & $1.06 \times 10^{-1}$ \\
Extraversion & 1.379 & $1.47 \times 10^{-2}$ \\
Honesty--Humility & 1.042 & $9.92 \times 10^{-7}$ \\
Openness & 0.884 & $2.72 \times 10^{-2}$ \\
\hline
\end{tabular}
\caption{Logistic regression predicting selection of trait-aligned response options from latent HEXACO trait scores.}
\label{tab:trait_prediction}
\end{table}

Five of the six HEXACO traits significantly predict the selection of their corresponding response options ($p < 0.05$), with consistently positive coefficients. This indicates that higher latent trait values increase the likelihood of selecting trait-aligned responses, supporting the interpretability of the recovered latent dimensions. Emotionality does not reach statistical significance, suggesting weaker alignment for that dimension under the current formulation.

Test--retest reliability is high across all traits (Table~\ref{tab:icc_values}), with intraclass correlation coefficients (ICC) exceeding 0.86, indicating that inferred trait scores are stable across repeated response-generation runs for the same personas.

\begin{table}[h!]
\centering
\begin{tabular}{lc}
\hline
\textbf{Trait} & \textbf{ICC} \\
\hline
\agreeableness\; & 0.992 \\
\conscientiousness\; & 0.990 \\
\emotionality\; & 0.930 \\
\extraversion\; & 0.958 \\
\honestyhumility\; & 0.868 \\
\openness\; & 0.944 \\
\hline
\end{tabular}
\caption{Trait-wise ICC values across repeated response-generation runs for SJT-based \hexaco\; trait scoring.}
\label{tab:icc_values}
\end{table}

Across analyses, the SJT framework demonstrates strong psychometric properties. Most scenarios exhibit moderate to high discrimination, indicating effective differentiation between personas. Persona identity contributes reliable variation in responses, although scenario context remains a dominant factor. The MIRT model recovers latent trait structure that predicts behavioral choices in a theoretically consistent manner.

Overall, these results indicate that the framework produces highly consistent and informative behavioral measurements, supporting its validity as a psychometric instrument.